%% file: acda.tex
\documentclass[lettersize,journal]{IEEEtran}
\usepackage{amsmath,amsfonts}
\usepackage{array}
\usepackage[caption=false,font=normalsize,labelfont=sf,textfont=sf]{subfig}
\usepackage{textcomp}
\usepackage{caption}
\usepackage{stfloats}
\usepackage{url}
\usepackage{verbatim}
\usepackage{graphicx}
\usepackage{cite}
\usepackage{color}
\usepackage{multirow, multicol}
\usepackage{varwidth}
\DeclareCaptionFormat{myformat}{%
  \begin{varwidth}{\linewidth}%
    \centering
    #1#2#3%
  \end{varwidth}%
}
\usepackage[ruled,linesnumbered]{algorithm2e}

\begin{document}

\title{SS-ADA: A Semi-Supervised Active Domain Adaptation Framework for Semantic Segmentation}

\author{Weihao Yan,
Yeqiang Qian,~\IEEEmembership{Member,~IEEE,} 
Yueyuan Li, Tao Li, \\
Chunxiang Wang,~\IEEEmembership{Member,~IEEE,}
and Ming Yang,~\IEEEmembership{Member,~IEEE}
\thanks{This work is supported by the National Natural Science Foundation of China under Grant 62103261. \emph{(Corresponding author: Yeqiang Qian; Ming Yang.)}}
\thanks{Weihao Yan, Yeqiang Qian, Yueyuan Li, Tao Li, Chunxiang Wang and Ming Yang are with the Department of Automation, Shanghai Jiao Tong University, Key Laboratory of System Control and Information Processing, Ministry of Education of China, Shanghai, 200240, China (email: qianyeqiang@sjtu.edu.cn; mingyang@sjtu.edu.cn).}%
}
\vspace{-0.5cm}

\markboth{Journal of \LaTeX\ Class Files,~Vol.~14, No.~8, August~2021}%
{Shell \MakeLowercase{\textit{et al.}}: A Sample Article Using IEEEtran.cls for IEEE Journals}


\maketitle
\input{tex/0_abstract.tex}

\input{tex/1_introduction.tex}

\input{tex/2_related_work.tex}

\input{tex/3_method.tex}

\input{tex/4_experiment.tex}

\input{tex/5_conclusion.tex}


\bibliographystyle{IEEEtran}
\bibliography{IEEEabrv,acda.bib}
\vspace{-0.5cm}



\begin{IEEEbiography}[{\includegraphics[width=1in,height=1.2in,clip,keepaspectratio]{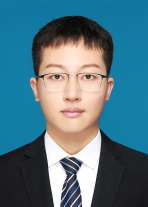}}]{\textbf{Weihao Yan}}
received a Bachelor's degree in Automation from Shanghai Jiao Tong University, Shanghai, China, in 2020. He is working towards a Ph.D. degree in Control Science and Engineering from Shanghai Jiao Tong University.

His main fields of interest are autonomous driving systems, computer vision, and domain adaptation. His current research activities include virtual-to-real transfer learning, scene segmentation, and foundation models.
\end{IEEEbiography}
\vspace{-12.5 mm}
\begin{IEEEbiography}[{\includegraphics[width=1in,height=1.2in,clip,keepaspectratio]{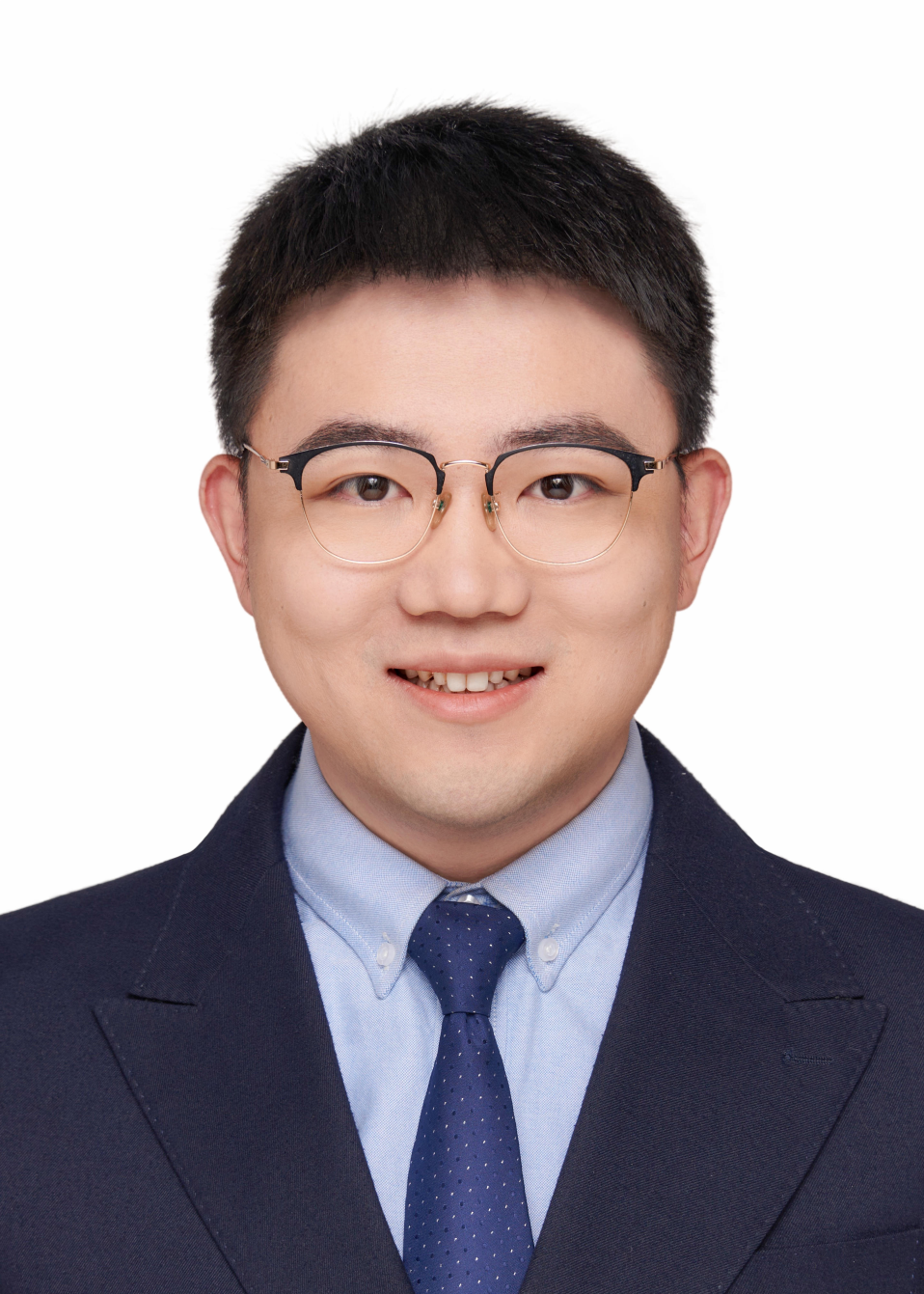}}]{\textbf{Yeqiang Qian}}
received a Ph.D. degree in control science and engineering from Shanghai Jiao Tong University, Shanghai, China, in 2020. He is currently a Tenure Track Associate Professor with the Department of Automation at Shanghai Jiao Tong University, Shanghai, China. \\ 
His main research interests include computer vision, pattern recognition, machine learning, and their applications in intelligent transportation systems.
\end{IEEEbiography}
\vspace{-12.5 mm}
\begin{IEEEbiography}[{\includegraphics[width=1in,height=1.25in,clip,keepaspectratio]{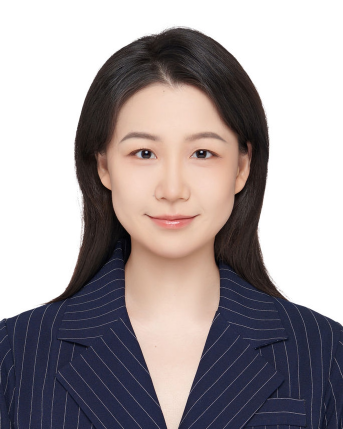}}]{Yueyuan Li} 
received a Bachelor's degree in Electrical and Computer Engineering from the University of Michigan-Shanghai Jiao Tong University Joint Insitute, Shanghai, China, in 2020. She is working towards a Ph.D. degree in Control Science and Engineering from Shanghai Jiao Tong University.

Her main fields of interest are the security of the autonomous driving system and driving decision-making. Her current research activities include driving decision-making models, driving simulation, and virtual-to-real model transferring.
\end{IEEEbiography}
\vspace{-12.5 mm}
\begin{IEEEbiography}[{\includegraphics[width=1in,height=1.25in,clip,keepaspectratio]{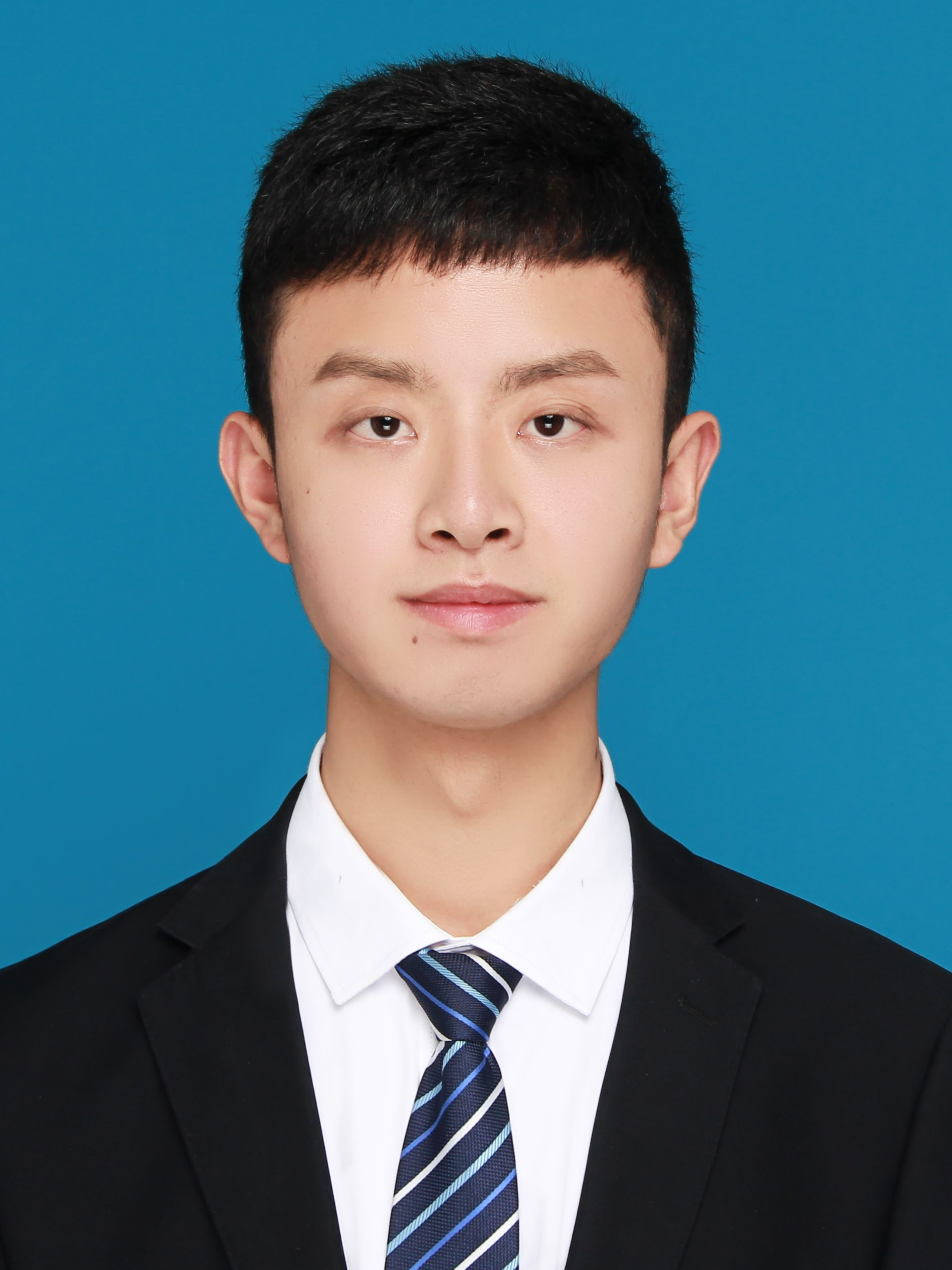}}]{Tao Li} received a Bachelor's degree in Automation from Shanghai Jiao Tong University, Shanghai, China, in 2020. He is working towards a Ph.D. degree in Control Science and Engineering from Shanghai Jiao Tong University. \\
  His main research interests include machine learning and optimization methods.
 \end{IEEEbiography}
 \vspace{-12.5 mm}
\begin{IEEEbiography}[{\includegraphics[width=1in,height=1.2in,clip,keepaspectratio]{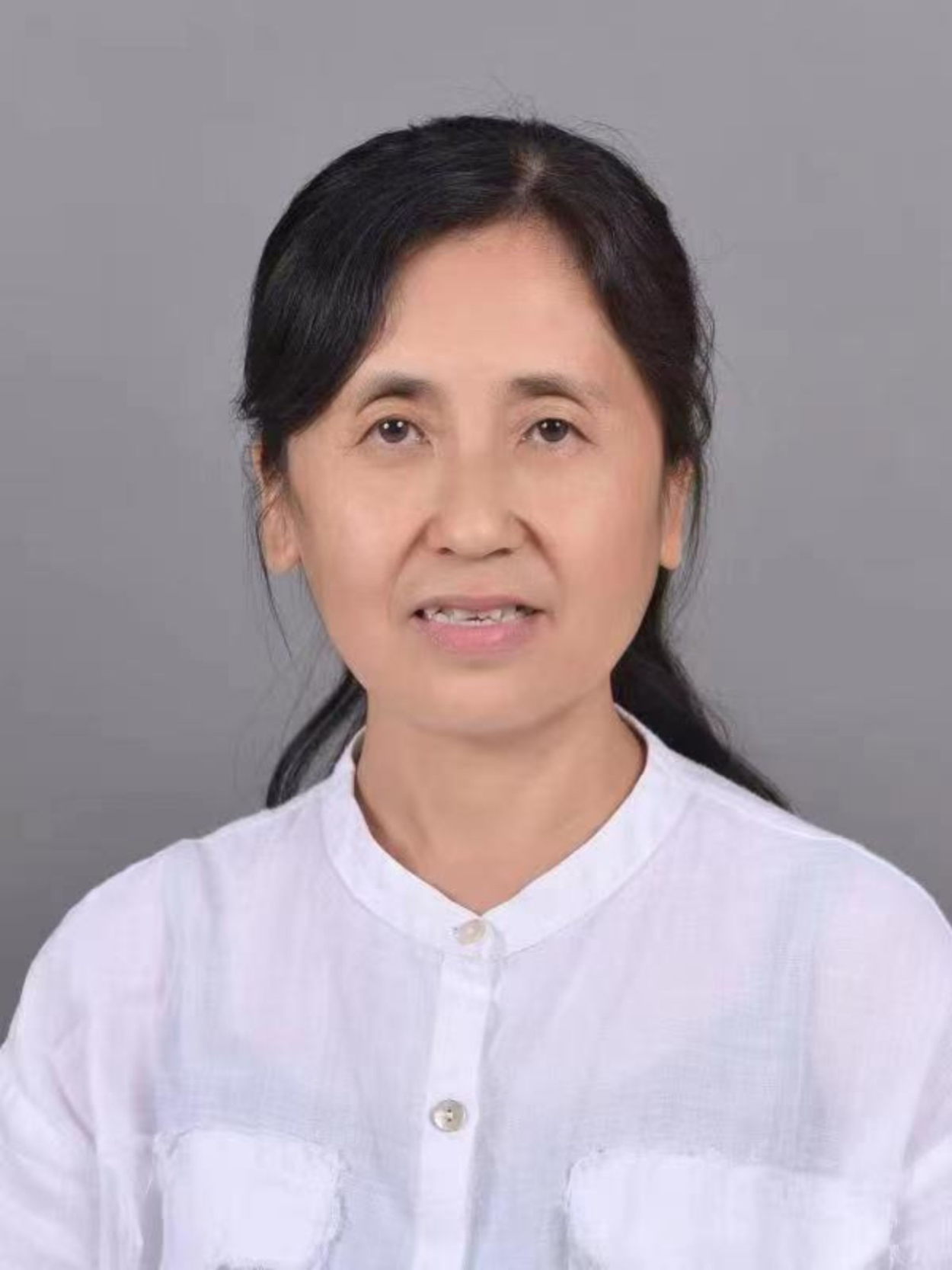}}]{\textbf{Chunxiang Wang}}
received the Ph.D. degree in mechanical engineering from the Harbin Institute of Technology, Harbin, China, in 1999.\\
She is currently an Associate Professor with the Department of Automation at Shanghai Jiao Tong University, Shanghai, China. Her research interests include robotic technology and electromechanical integration.
\end{IEEEbiography}
\vspace{-12.5 mm}
\begin{IEEEbiography}[{\includegraphics[width=1in,height=1.2in,clip,keepaspectratio]{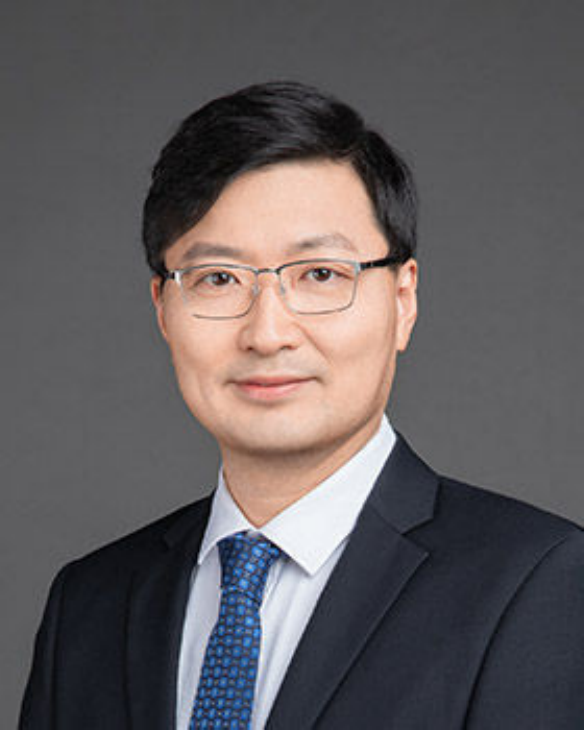}}]{\textbf{Ming Yang}}
received the Master and Ph.D. degrees from Tsinghua University, Beijing, China, in 1999 and 2003, respectively.\\
He is currently the Full Tenure Professor at Shanghai Jiao Tong University, the deputy director of the Innovation Center of Intelligent Connected Vehicles. He has been working in the field of intelligent vehicles for more than 20 years. He participated in several related research projects, such as the THMR-V project (first intelligent vehicle in China), European CyberCars and CyberMove projects, CyberC3 project, CyberCars-2 project, ITER transfer cask project, AGV, etc.
\end{IEEEbiography}

\vfill

\end{document}

%% file: tex/0_abstract.tex
\begin{abstract}
    Semantic segmentation plays an important role in intelligent vehicles, providing pixel-level semantic information about the environment. However, the labeling budget is expensive and time-consuming when semantic segmentation model is applied to new driving scenarios. To reduce the costs, semi-supervised semantic segmentation methods have been proposed to leverage large quantities of unlabeled images. Despite this, their performance still falls short of the accuracy required for practical applications, which is typically achieved by supervised learning. A significant shortcoming is that they typically select unlabeled images for annotation randomly, neglecting the assessment of sample value for model training. In this paper, we propose a novel semi-supervised active domain adaptation (SS-ADA) framework for semantic segmentation that employs an image-level acquisition strategy. SS-ADA integrates active learning into semi-supervised semantic segmentation to achieve the accuracy of supervised learning with a limited amount of labeled data from the target domain. Additionally, we design an IoU-based class weighting strategy to alleviate the class imbalance problem using annotations from active learning. We conducted extensive experiments on synthetic-to-real and real-to-real domain adaptation settings. The results demonstrate the effectiveness of our method. SS-ADA can achieve or even surpass the accuracy of its supervised learning counterpart with only 25\% of the target labeled data when using a real-time segmentation model. The code for SS-ADA is available at \url{https://github.com/ywher/SS-ADA}.
\end{abstract}


%% file: tex/1_introduction.tex
\section{Introduction}
\IEEEPARstart{I}{mage} semantic segmentation is a crucial perception task in intelligent transportation system \cite{hyseg,restricted,gated,segtransconv,transfer,cmx}, facilitating various downstream tasks such as drivable area detection, semantic-based simultaneous localization and mapping. Recent years have seen remarkable progress in semantic segmentation, driven by the availability of large-scale annotated datasets and advancements in deep learning. However, deep learning-based semantic segmentation methods require extensive annotated datasets for training, and obtaining pixel-wise labels is costly. For instance, the labor-intensive annotation process takes approximately 1.5 hours for a single Cityscapes image, and up to 3.3 hours for an image captured under adverse weather conditions\cite{cityscapes, acdc}. 

\begin{figure}[tbp]
    \centering
    \captionsetup[subfloat]{font=scriptsize,labelfont=scriptsize,labelformat=empty}
    \vspace{-0.2cm}
    \hfill \subfloat[]{\includegraphics[width=0.98\linewidth]{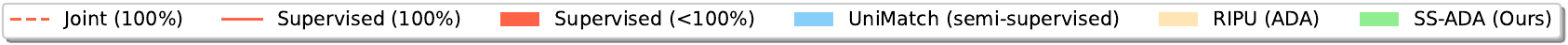}}\\ \vspace{-0.7cm}
    \subfloat[GTA5-to-Cityscapes]{\includegraphics[width=0.5\linewidth]{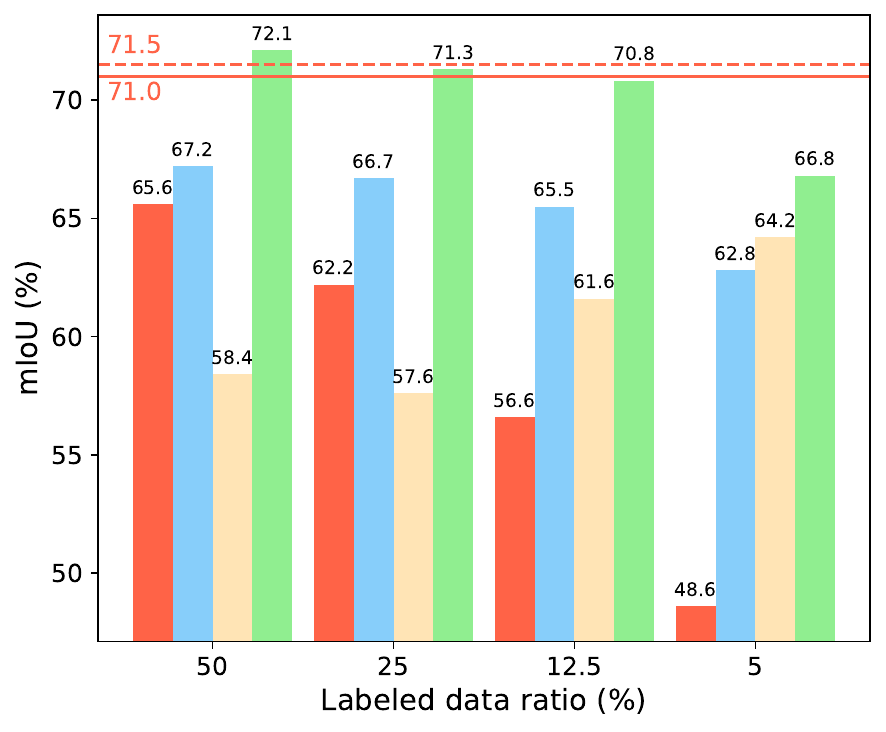}}
    \subfloat[Cityscapes-to-ACDC]{\includegraphics[width=0.5\linewidth]{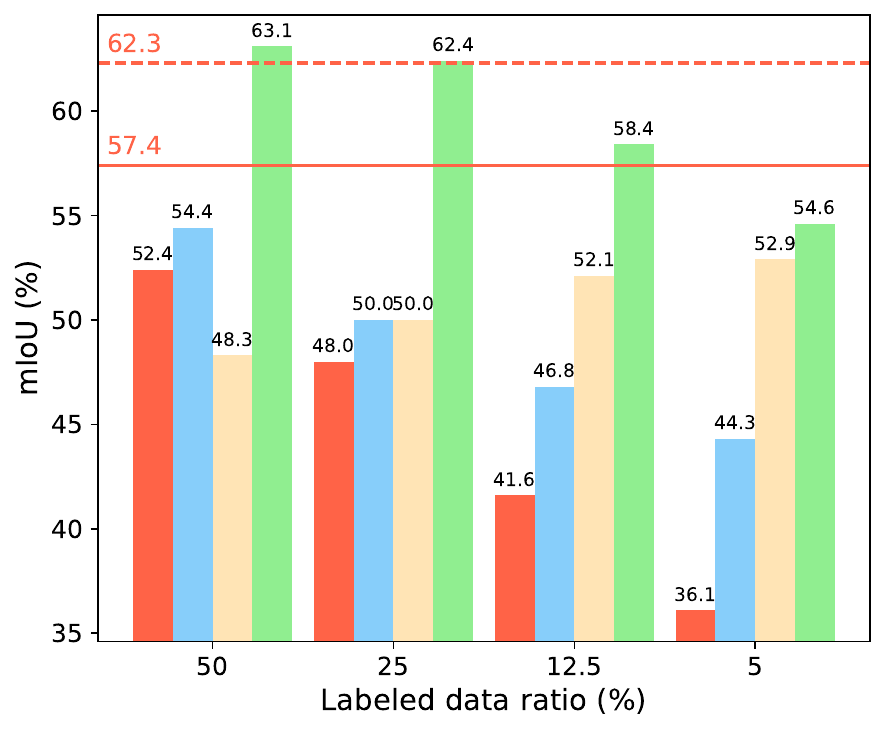}}
    \caption{The performance of joint training, supervised and semi-supervised learning, active domain adaptation and SS-ADA on synthetic-to-real and real-to-real domain adaptation settings.}
    \label{fig:intro}
    \vspace{-0.5cm}
\end{figure}

To reduce the labeling cost, semi-supervised semantic segmentation methods have garnered attention in recent years\cite{tits-semi,u2pl,unimatch}. These methods use a small amount of labeled data alongside a large amount of unlabeled data to enhance the model's performance\cite{u2pl,unimatch,seg2depth,tits-semi}. However, as shown in Figure~\ref{fig:intro}, even with 50\% of the data annotated, the performance of semi-supervised semantic segmentation methods has not yet matched the accuracy of fully supervised methods\cite{u2pl,unimatch}, which is more desirable in practical applications due to its higher precision. One potential reason is that they typically employ simple random sampling for annotating the unlabeled data. This approach overlooks the selection of particularly informative samples that could enhance model training. In this paper, we explore incorporating active learning into semi-supervised semantic segmentation to perform valuable sample selection. Additionally, we introduce domain adaptation and leverage a large amount of annotated source domain data from driving simulators or public datasets to aid training.

Active learning methods employ a human-in-the-loop mechanism to enhance model performance using a limited amount of annotated data. These methods develop acquisition strategies to select informative samples based on uncertainty sampling, diversity sampling, or a combination of both \cite{deepbayesian, wu2021redal, ripu}. Building on active learning, active domain adaptation (ADA) methods aim to transfer knowledge from a labeled source domain to an unseen target domain\cite{activecontrastive,active_tits,activesurvey,ripu,d2ada}. ADA methods leverage the abundance of synthetic data generated from driving simulators to facilitate model training\cite{playingfordata,synthia,simulator}. However, the incorporation of ADA into semi-supervised semantic segmentation remains under-explored. Meanwhile, pixel or region-based ADA methods suffer from class imbalance problem\cite{ripu}. As the labeling budget exceeds a certain threshold, such as 10\%, the model's performance tends to gradually deteriorate\cite{ripu,halo}, as highlighted in Figure~\ref{fig:intro}. Research by \cite{halo} revealed that applying all labels in domain adaptation often results in the majority of labels belonging to a few dominant classes, such as road, building, and vegetation. This issue limits the effectiveness of these methods, leading to the use of only a small proportion of annotations (e.g., 5\%) in ADA experiments, which constrains further reduction of the gap between semi-supervised and supervised accuracy.
In this paper, we propose a novel semi-supervised active domain adaptation framework named SS-ADA for semantic segmentation, which integrates active learning into semi-supervised semantic segmentation methods. The framework consists of three core modules: a semi-supervised learning module, an active learning module, and an IoU-based class weighting module. The semi-supervised learning module leverages unlabeled data from the target domain, while the active learning module selects informative samples from the unlabeled data for manual annotation. The Intersection-over-Union (IoU) based class weighting strategy assigns greater weights to classes with lower IoU to alleviate the class imbalance problem. We conduct extensive experiments on synthetic-to-real and real-to-real domain adaptation settings. The experimental results demonstrate that SS-ADA achieves accuracy comparable to that of fully supervised training using only 25\% of labeled data from the target driving scene. This significantly reduces the annotation requirements for deploying semantic segmentation models in target driving scenarios. Our contributions are summarized as follows:  

\begin{itemize}
    \item We propose a novel semi-supervised active domain adaptation framework, SS-ADA, for semantic segmentation. SS-ADA introduces active learning into semi-supervised segmentation methods to select the more informative unlabeled samples for human annotation and model training. 
    \item We design an IoU-based class weighting strategy based on the annotation from active learning. It assigns higher training loss weights to classes with lower IoU and vice versa, alleviating the class imbalance problem. 
    \item Extensive experiments on synthetic-to-real and real-to-real domain adaptation settings  demonstrate the effectiveness of SS-ADA. It achieves accuracy comparable to fully supervised learning counterparts using only 25\% of labeled data in the target domain. 
\end{itemize}

%% file: tex/2_related_work.tex
\section{Related Work}

\subsection{Semantic Segmentation}

Semantic segmentation is a pivotal perception task in intelligent transportation system\cite{restricted,gated,segtransconv,hyseg,transfer,cmx}.  One category of methods aims at enhancing segmentation accuracy through robust model architectures. For instance, CMX proposes a universal transformer-based cross-modal fusion architecture for RGB-X semantic segmentation\cite{cmx}. Another category of research focuses on balancing inference speed and model precision. SegTransConv is based on the hybrid architecture of CNN and transformer for real-time semantic segmentation of autonomous vehicles\cite{segtransconv}. BiSeNet\ achieves this balance by integrating a spatial path and a context pathcite{bisenet}. In our experiments, we adopt the BiSeNet for real-time inference. 
\vspace{-0.2cm}


\subsection{Semi-supervised Semantic Segmentation}
Semi-supervised semantic segmentation methods aim to improve the segmentation performance using a limited amount of labeled data and a large-scale unlabeled data directly from the target domain. These methods can be mainly divided into two categories: consistency regularization and entropy minimization. Consistency regularization methods enforce the model to produce consistent predictions on unlabeled data under perturbations such as data augmentation, mixed samples, or feature perturbations\cite{Augmentation, classmix, unimatch}. Entropy minimization methods focus on assigning pseudo-labels to unlabeled target data and using them for self-training, such as in ST++ and USRN\cite{st++,usrn}. Recently, CorrMatch has used correlation matching to discover more accurate high-confidence regions\cite{CorrMatch}. Despite making steady progress in recent years, semi-supervised methods still fall short of the accuracy achieved by supervised learning counterparts due to noisy pseudo-labels and a lack of accurate semantic guidance. Additionally, these methods often overlook the utilization of label-rich source domains and the selection of informative samples.
\vspace{-0.3cm}

\subsection{Active Domain Adaptation for Semantic Segmentation}

Domain adaptation (DA) methods aim to transfer the knowledge learned from a labeled source domain to a target domain with different distribution. Recent advancements in DA for semantic segmentation primarily focus on unsupervised domain adaptation\cite{cat_adversal,daformer,mic,sam4udass}. They assume that the target domain has no labeled data and the corresponding performance still lag far behind supervised learning counterparts.

Researchers have turned to active learning to improve adaptation performance with human annotation. Active learning aims to maximize model performance with limited labeled data by selecting the more valuable samples for annotation \cite{tits_active}. For example, Gao proposes an active and contrastive learning framework for fine-grained off-road semantic segmentation\cite{activecontrastive}. LabOR introduces a novel point-based annotation strategy with an adaptive human-in-the-loop pixel selector but overlooks the contextual relationships between pixels in an image\cite{labor}. RIPU designs a region-based acquisition strategy that uses region impurity and prediction uncertainty to select regions that are uncertain in prediction and diverse in spatial adjacency\cite{ripu}. Subsequently, D2ADA  employs density-aware measurements to acquire label annotations\cite{d2ada}. HALO  introduces a hyperbolic radius into the acquisition strategy for more effective data annotation and highlights the class imbalance problem with increasing budget\cite{halo}.

Current ADA methods for semantic segmentation are dominant by pixel or region-level acquisition strategies. However, these forms are not suitable for human labeling, which may not actually reduce the difficulty and cost of annotation\cite{iterative}. Moreover, most existing ADA methods neglect the  potential of unlabeled data for training, which is available and usually large-scale in driving scenarios. In this paper, we introduce active learning into semi-supervised semantic segmentation methods to select informative samples for human annotation and model training. We adopt an image-based acquisition strategy and annotation form to preserve contextual information. 

%% file: tex/3_method.tex
\section{Method}

\subsection{Preliminaries}
\subsubsection{Data representation}
Let the labeled data in the source domain be denoted as $D_s^l=(x_s^l, y_s^l)$, where $x_s^l \in R^{3 \times H \times W}$ and $y_s^l \in R^{H \times W}$ present the image and label, H and W are the height and width, respectively. The images in the target domain are initially without annotations and are denoted as $D_t^u=x_t^u$. After human annotation, the target labeled data is represented as $D_t^l=(x_t^l, y_t^l)$.

\subsubsection{Semi-supervised semantic segmentation} Both $D_t^l$ and $D_t^u$ in the target domain are used for model training. For $D_t^l$, the supervised cross-entropy loss can be applied directly. To utilize the potential of $D_t^u$, consistency regularization loss, entropy minimization loss or other semi-supervised learning methods can be employed. Denote the model's prediction for $x_t^u$ as $p_t^u$, the general loss function for semi-supervised semantic segmentation is defined as follows:
\begin{equation}
    \centering
    L_{semi} = L(p_t^l, y_t^l) + \lambda L(p_t^u)
\end{equation}
where $L(p_t^u)$ is the unsupervised loss designed by semi-supervised semantic segmentation methods, $\lambda$ is the coefficient for loss balancing, $L(p, y)$ is usually the cross-entropy loss:
\begin{equation}
    \centering
    L(p, y) = -\frac{1}{N} \sum_{i=1}^{N} \sum_{j=1}^{C} y^{(i,j)} \log p^{(i,j)}
    \label{eq:ce_loss}
\end{equation}
where $N$ is the number of pixels, $C$ is the number of classes. $i$ is the pixel index, and $j$ is the channel index.

\subsubsection{ADA for semantic segmentation} 
In active learning, part of the samples in $D_t^u$ are selected for human annotation to obtain the target labeled data $D_t^l=(x_t^l, y_t^l)$. The selection and annotation are conducted at $N$ active learning trigger epoches $T_{ac}=[T_{a1}, T_{a2}, \ldots, T_{aN}]$ to acquire a total of $N_s$ samples. The goal of ADA for semantic segmentation is to leverage $D_s^l$ and $D_t^l$ to train a segmentation model that performs well on the target domain. Denote the model's prediction results for $x_s^l$ and $x_t^l$ as $p_s^l$ and $p_t^l$, respectively. The general loss function of ADA for semantic segmentation is defined as follows:

\begin{equation}
    \centering
    L_{active} = L(p_s^l, y_s^l) + L(p_t^l, y_t^l)
    \label{eq: loss_active}
\end{equation}
where $L(p_s^l, y_s^l)$ and $L(p_t^l, y_t^l)$ are usually cross-entropy losses calculated according to Equation~\eqref{eq:ce_loss}.

\subsection{Overall Framework}
\begin{figure*}[htbp]
    \centering
    \includegraphics[width=0.999\linewidth]{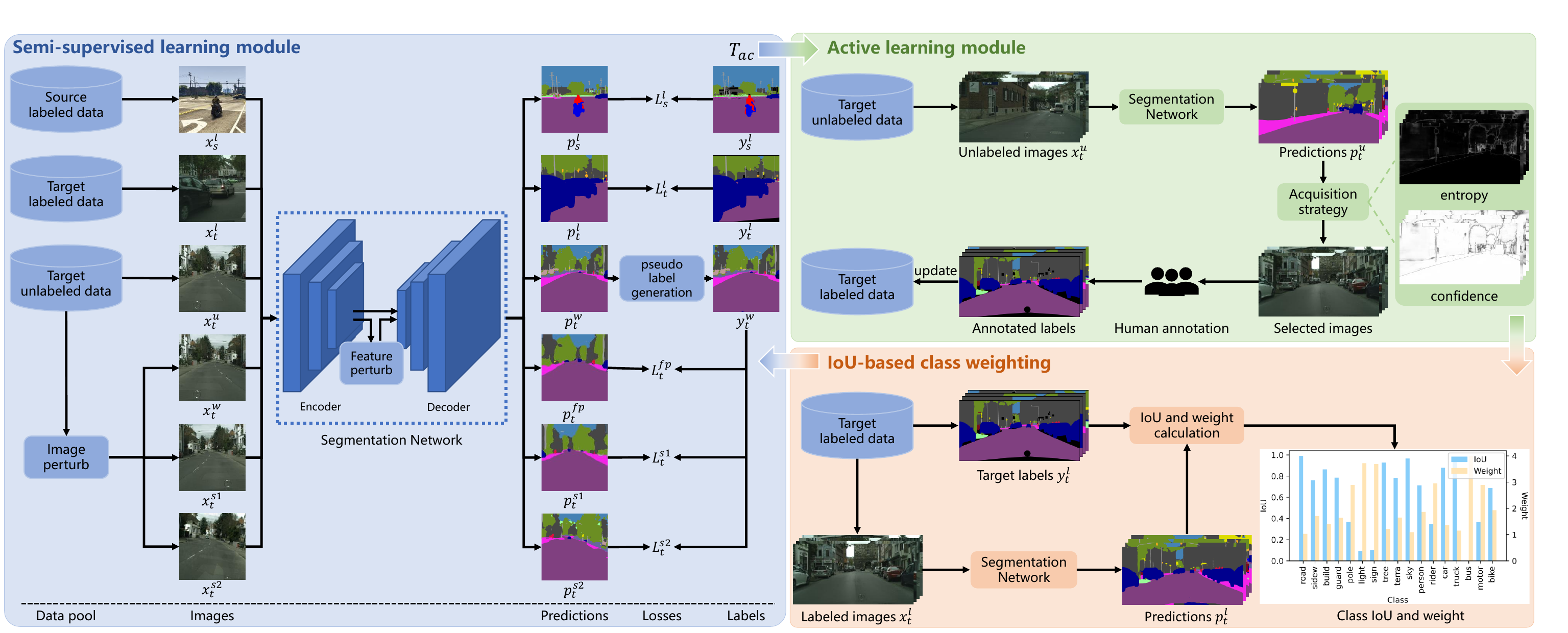}
    \caption{The training framework of SS-ADA for semantic segmentation. The semi-supervised learning module leverages source labeled data, target labeled data, and target unlabeled data for training. The active learning module is triggered at specified training epochs $T_{ac}$ to evaluate the value of the unlabeled data, select a portion of them for manual annotation, and update the target domain data accordingly. IoU-based class weighting strategy is applied after active learning module and calculate the class weights. After that, the training process goes back to semi-supervised learning module.} 
    \label{fig:whole_framework}
    \vspace{-0.4cm}
\end{figure*}

The overall training framework of SS-ADA for semantic segmentation is shown in Figure~\ref{fig:whole_framework}. SS-ADA has three key components: the semi-supervised learning module, the active learning module and the IoU-based class weighting. The semi-supervised learning module is designed to leverage the source labeled data $D_{s}^l$, target labeled data $D_{t}^l$, and target unlabeled data $D_{t}^u$ for model training, fully utilizing the available data, especially the potential of unlabeled target data. The training of the segmentation model is primarily conducted within the semi-supervised module. 

When the training epoch reaches the specified $T_{ac}$ rounds, the active learning module is triggered. It evaluates the value of the remaining unlabeled target domain data $D_{t}^u$ and selects a portion of them based on the acquisition strategy for manual annotation. The selected images and corresponding annotated labels are added to $D_t^l$ for subsequent calculation and training. 

After $D_t^l$ is updated in active learning module, the IoU-based class weighting strategy is applied. It calculates the class-wise IoU of the data in $D_t^l$ and determines the loss weights for the semi-supervised learning module. Following this, the training process goes back to semi-supervised module with the updated $D_t^l$, $D_t^u$, and class weights. 

These three modules work together to improve the model's performance in the target domain. The details of these components are described in the following sections. 
\vspace{-0.2cm}

\subsection{Semi-supervised Learning Module}
The data used for semi-supervised learning includes source labeled data $D_{s}^l$, target labeled and unlabeled data $D_{t}^l$ and $D_{t}^u$. Since the images from the target domain lack annotations initially, the semi-supervised learning process cannot start directly. Therefore, we uniformly randomly sampled 1\% of the target data for human annotation, forming the initial target labeled data pool $D_{t}^l$. Optionally, UDA methods can also be utilized for model initialization. Once the active learning module is triggered, a portion of the data in $D_{t}^u$ is annotated and added to $D_{t}^l$, allowing the model to be trained normally within the semi-supervised learning module. 

For the labeled images $x_s^l$ and $x_t^l$, they pass through the network and obtain their predictions $p_s^l$ and $p_t^l$. The cross-entropy losses $L_s^l(p_s^l, y_s^l)$ and $L_t^l(p_t^l, y_t^l)$ are calculated for the predictions and labels according to Equation~\eqref{eq:ce_loss}.

For the unlabeled images $x_t^u$, we use the weak-to-strong consistency training inspired by\cite{unimatch,TUFL}. The core idea is to apply different levels of perturbations to the images, including image-level perturbations and feature-level perturbations, ensuring that the model produces consistent predictions for the perturbed image pairs. Denote the weak and strong image perturbations as $g_w$ and $g_s$, respectively. The weak perturbations $g_w$ include random resizing, cropping and random horizontal flipping. The strong perturbations $g_s$ consist of color jitter, grayscale conversion, Gaussian blur and Cutmix\cite{classmix}. Denote the feature perturbation as $g_f$, which refers to applying the Dropout\cite{dropout} to the features extracted by the encoder. 

In the ``Image perturb'' step in Figure~\ref{fig:whole_framework}, one instance of $g_w$ and two instances of $g_s$ are applied to $x_t^u$, resulting in $x_t^w$, $x_t^{s1}$ and $x_t^{s2}$. The corresponding model predictions of these perturbed images are denoted as $p_t^w$, $p_t^{s1}$ and $p_t^{s2}$, respectively. To construct a broader perturbation space, the feature perturbation $g_f$ is applied to the features of $x_t^w$ extracted by the encoder in the ``Feature perturb'' step in Figure~\ref{fig:whole_framework}, additionally producing the prediction $p_t^{fp}$. Following previous UDA methods\cite{daformer, mic,sam4udass}, $p_t^w$ is used to generate the pseudo-label $y_t^{w}$ for the unlabeled data based on the confidence threshold $c_{t}$: 
\begin{equation}
    y_t^w=\underset{\text{dim}=C}{argmax}(p_t^w)[\underset{\text{dim}=C}{max}(p_t^w)>c_{t}]
    \label{eq: pseudo_label}
\end{equation}
where $\underset{\text{dim}=C}{argmax}(p_{t\_w})$ and $\underset{\text{dim}=C}{max}(p_{t\_w})$ denote the prediction and the confidence of the model, respectively. $c_{t}$ is within the range of $[0, 1]$.
The cross-entropy losses $L_t^{fp}(p_t^{fp}, y_t^{w})$, $L_t^{s1}(p_t^{s1}, y_t^{w})$ and $L_t^{s2}(p_t^{s2}, y_t^{w})$ are also used for these unlabeled predictions and pseudo-labels following Equation~\eqref{eq:ce_loss}. The loss function of the semi-supervised learning module is:
\begin{equation}
    \centering
    L_{semi} = L_s^l + L_t^l + \lambda (L_t^{fp} + L_t^{s1} + L_t^{s2})
\end{equation}
where $\lambda$ is the weight for the consistency regularization loss and is set as $\frac{1}{3}$ in our experiments.



\subsection{Active Learning Module}

The active learning module is triggered at specified training epochs $T_{ac} = [T_{a1}, T_{a2}, \ldots, T_{aN}]$ and select a total of $N_s$ images from the target unlabeled data $D_t^u$ for manual annotation. These annotated images and labels are then incorporated into the target labeled data $D_{t}^l$. In our experiments, the triggering epochs for active learning are set within the first half of the total training epochs. The number of samples needed for annotation is uniformly distributed among these sampling epochs. Specifically, at training epoches $[T_{a1}, T_{a2}, \ldots, T_{aN}]$, the active learning module will select $\frac{N_s}{N}$ unlabeled samples from $D_t^u$ for manual annotation and update $D_{t}^l$.

\begin{figure}[tbp]
    \vspace{-0.3cm}
    \centering
    \captionsetup[subfloat]{font=scriptsize,labelfont=scriptsize,labelformat=empty}
    \subfloat{\includegraphics[width=0.249\linewidth]{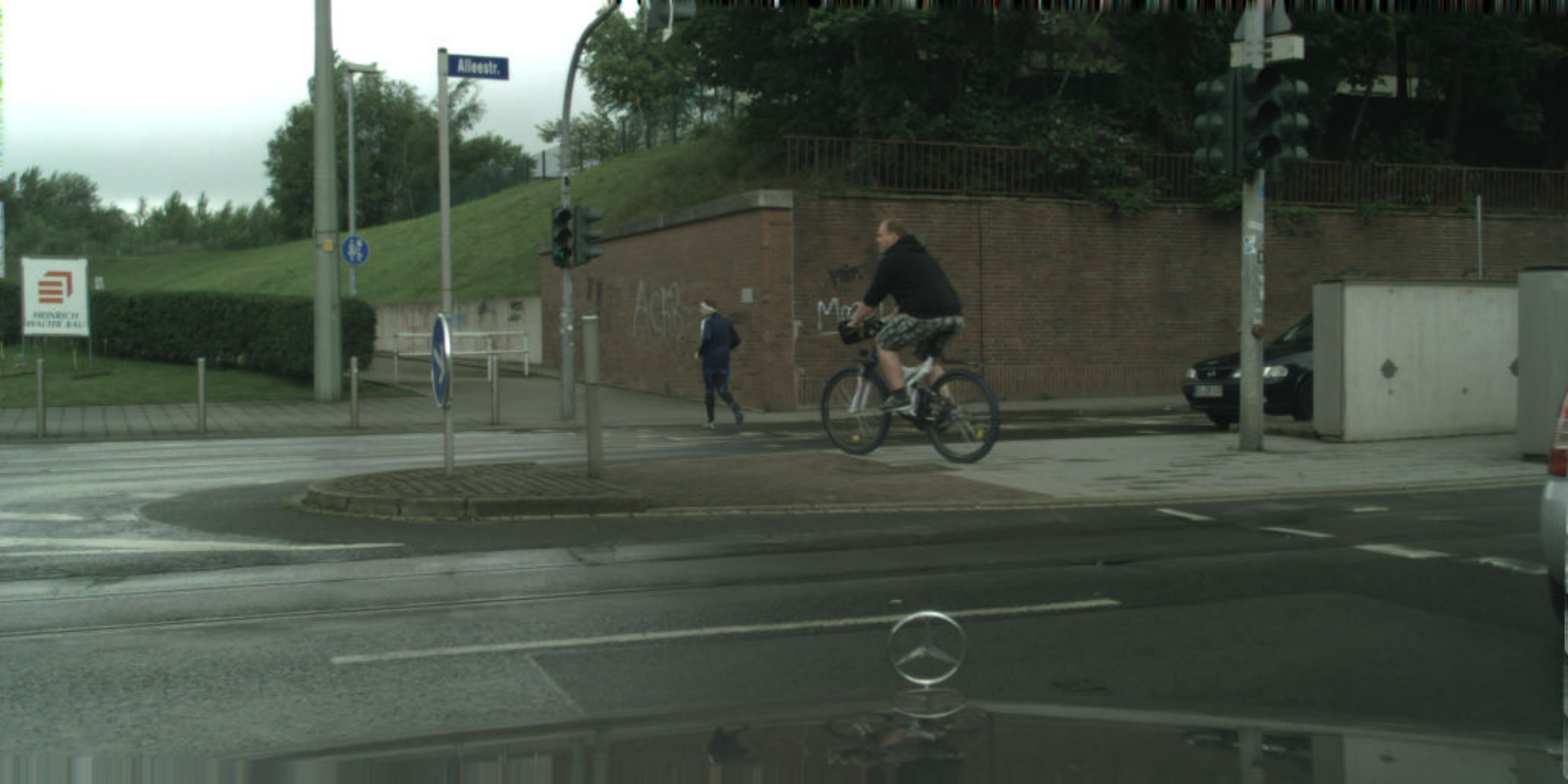}}\hfill
    \subfloat{\includegraphics[width=0.249\linewidth]{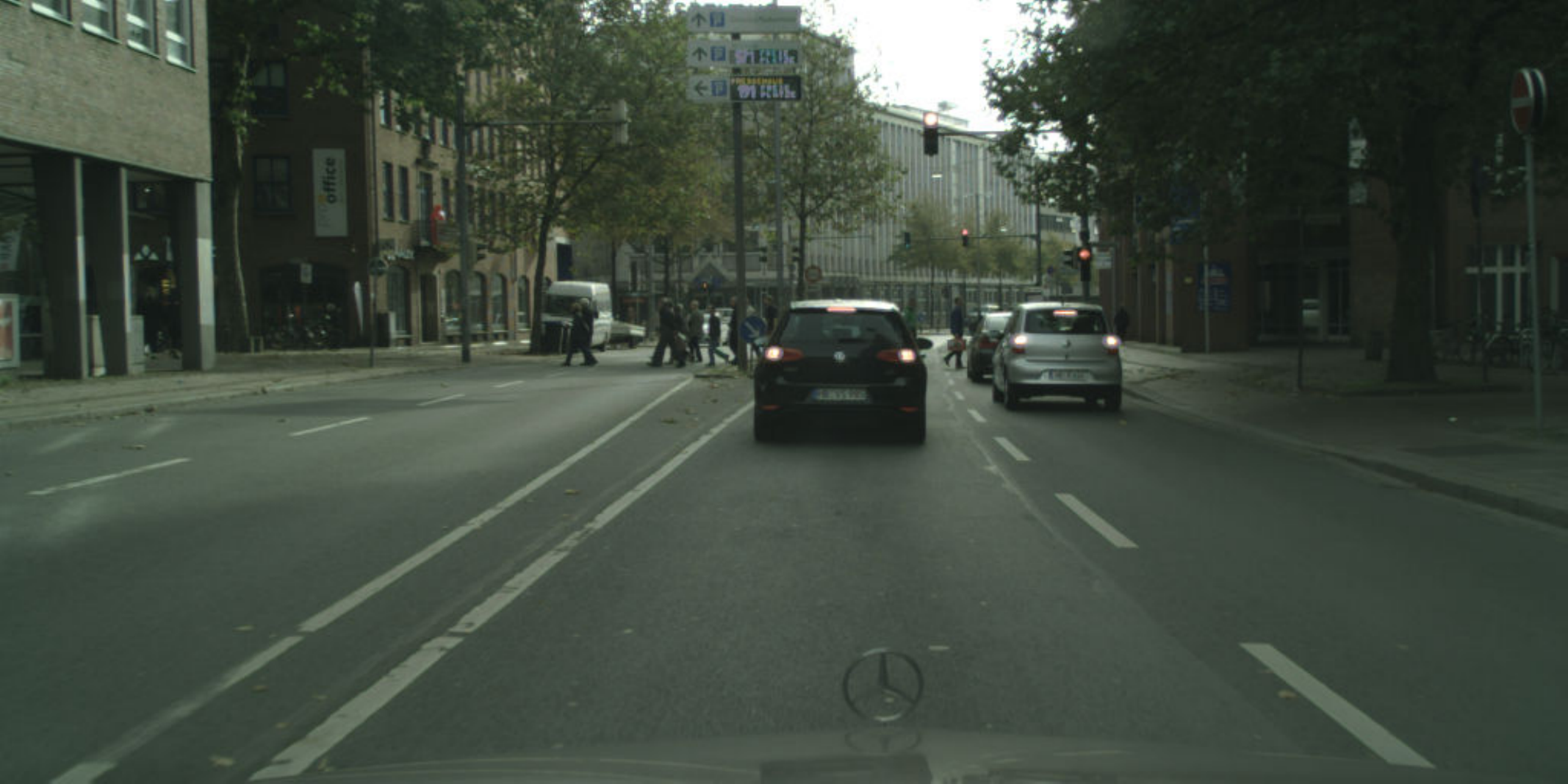}}\hfill
    \subfloat{\includegraphics[width=0.249\linewidth]{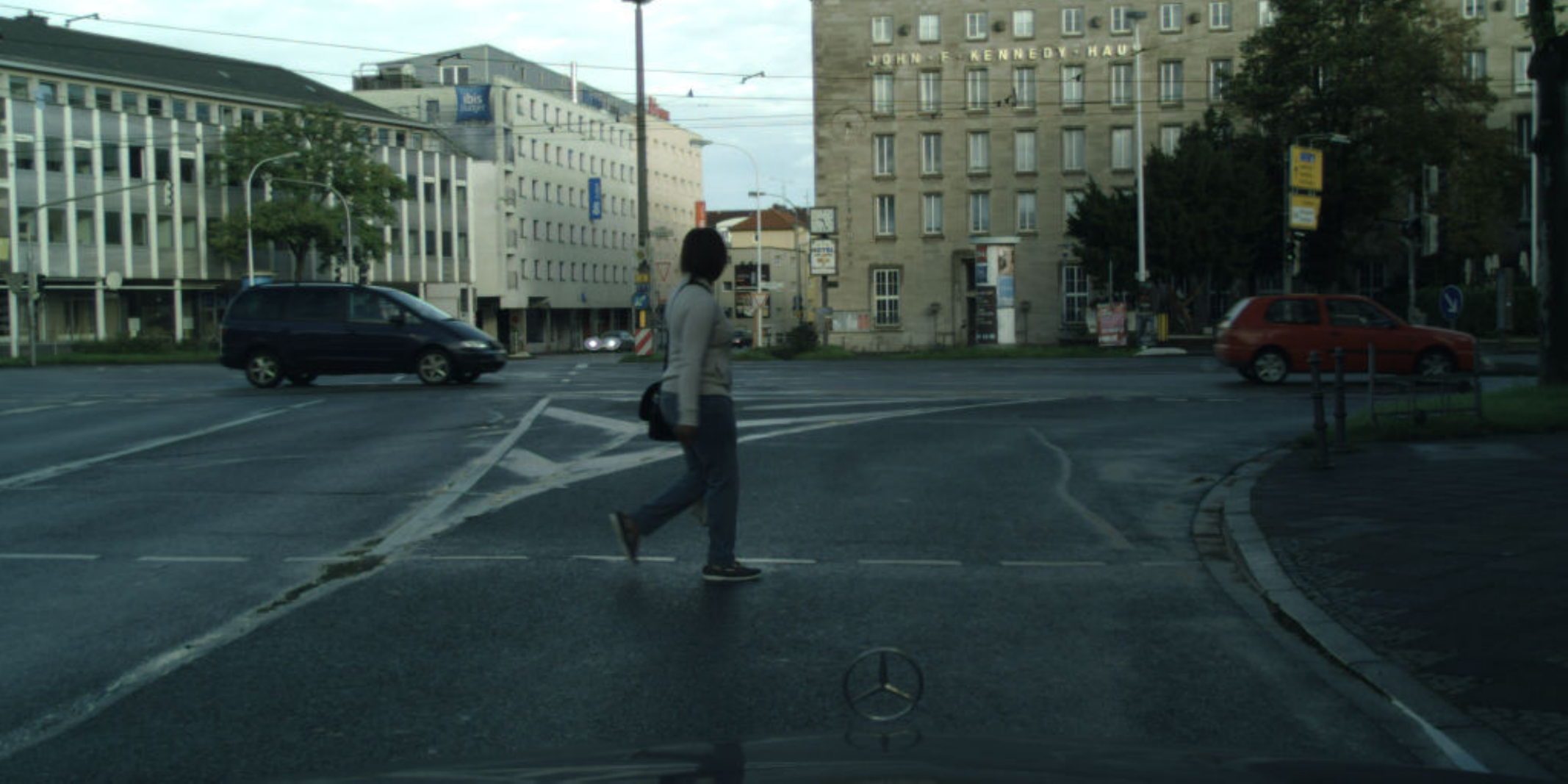}}\hfill
    \subfloat{\includegraphics[width=0.249\linewidth]{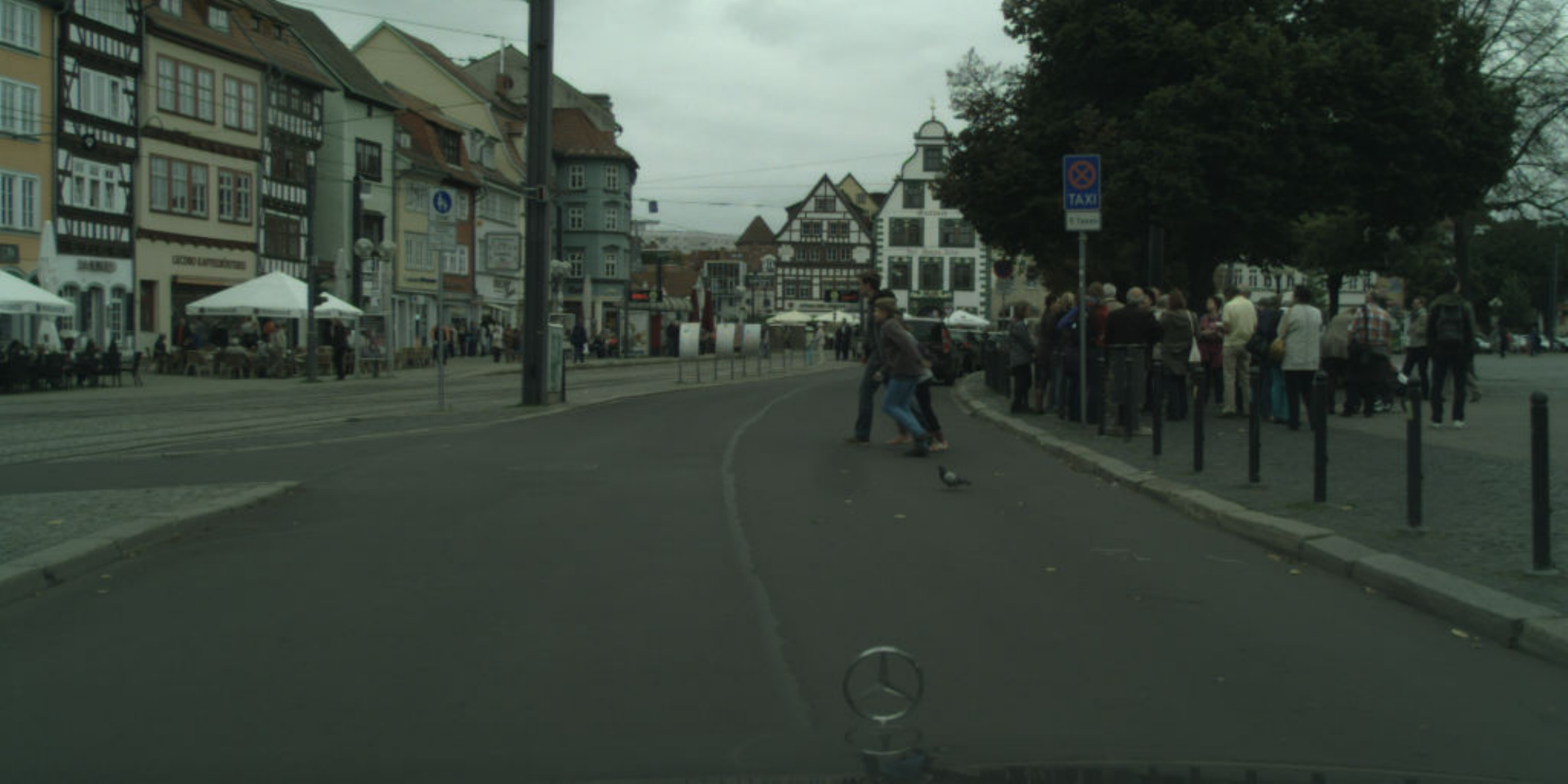}}\hfill \\ \vspace{-0.325cm}
    \subfloat{\includegraphics[width=0.249\linewidth]{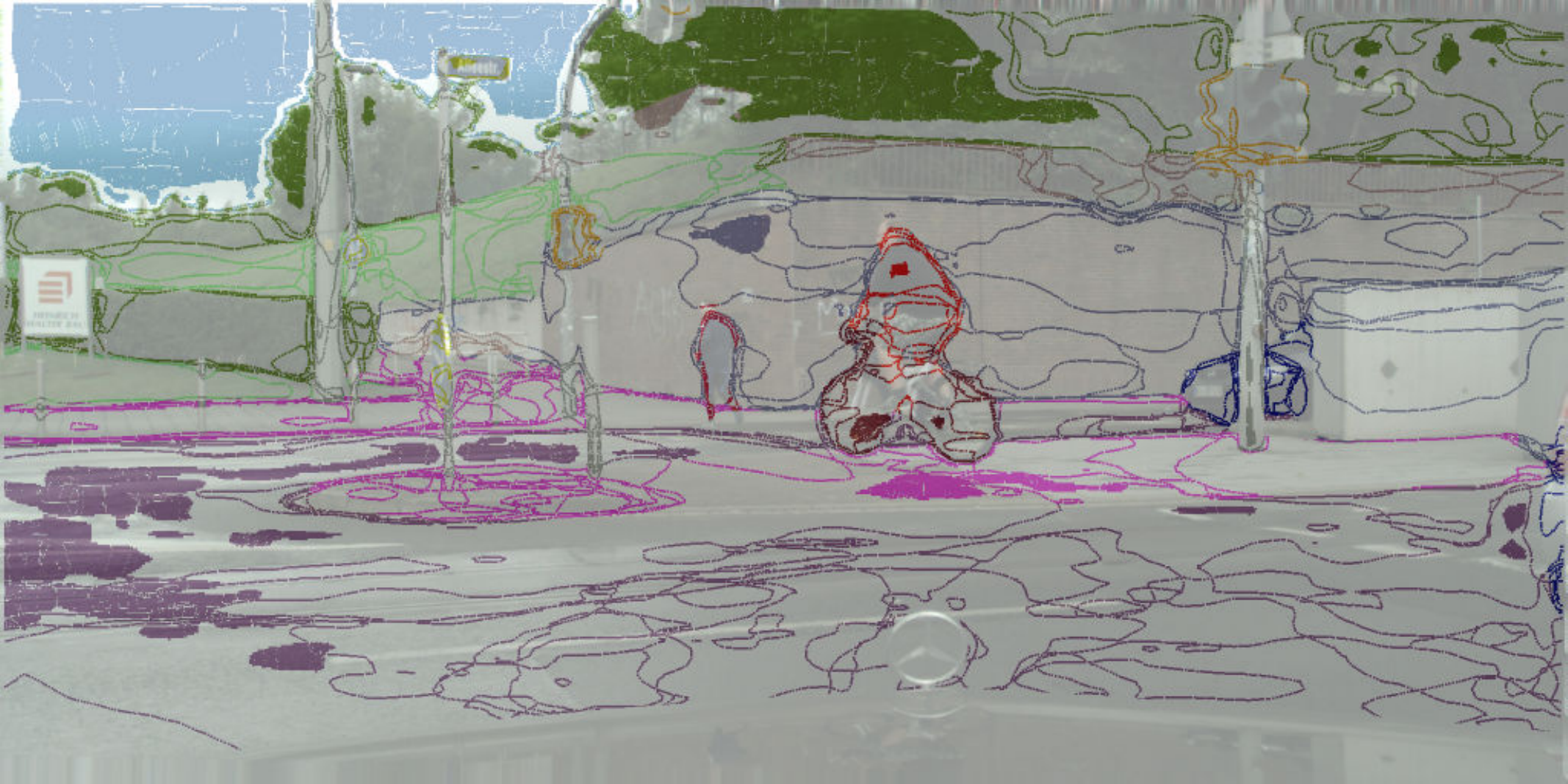}}\hfill
    \subfloat{\includegraphics[width=0.249\linewidth]{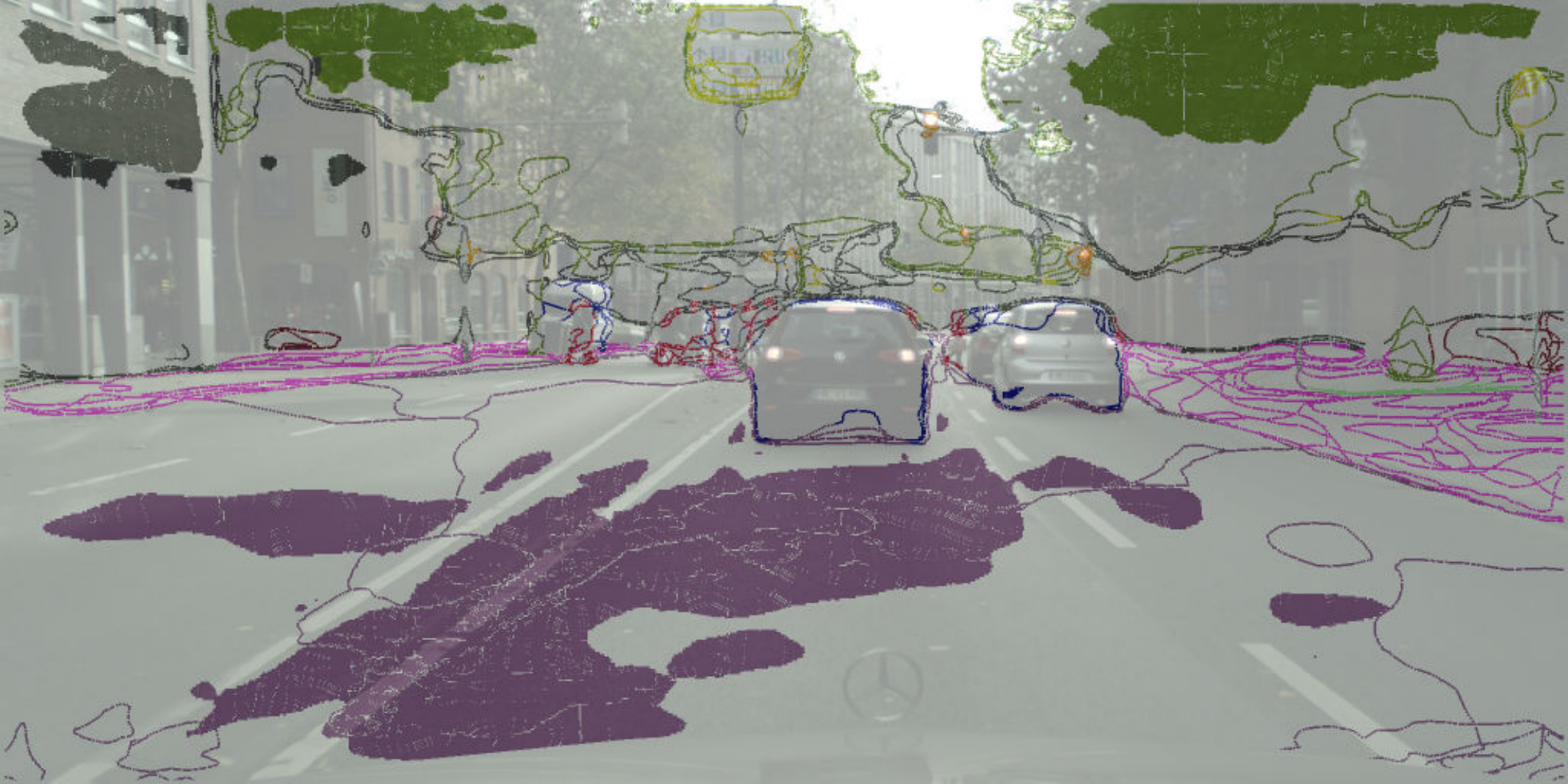}}\hfill
    \subfloat{\includegraphics[width=0.249\linewidth]{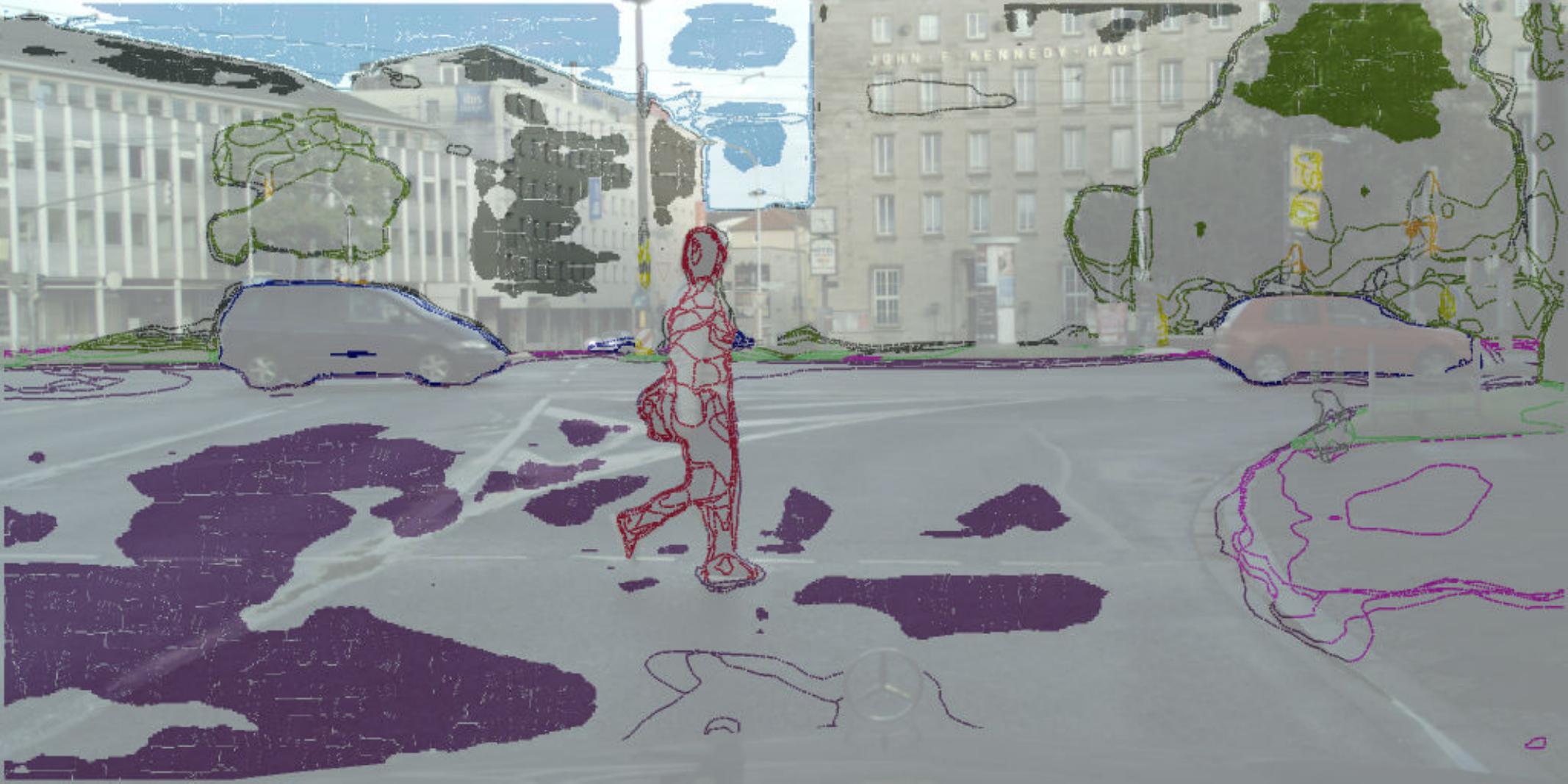}}\hfill
    \subfloat{\includegraphics[width=0.249\linewidth]{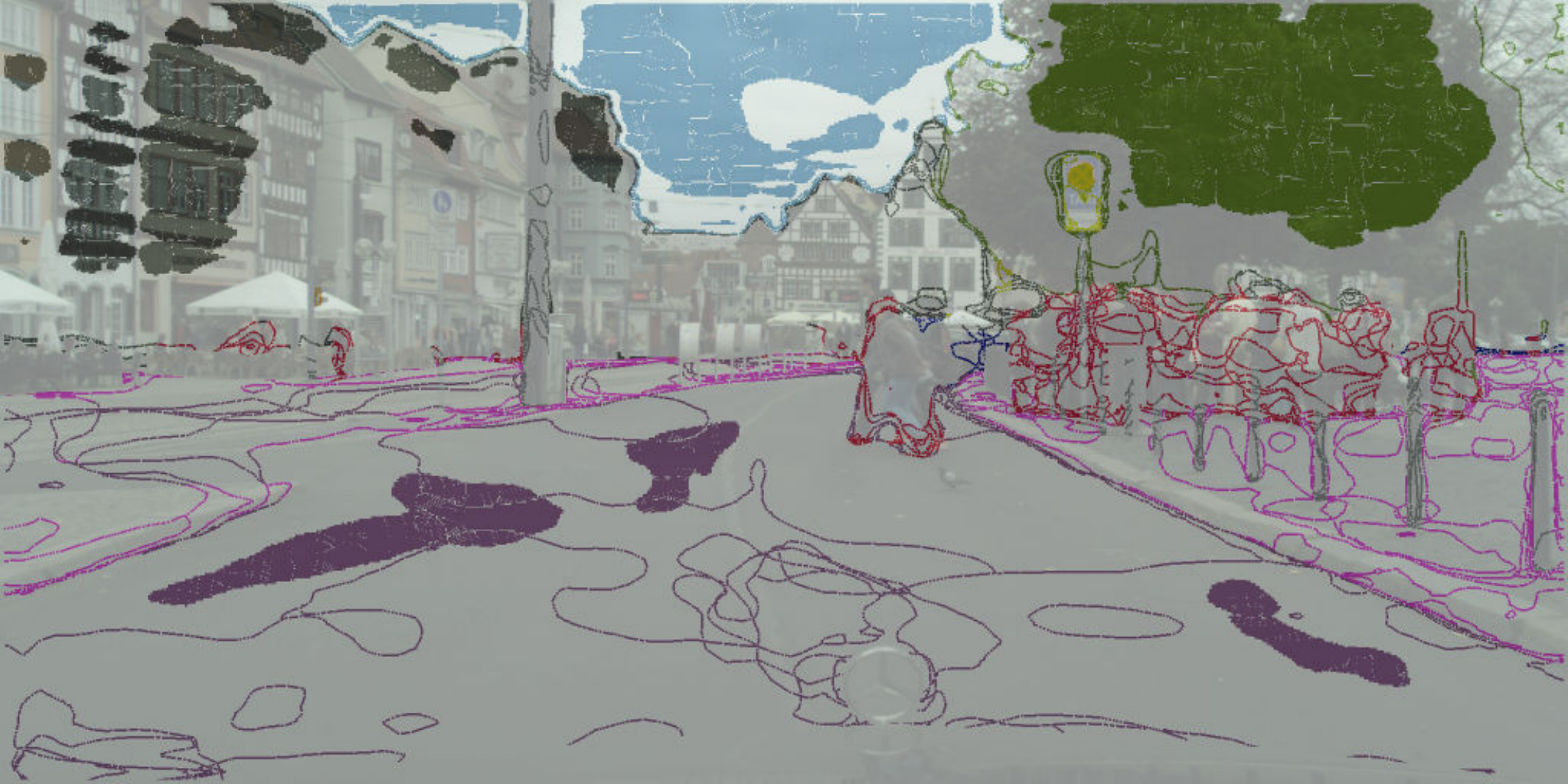}} \\ \vspace{-0.325cm}
    \subfloat{\includegraphics[width=0.249\linewidth]{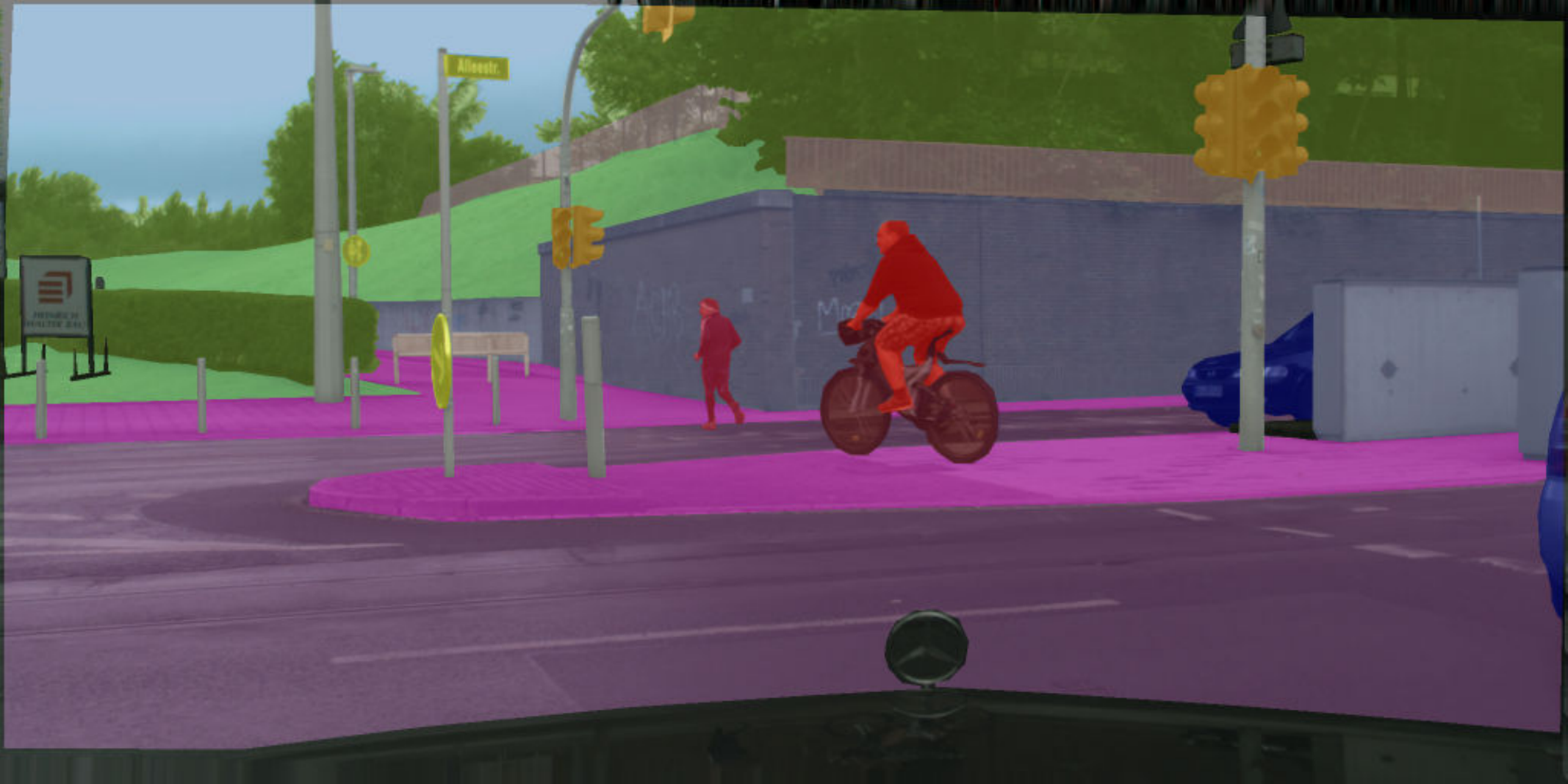}} \hfill
    \subfloat{\includegraphics[width=0.249\linewidth]{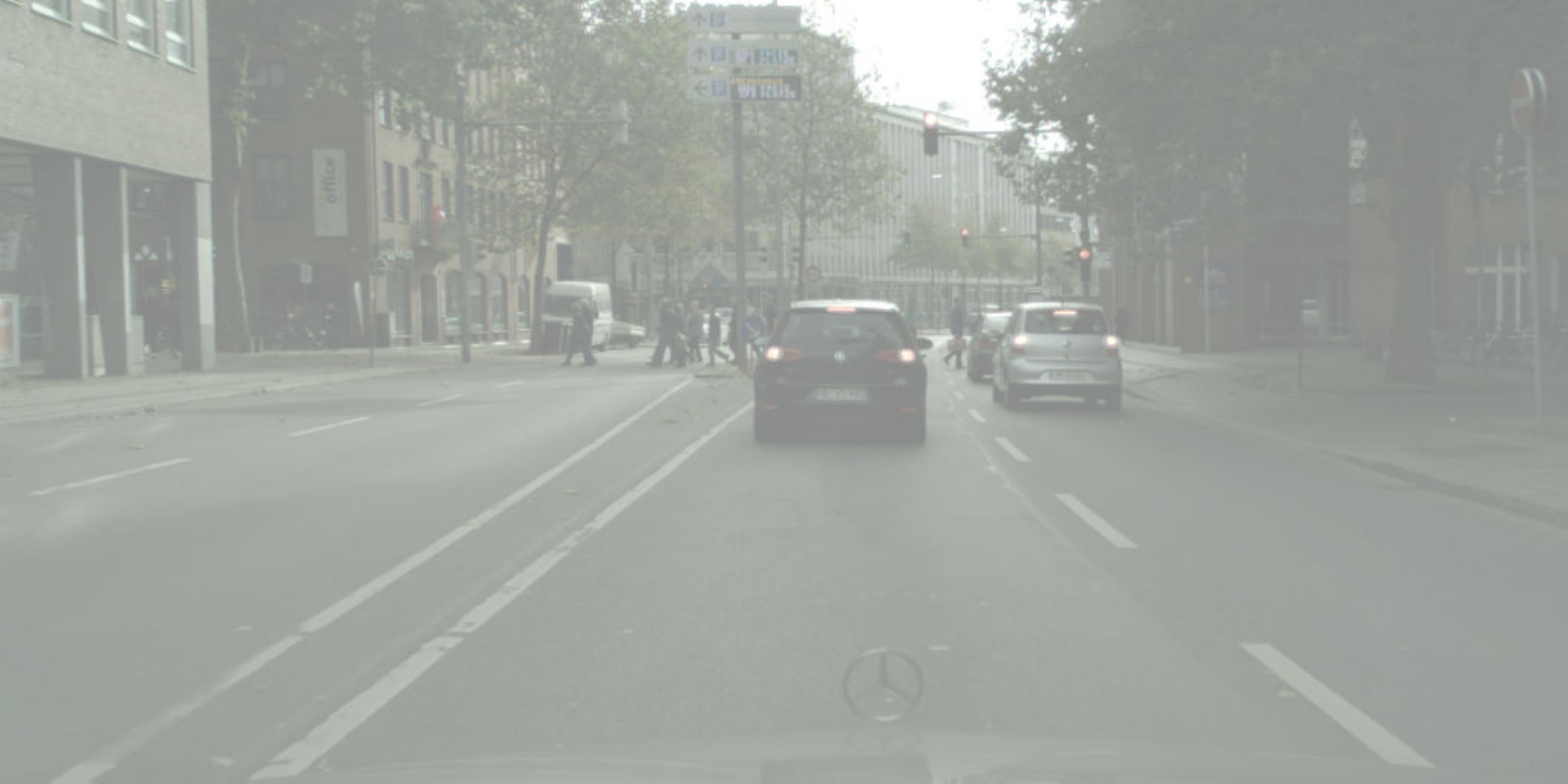}} \hfill
    \subfloat{\includegraphics[width=0.249\linewidth]{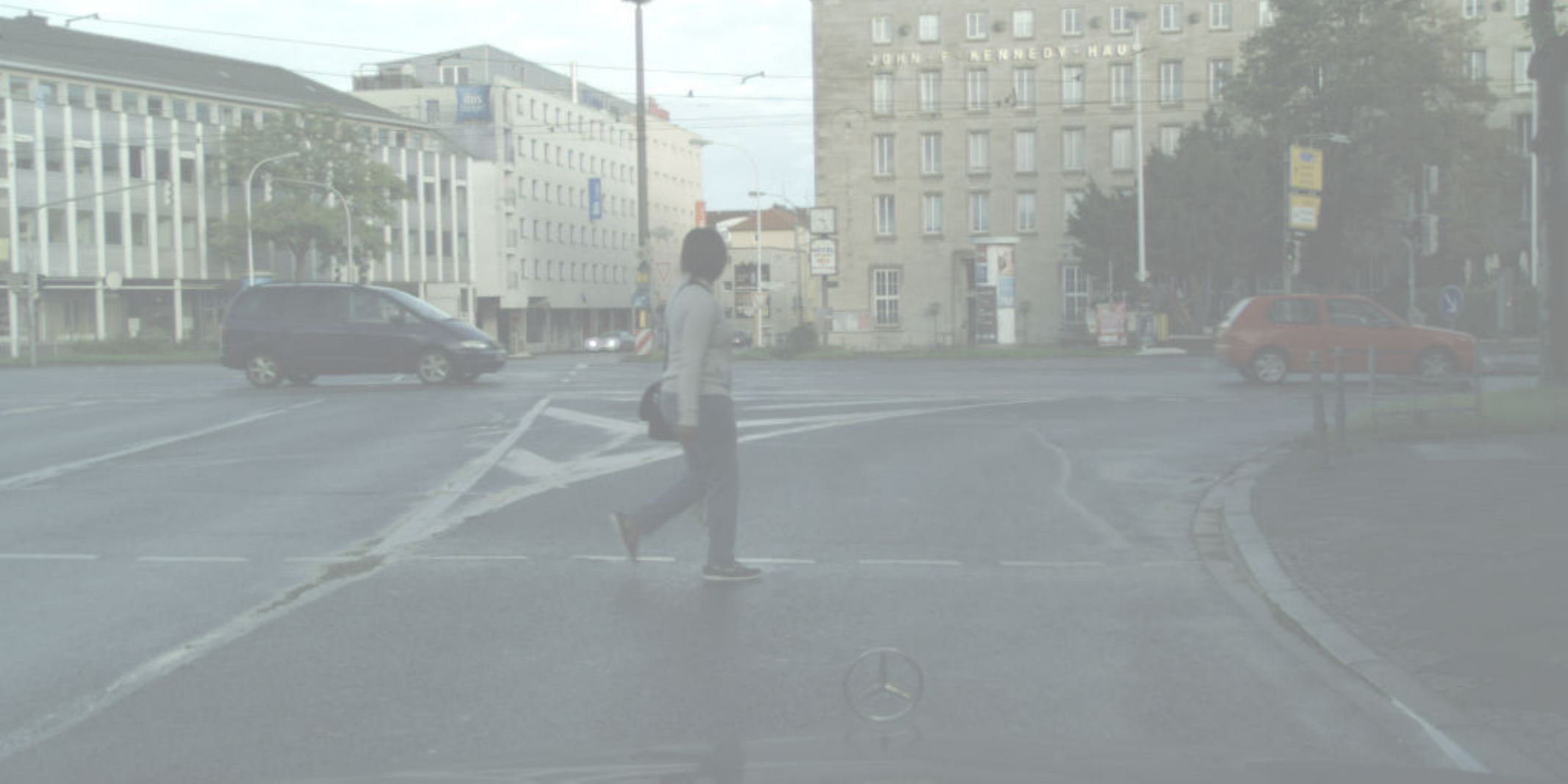}} \hfill
    \subfloat{\includegraphics[width=0.249\linewidth]{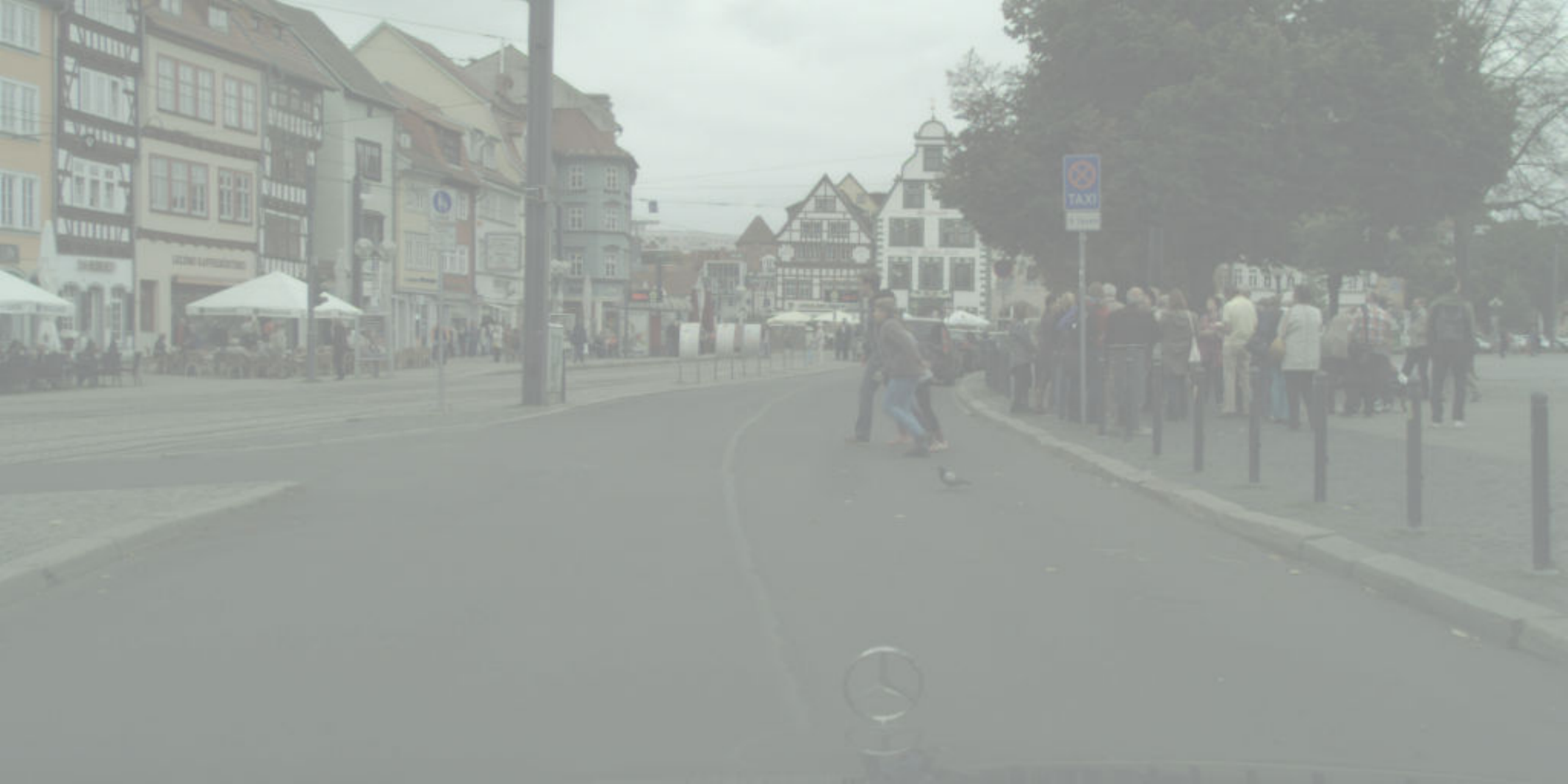}}
    \vspace{-0.2cm}
    \caption{Illustration of annotation forms for ADA. From top to bottom: original images, pixel/region-level (annotating 25\% of the pixels in each image), and image-level (annotating 25\% of the images). Colored regions indicate the areas to be annotated, while white regions represent the unlabeled areas.} 
    \label{fig:annotation_form}
    \vspace{-0.4cm}
\end{figure}

The images from $D_t^u$ are fed into the trained segmentation model from semi-supervised learning module, and prediction logits $p$ are obtained. Then, the images are ranked and selected based on the acquisition strategy. The acquisition strategy in active learning module is crucial for acquiring the informative samples. In SS-ADA, we adopt the image-based acquisition strategy, which is more suitable for manual labeling than pxiel and region-based ones, as shown in Figure~\ref{fig:annotation_form}. The value of the image is evaluated based on the model's uncertainty. We use classic metrics such as entropy and confidence to assess the uncertainty of the model's predictions. The image-level entropy $Ent(p)$ and confidence $Con(p)$ are defined as:  
\begin{equation}
    \centering
    Ent(p) = -\frac{1}{N}\sum_{i=1}^{N}\sum_{j=1}^{C} p^{(i,j)} \log p^{(i,j)}
    \label{eq:acquire_entropy}
\end{equation}

\vspace{-0.3cm}
\begin{equation}
    \centering
    Con(p) = \frac{1}{N}\sum_{i}^{N}\underset{\text{dim}=C}{\max}(p)^{i}
    \label{eq:acquire_confidence}
\end{equation}
where $p$ presents the prediction logits after applying softmax, with the shape of (C, H, W). C is the number of classes, H and W are the height and width of the image, and N is the number of pixel. The image-level entropy and confidence are calculated by averaging the pixel-level entropy and confidence of the predictions. If $Ent(p)$ is used as the acquisition strategy, the images with the highest entropy are selected. If $Con(p)$ is used, the images with the lowest confidence are selected. The selected images are annotated by humans and added to $D_{t}^l$.  

\subsection{IoU-based Class Weighting}

\begin{figure}
    \centering
    \captionsetup[subfloat]{font=scriptsize,labelfont=scriptsize,labelformat=empty}
    \subfloat{\includegraphics[width=0.8\linewidth]{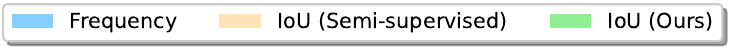}}\\ \vspace{-0.35cm}
    \subfloat{\includegraphics[width=1.0\linewidth]{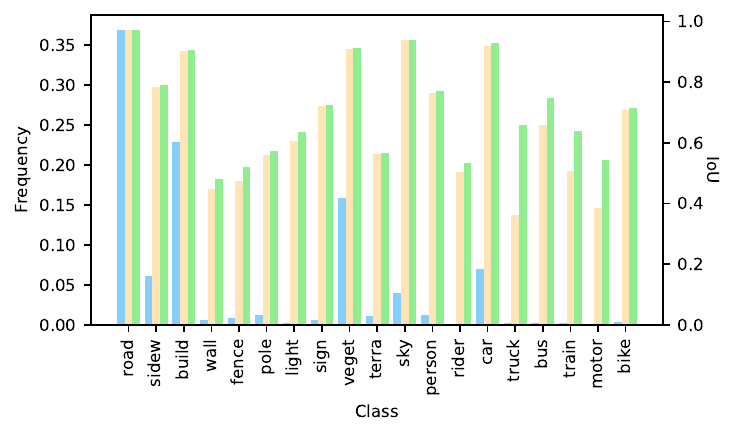}} \\ \vspace{-0.1cm}
    \caption{The class imbalance problem in GTA5-to-Cityscapes.}
    \label{fig:cls_imbalance}
    \vspace{-0.4cm}
\end{figure}

To alleviate the class imbalance problem in the target domain, we design a class weighting strategy based on the IoU of labeled data in the target domain $D_t^l$, obtained from the active learning module. It adjusts the weights of each class in the loss function during training, assigning higher weights to classes with lower IoU and vice versa. 
Previous methods typically calculate weights based on class frequency: 
\begin{equation}
    F_{i} = \frac{N_i}{\sum_{j=1}^{C}N_j}
    \label{eq:fre_weight}
\end{equation}
where $N_i$ is the number of pixels of class $i$, and $C$ is the number of classes. We calculate the frequency of each class using the ground truth of the Cityscapes validation set, and the IoU of the semi-supervised learning method UniMatch \cite{unimatch} with 25\% target labeled data in the GTA5-to-Cityscapes task, as shown in Figure~\ref{fig:cls_imbalance}. We observed that some classes, such as traffic signs and people, despite having low class frequencies, achieve relatively good class IoU (above 72\%). On the other hand, classes like trucks and trains have both low class frequencies and low IoU (below 60\%), thus requiring higher weights.  This observation motivated us to calculate class weights based on IoU. 

Specifically, after the active learning module is triggered at specified epochs and $D_{t}^l$ is updated, the semantic segmentation model makes predictions on the images in $D_{t}^l$ and calculates the IoU for each class. Let $x_t^l$, $\hat{y}_t^l$ and $y_t^l$ denote the image from target labeled domain, its prediction and ground truth, respectively. The IoU of each class, i.e., $IoU_i$ is calculated as:  

\begin{small}
    \begin{equation}
        \centering
        IoU_i = \frac{\sum I[y_t^l=i, \hat{y}_t^l=i]}{\sum I[y_t^l=i]+\sum I[\hat{y}_t^l=i]-\sum I[y_t^l=i, \hat{y}_t^l=i]}
        \label{eq:iou_cal}
    \end{equation}
\end{small}%
where $i$ is the class index, $I$ is the indicator function, which equals 1 when the condition is satisfied, and 0 otherwise. The loss weight for class $i$, denoted as $w_i$, is updated as:

\begin{equation}
    \centering
    w_i = (1-IoU_i) \times (u-1) + 1
\end{equation}
where $u$ is the upper bound of the class weight and is set as 2 in our experiments. The range of the class weight is $[1, u]$ and $w_i$ is incorporated into the loss calculation for $L_s^l$, $L_t^l$, $L_t^{fp}$, $L_t^{s1}$, and $L_t^{s2}$. Take the loss $L_s^l$ as an example, the loss function with the IoU-based class weighting is:
\begin{equation}
    \centering
    L_s^l(p_s^l, y_s^l) = -\frac{1}{N} \sum_{i=1}^{N} \sum_{j=1}^{C} w_j y_s^{l,(i,j)} \log p_s^{l,(i,j)}
    \label{eq: loss_s_l_weight}
\end{equation}
where $i$ and $j$ are the pixel and channel indices, $N$ and $C$ are the number of pixels and classes, $w_j$ is the weight for class $j$.

\subsection{Training Process}
The pseudocode of the training process of SS-ADA is outlined  in Algorithm~\ref{alg:training}. The function ``$acquire(p, \frac{N_{s}}{N})$'' denotes the selection of $\frac{N_s}{N}$ samples for annotation based on the model prediction $p$, Equation~\eqref{eq:acquire_entropy}, and~\eqref{eq:acquire_confidence} in the active learning module. The function ``$human\_annotate(x_t^{select})$'' signifies the human annotation for the selected unlabeled images $x_t^{select}$. Lastly, the function ``$iou\_cal(\hat{y}_t^l, y_t^l)$'' involves calculating the class IoU using $\hat{y}_t^l$ and $y_t^l$ based on Equation~\eqref{eq:iou_cal}.

\begin{algorithm}[h]
    \caption{Pseudocode of the training process}\label{alg:training}
    \KwIn{Source labeled data: $D_s^l=(x_s^l, y_s^l)$, target unlabeled data: $D_t^u=x_t^u$, total training epoches: $T_{t}$, active learning trigger epoches: $T_{ac}=[T_{a1}, T_{a2}, \ldots, T_{aN}]$, sampling number: $N_{s}$, semantic segmentation model $S$
            }
    \KwOut{Trained model $S$}
    Uniformly Randomly annotate 1\% data of $D_t^u$ as $D_t^l$\;

    \For{{$T_i$} in {$T_{t}$}}{
        \eIf{$T_i$ in $T_{ac}$}{
            \# Make prediction on $D_t^u$:\\
            $p = S(D_t^u)$\;
            \# Select the informative samples for annotation:\\
            $x_t^{select} = acquire(p, \frac{N_{s}}{N})$\;
            $y_t^{select} = human\_annotate(x_t^{select})$\;
            \# Update $D_t^l$ with ($x_t^{select}$, $y_t^{select}$):\\
            $D_t^l= D_t^l \bigcup (x_t^{select}, y_t^{select})$\;
            \# Make prediction on $D_t^u$: \\
            $\hat{y}_t^l = S(D_t^l)$\;
            \# Calculate $IoU_i$ and $w_{i}$:\\
            $IoU_i = iou\_cal(\hat{y}_t^l, y_t^l)$\;
            $w_{i} = 1 - IoU_i \times (u-1) + 1$\;
        }
        {
            \# Calculate semi-supervised loss with $w_{i}$:\\
            $L_{semi} = L_s^l + L_t^l + \lambda (L_t^{fp} + L_t^{s1} + L_t^{s2})$;\\
            Update S by $L_{semi}$
        }
        
    }
    
\end{algorithm}
\vspace{-0.2cm}


%% file: tex/4_experiment.tex
\section{Experiment}
\subsection{Dataset}
\begin{figure}[!tbp]
    \centering
    \captionsetup[subfloat]{font=scriptsize,labelfont=scriptsize,labelformat=empty}
    \subfloat[front-view]{\includegraphics[width=0.25\linewidth]{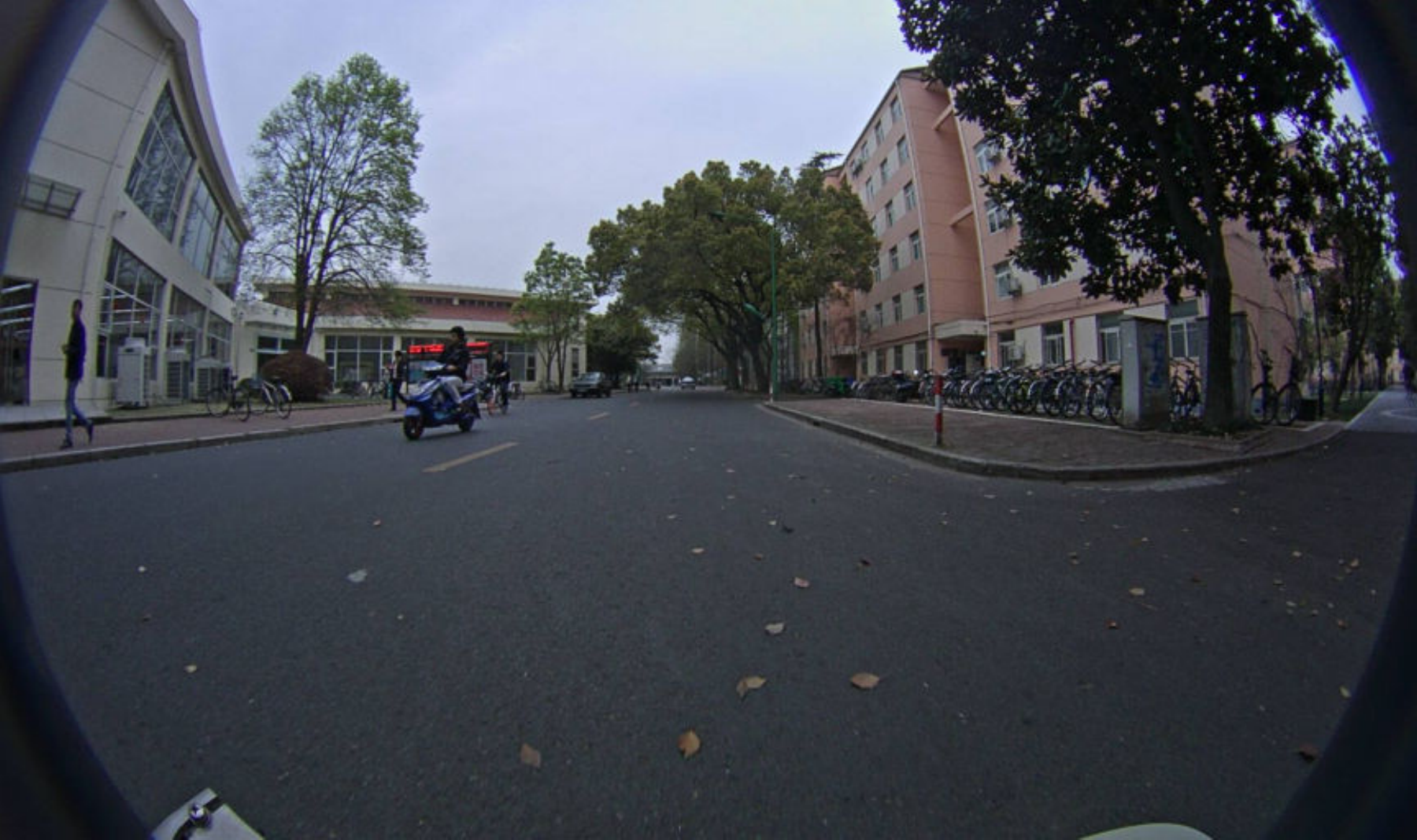}}\hfill
    \subfloat[rear-view]{\includegraphics[width=0.25\linewidth]{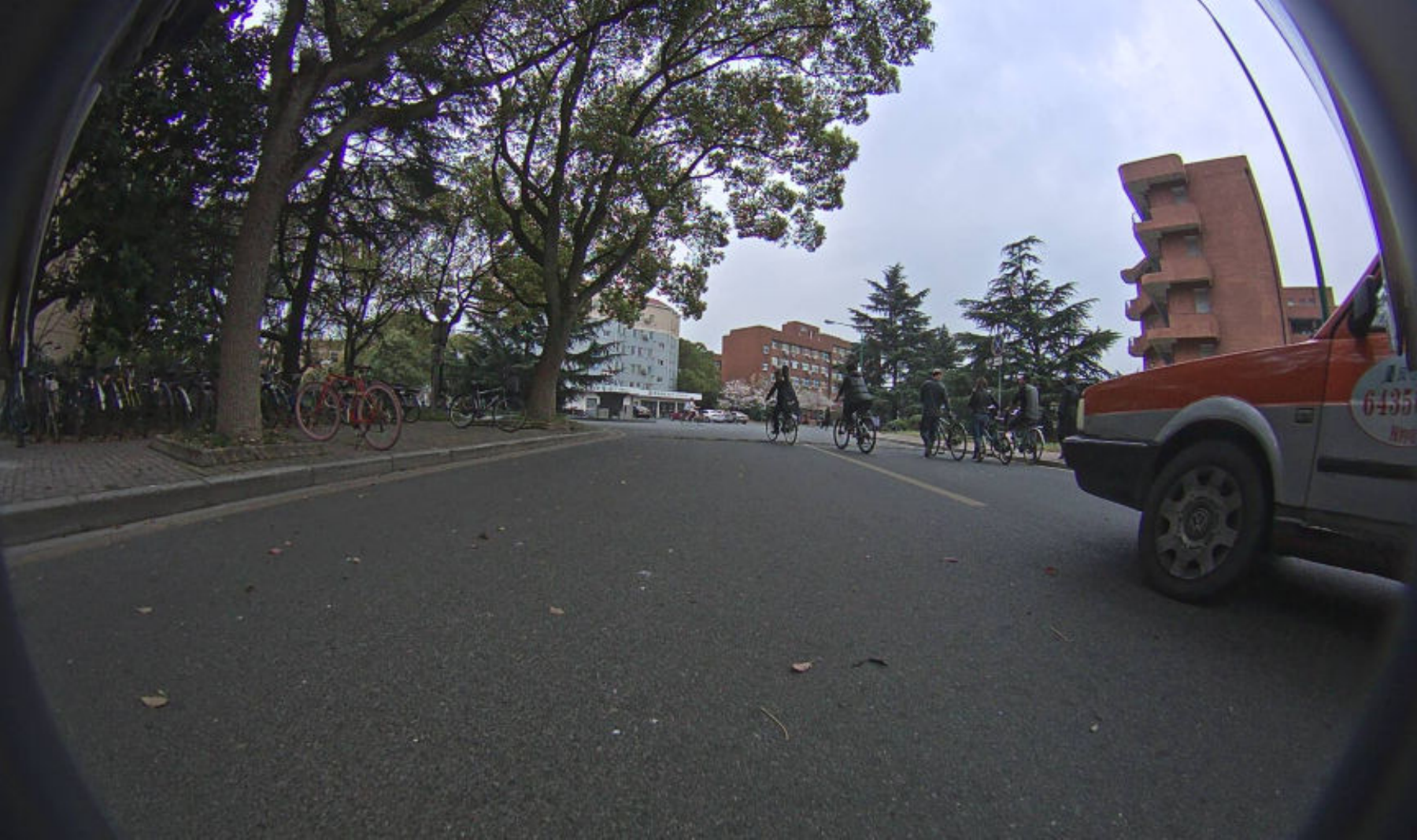}}\hfill
    \subfloat[left-view]{\includegraphics[width=0.25\linewidth]{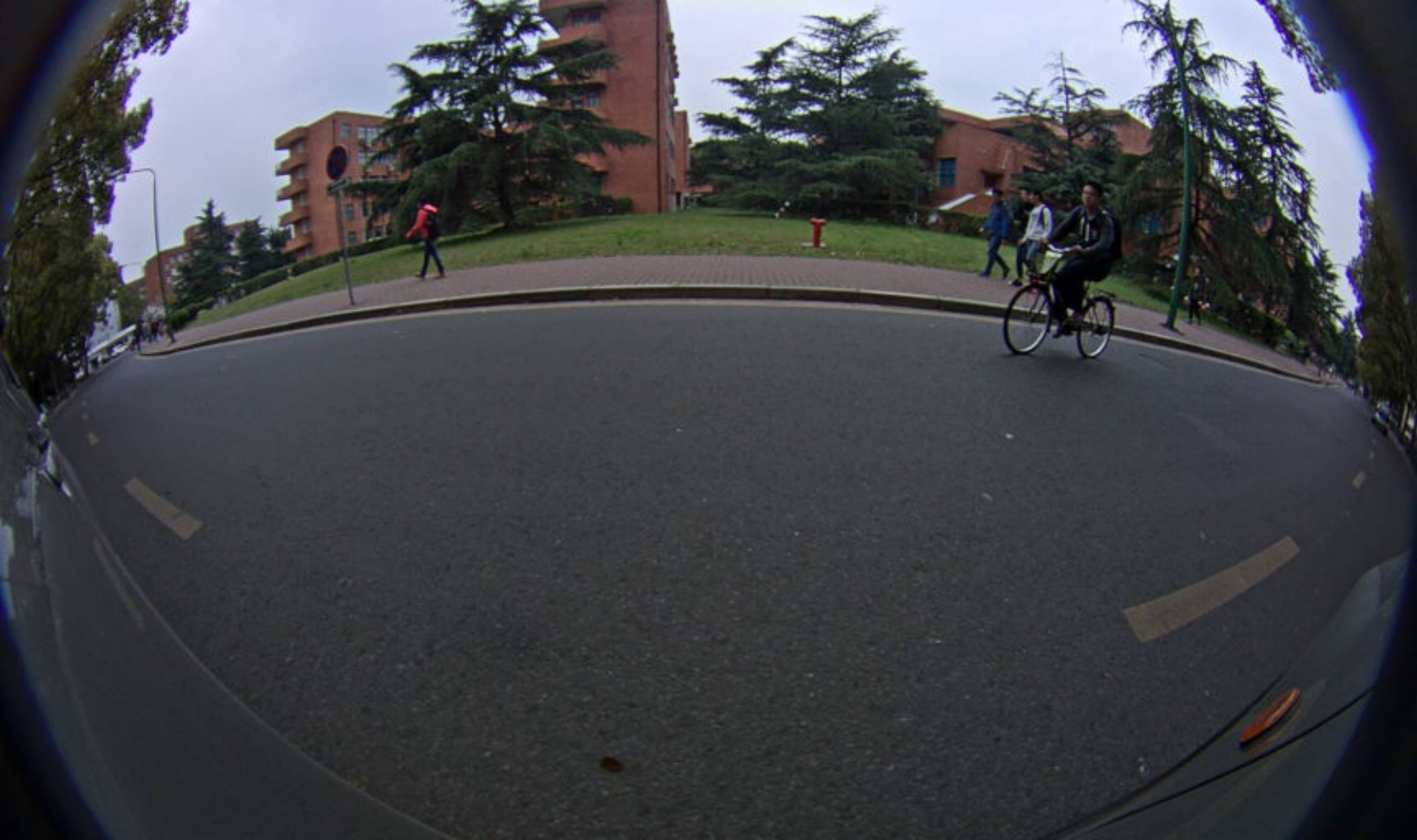}}\hfill
    \subfloat[right-view]{\includegraphics[width=0.25\linewidth]{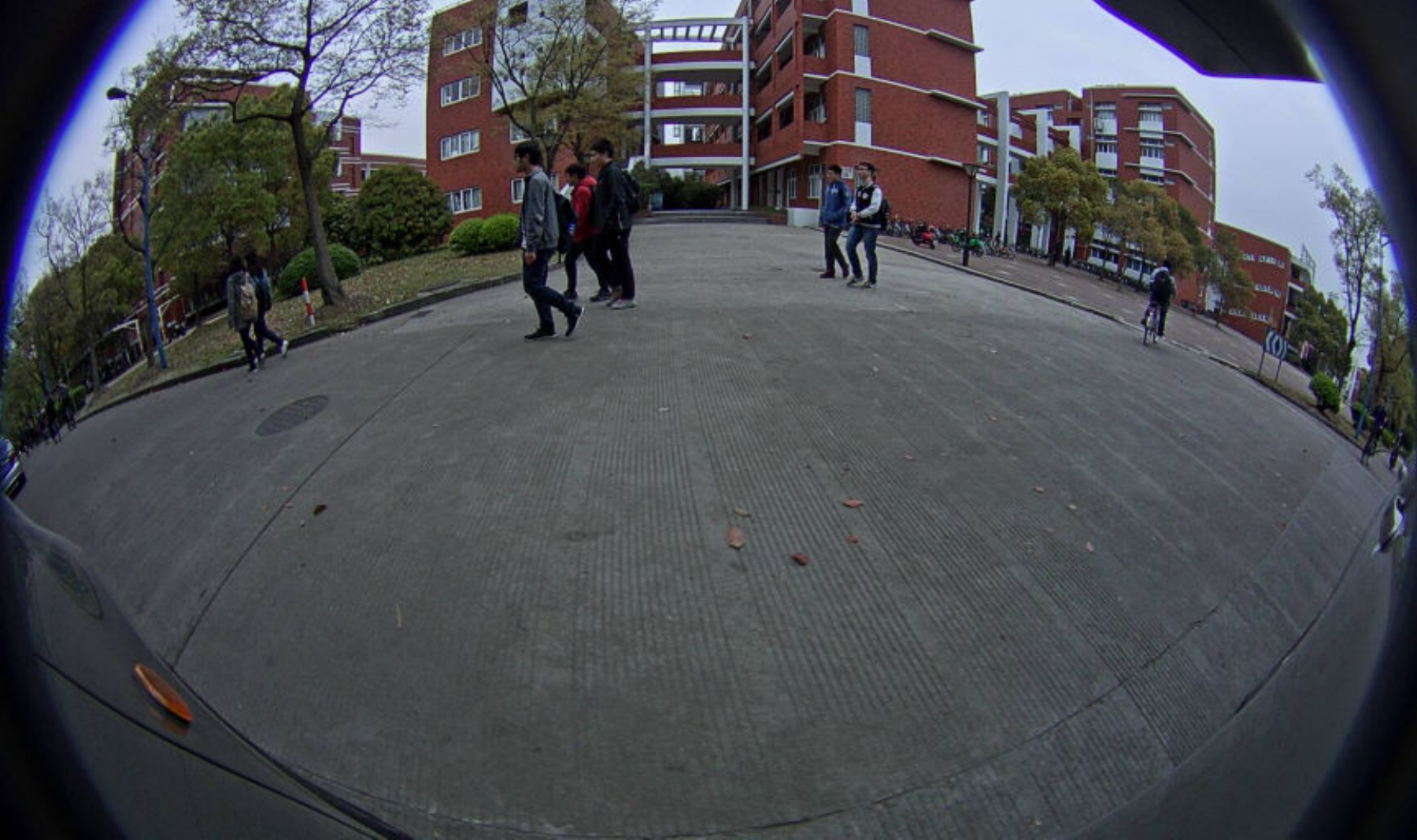}}
    \caption{Some examples of the FishEyeCampus dataset.}
    \label{fig:fisheye}
    \vspace{-0.4cm}
\end{figure}

We conduct synthetic-to-real and real-to-real domain adaptation experiments that include three public settings: GTA5-to-Cityscapes, SYNTHIA-to-Cityscapes, Cityscapes-to-ACDC and our customized setup: Cityscapes-to-FishEyeCampus. The characteristics of these datasets, including their type (simulated or real), the count of training and validation sets, resolution, and the number of semantic categories, are summarized in Table~\ref{tab:dataset}. GTA5 originates from the video game Grand Theft Auto V, while SYNTHIA is a product of a virtual city environment. Cityscapes consists of diverse urban street scenes captured in Germany and nearby regions. ACDC comprises 1000 foggy, 1006 nighttime, 1000 rainy and 1000 snowy images for semantic understanding in challenging visual conditions. These datasets are widely employed in domain adaptation studies, covering scenarios from simulation to real-world environments and from normal to adverse weather conditions. FishEyeCampus, our proprietary dataset, was collected on campus using four surround-view fisheye cameras. Selected samples from FishEyeCampus are shown in Figure~\ref{fig:fisheye}. This dataset allows us to further evaluate SS-ADA's effectiveness in real-world contexts, demonstrating its adaptability from pinhole camera images (Cityscapes) to fisheye camera images.

\begin{table}[!tbp]
    \centering
    \caption{The information of datasets used in the experiments.}
    \label{tab:dataset}
    \setlength{\tabcolsep}{5pt}
    \begin{tabular}{c|ccccc}
        \hline
        Dataset & Type & Train & Val  & Resolution & Classes \\ \hline
        GTA5 & Simulated & 24966 & - & 1914 $\times$ 1052 & 19 \\
        SYNTHIA & Simulated & 9400 & - & 1280 $\times$ 760 & 16 \\
        Cityscapes & Real & 2975 & 500 & 2048 $\times$ 1024 & 19 \\
        ACDC & Real & 1600 & 406 & 1920 $\times$ 1080 & 19 \\
        FishEyeCampus & Real & 450 & 150 & 864 $\times$ 512 & 17 \\ \hline
    \end{tabular}
    \vspace{-0.4cm}
\end{table}

\subsection{Evaluation Metric}
We employ the widely used evaluation metric, the mean Intersection over Union (mIoU), for semantic segmentation evaluation. It calculates the IoU for each class and then averages them. The mIoU is calculated as follows:
\begin{equation}
    mIoU = \frac{1}{C} \sum_{i=1}^{C} \frac{TP_i}{TP_i + FP_i + FN_i}
    \label{eq:mIoU}
\end{equation}
where $C$ is the number of classes, $TP_i$, $FP_i$ and $FN_i$ are the true positive, false positive and false negative of class $i$.

\subsection{Experimental Setup}

For real-time inference and deployment, we adopt BiSeNet\cite{bisenet} as the segmentation model unlike previous methods that use DeepLabV2 or DeeplabV3+\cite{deeplabv2}. In the semi-supervised learning module, the training parameters are set following \cite{u2pl,unimatch}. The model is trained using the SGD optimizer with a learning rate of 0.005, momentum of 0.9, and weight decay of 0.0001. A warm-up strategy for the learning rate is employed in the first 1000 iterations, and the batch size is set as 4. For the former three domain adaptation datasets, the images are cropped into (768, 768), while for Cityscapes-to-FishEyeCampus, the images are cropped into (512, 512) during the training process. The total training epochs are set as 240 for GTA5-to-Cityscapes and SYNTHIA-to-Cityscapes, 200 for Cityscapes-to-ACDC, and Cityscapes-to-FishEyeCampus. When reproducing U2PL and UniMatch using BiSeNet as the segmentation model, we adopt the same training parameters following\cite{u2pl,unimatch}. The active learning module is triggered at 20, 40, and 60 epochs for all datasets. The upper bound of the class weight $u$ is set to 2. The confidence threshold for pseudo-label generation is set to 0 and can be improved by existing self-training methods. To ensure a fair comparison, we also replace the segmentation model in D2ADA\cite{d2ada} and RIPU\cite{ripu} with BiSeNet and maintain the original training parameters in their respective papers. Our goal is to achieve the accuracy of supervised learning using fewer target labeled samples. We use the entire target domain dataset to perform supervised training with BiSeNet as the target accuracy. Additionally, we include the accuracy obtained from joint training with both source and target domain datasets. The experiments are conducted on a single NVIDIA RTX 3090 GPU. More experimental details can be found at \url{https://github.com/ywher/SS-ADA}.

\subsection{Quantitative Results and Analysis}

    


\subsubsection{GTA5-to-Cityscapes}
\begin{table*}[!htbp]
    \centering
    \setlength{\tabcolsep}{3pt}
    \renewcommand{\arraystretch}{1.05}
    \captionsetup{format=myformat}
    \caption{The performance of SS-ADA on GTA5-to-Cityscapes.} 
    \vspace{-0.1cm}
    \label{tab:gta5_comparison}
    \resizebox{\linewidth}{!}{
    \begin{tabular}{c|c|c|cccccccccccccccccccc}
        \hline
        Methods & Type & Ratio & mIoU & road & sw & build & wall & fence & pole & light & sign & vege & terrain & sky & person & rider & car & truck & bus & train & motor & bike \\ \hline
        Neutral\cite{neutral} & \multirow{2}{*}{UDA} & 0 & 49.5 & 90.9  & 52.9  & 86.1  & 38.5  & 24.5  & 41.5  & 46.7  & 42.1  & 86.7  & 37.0  & 86.4  & 63.3  & 24.2  & 88.1  & 36.8  & 42.4  & 2.3  & 19.0  & 30.4 \\
        TUFL\cite{TUFL} & & 0 & 55.4 & 89.8  & 54.9  & 87.1  & 47.0  & 34.0  & 41.8  & 53.0  & 58.6  & 86.6  & 38.4  & 92.3  & 57.7  & 28.3  & 87.3  & 51.0  & 50.2  & 9.3  & 26.7  & 58.3 \\ 
        \hline \hline
        \multirow{2}{*}{UniMatch\cite{unimatch}} & joint & 100 & \textbf{71.5} & \underline{97.2}  & \textbf{79.3}  & \textbf{90.6}  & 36.5  & \textbf{52.1}  & 56.7  & \textbf{64.2}  & \underline{72.1}  & \underline{91.1}  & \underline{59.0}  & {93.6}  & \underline{77.5}  & \underline{53.5}  & \textbf{93.4}  & \textbf{73.8}  & \textbf{79.8}  & \textbf{67.8}  & 49.6  & 71.2 \\
        & sup & 100 & {71.0} & \underline{97.2}  & \underline{79.0}  & \underline{90.5}  & 44.4  & 50.8  & \textbf{58.1}  & \underline{63.8}  & 71.9  & \textbf{91.2}  & 58.6  & 92.6  & \textbf{78.1}  & \textbf{57.1}  & \underline{93.1}  & 66.0  & 73.2  & 57.8  & \underline{53.6}  & \textbf{72.4} \\ \hline 
        \multirow{3}{*}{Ours (UniMatch)} & \multirow{3}{*}{ADA} & 5 & 66.8 & 96.4  & 75.0  & 89.2  & 43.7  & 45.1  & 53.3  & 58.2  & 68.8  & 90.7  & 55.4  & \textbf{93.8}  & 75.8  & 49.7  & 91.6  & 54.6  & 67.4  & 43.6  & 47.2  & 69.4 \\ 
        & & 12.5 & 70.8  & {97.0}  & 78.1  & \underline{90.5}  & \textbf{52.9}  & 48.9  & 55.8  & {62.8}  & 71.4  & 90.9  & \textbf{59.6}  & {93.4}  & 76.4  & 50.5  & 92.7  & \underline{66.6}  & \underline{74.6}  & \underline{65.6}  & 46.8  & 71.3 \\ 
        & & 25 & \underline{71.3}  & \textbf{97.3} & {78.9}  & 90.4  & \underline{48.1}  & \underline{52.0}  & \underline{57.1}  & {63.5}  & \textbf{72.4}  & \textbf{91.2}  & 56.5  & \underline{93.7}  & 77.0  & {53.4}  & {92.8}  & 65.9  & \underline{74.6}  & {64.0}  & \textbf{54.3}  & \underline{71.6} \\ 
        \hline \hline
        \multirow{2}{*}{U2PL\cite{u2pl}} & joint & 100 & \underline{70.5} & \textbf{97.6} & \textbf{82.1} & \underline{90.7} & 39.3 & \underline{53.4} & \textbf{58.0} & \textbf{64.2} & \textbf{74.0} & \underline{90.9} & \textbf{60.2} & \textbf{93.7} & 77.0 & 49.8 & \textbf{93.7} & 64.8 & \textbf{77.9} & 47.9 & \underline{51.2} & \textbf{73.6} \\
        & sup & 100 & {70.1} & \textbf{97.6} & \underline{81.4} & 90.4 & 45.9 & 50.9 & \underline{57.2} & \underline{64.0} & 72.4 & \textbf{91.1} & \underline{60.0} & {92.2} & 76.4 & \underline{52.9} & 92.7 & \underline{68.6} & 74.5 & \underline{54.9} & 38.6 & 70.6 \\
        \hline
        \multirow{3}{*}{Ours (U2PL)} & \multirow{3}{*}{ADA} & 5 & 59.6 & 88.6 & 49.7 & 88.1 & 44.6 & 40.0 & 52.3 & 53.3 & 62.3 & 90.3 & 53.1 & 87.8 & 72.4 & 39.3 & 90.0 & 44.2 & 56.4 & 10.7 & 41.5 & 67.0 \\
        & & 12.5 & {68.3} & 96.5 & 75.5 & 89.7 & \underline{47.1} & 47.7 & 55.3 & 60.6 & 68.1 & 90.6 & 55.3 & {92.1} & \underline{77.4} & 52.5 & 92.5 & {67.1} & 67.8 & 41.2 & 49.9 & 70.8 \\
        & & 25 & \textbf{72.4} & \underline{97.0} & {78.7} & \textbf{91.0} & \textbf{56.8} & \underline{54.3} & \textbf{58.0} & {63.5} & \textbf{72.9} & \textbf{91.1} & {57.1} & \underline{92.9} & \textbf{78.4} & \textbf{54.8} & \underline{93.4} & \textbf{69.5} & \underline{77.2} & \textbf{66.2} & \textbf{51.8} & \underline{71.7} \\
        \hline
    \end{tabular}}
    \caption*{Ours (UniMatch) and Ours (U2PL): SS-ADA with UniMatch and U2PL as the semi-supervised learning module, respectively. joint: joint training; sup: supervised learning; Ratio: the ratio of target labeled data. The highest and second-highest values in each semi-supervised series are presented in \textbf{bold} and \underline{underline}, respectively.}
    \vspace{-0.46cm}
\end{table*}

\begin{figure*}[!htbp]
    \centering
    \captionsetup[subfloat]{font=scriptsize,labelfont=scriptsize,labelformat=empty}
    \subfloat[]{\includegraphics[width=0.75\linewidth]{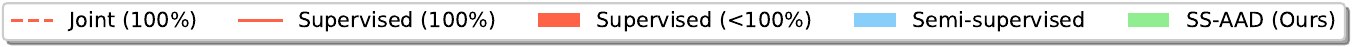}}\\ \vspace{-0.7cm}
    \subfloat[]{\label{fig:uni_gta5}\includegraphics[width=0.25\linewidth]{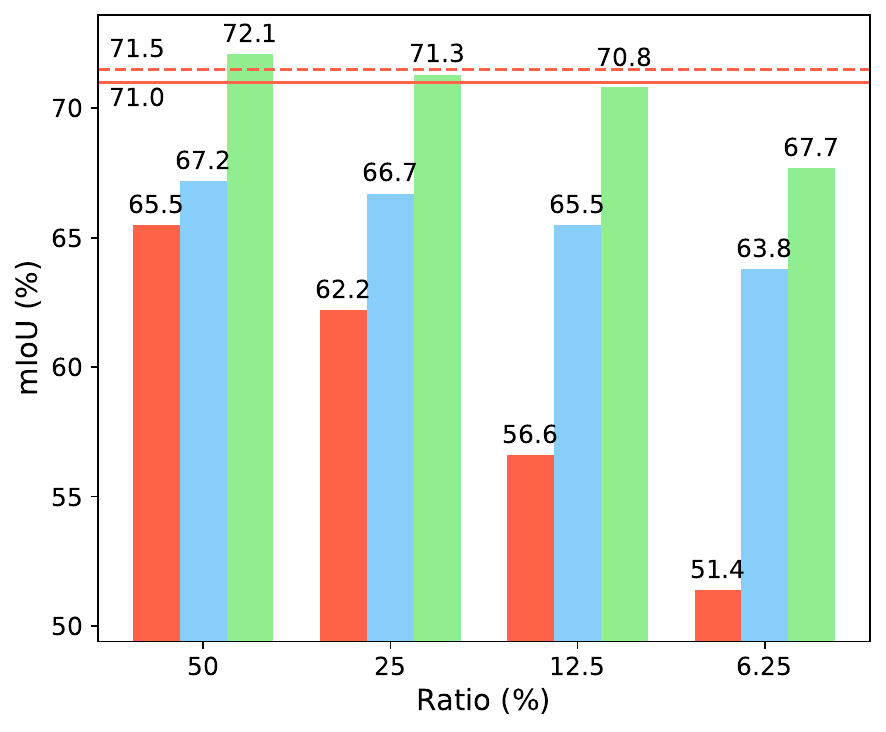}}
    \subfloat[]{\label{fig:uni_syn}\includegraphics[width=0.25\linewidth]{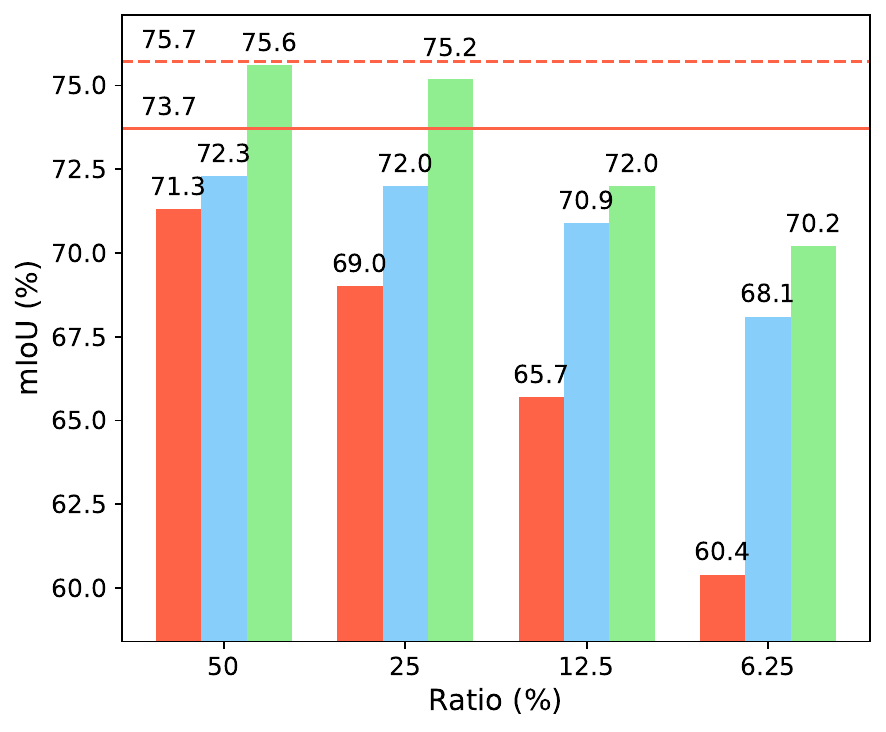}}
    \subfloat[]{\label{fig:uni_acdc}\includegraphics[width=0.25\linewidth]{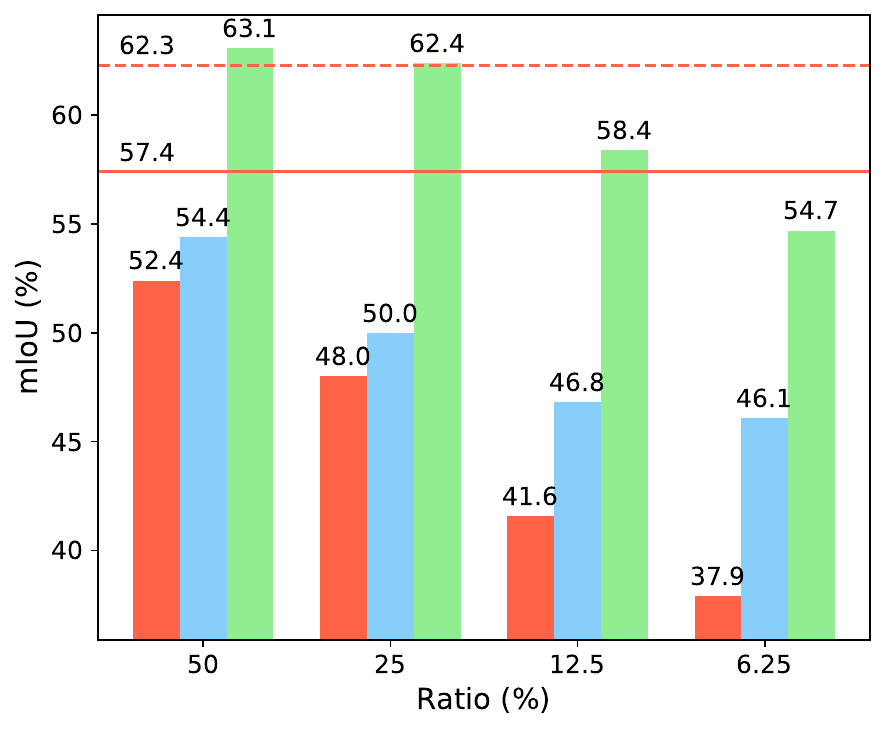}}
    \subfloat[]{\label{fig:uni_school}\includegraphics[width=0.25\linewidth]{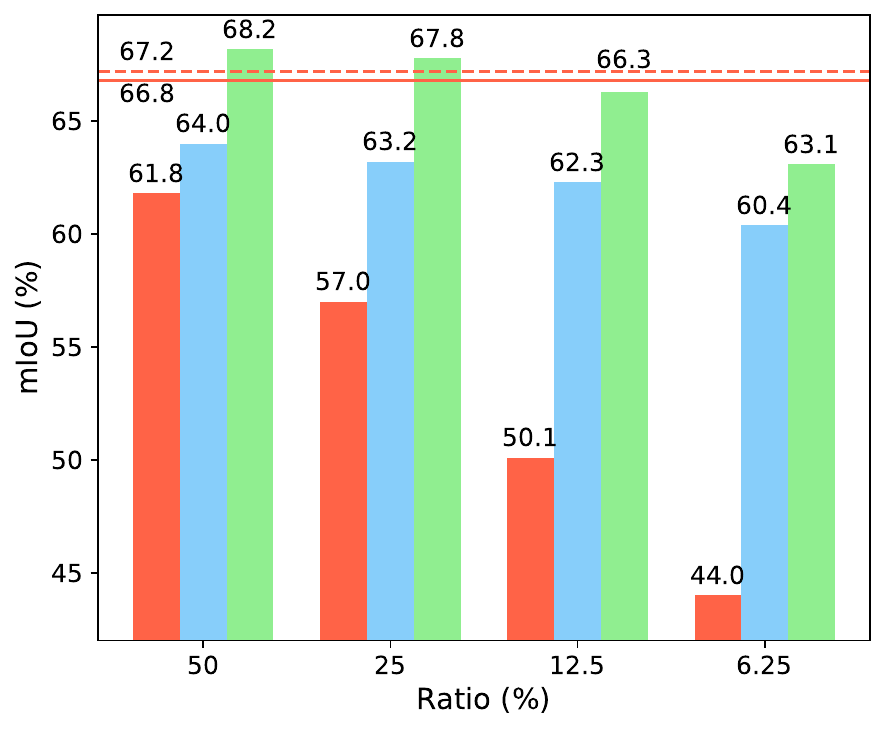}}\\ \vspace{-0.7cm}
    \subfloat[GTA5-to-Cityscapes]{\label{fig:u2_gta5}\includegraphics[width=0.25\linewidth]{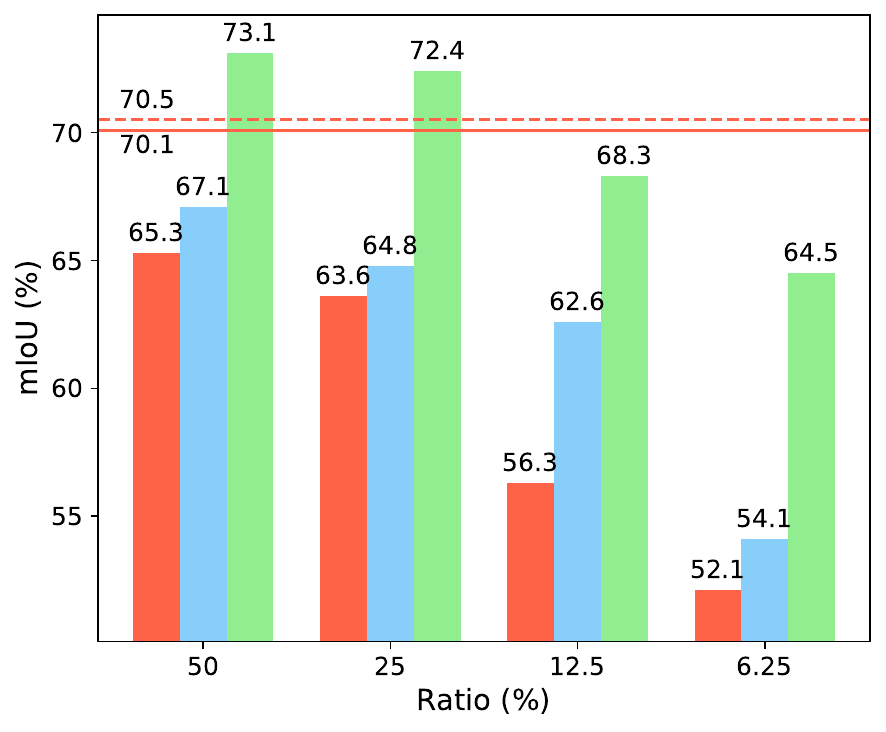}}
    \subfloat[SYNTHIA-to-Cityscapes]{\label{fig:u2_syn}\includegraphics[width=0.25\linewidth]{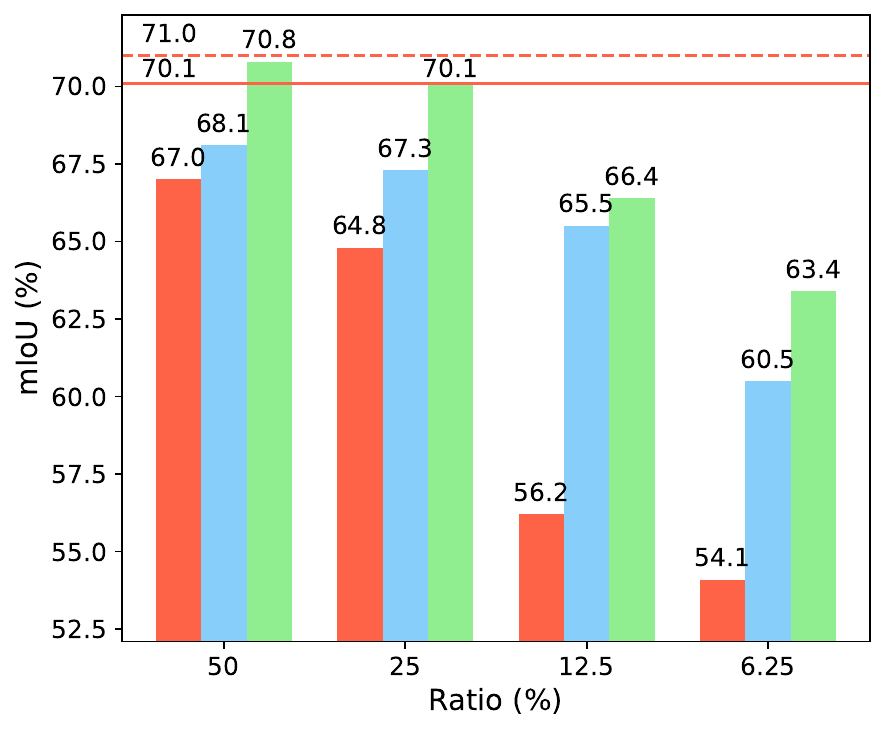}}
    \subfloat[Cityscapes-to-ACDC]{\label{fig:u2_acdc}\includegraphics[width=0.25\linewidth]{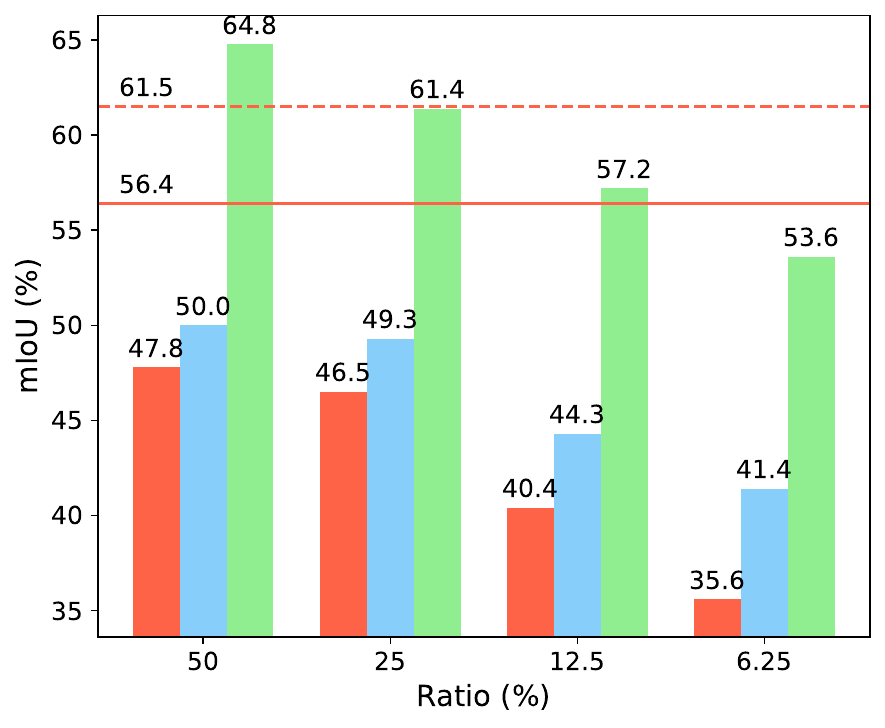}}
    \subfloat[Cityscapes-to-FishEyeCampus]{\label{fig:u2_school}\includegraphics[width=0.25\linewidth]{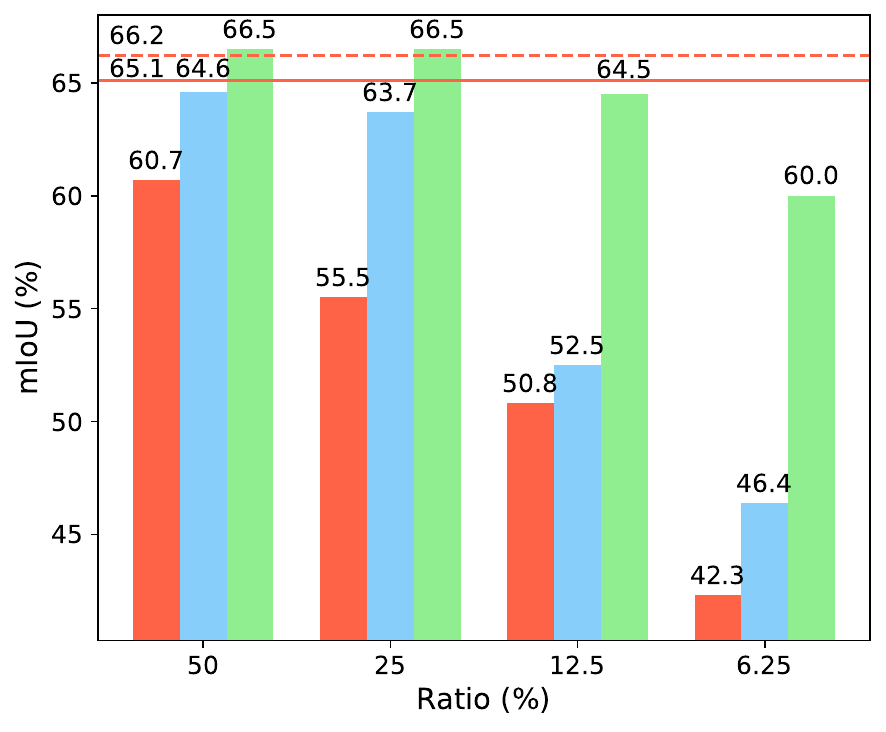}}
    \caption{The performance of joint training (100\% source and target labeled data), supervised learning (100\% and smaller proportions of target labeled data), semi-supervised learning (a portion of target labeled data and the remaining unlabeled data), and SS-ADA (100\% source labeled data and a portion of target labeled data) on four domain adaptation datasets. The semi-supervised learning methods used in the first and second rows are UniMatch and U2PL, respectively.}
    \label{fig:sup_semi_our_comparison_u2pl}
    \vspace{-0.25cm}
\end{figure*}

        


The quantitative results on GTA5-to-Cityscapes are shown in Table~\ref{tab:gta5_comparison}. When using UniMatch, the mIoU for supervised learning and joint training are 71.0\% and 71.5\%, respectively. With 25\% of labeled data in Cityscapes, SS-ADA achieves 71.3\% mIoU, surpassing that of supervised learning. To demonstrate the generalizability, we  replace the semi-supervised learning module with U2PL. The mIoU for supervised learning and joint training are 70.1\% and 70.5\%, respectively. With 25\% target labeled data, SS-ADA achieves an mIoU of 72.4\%, higher than both. 



In the first column of Figure~\ref{fig:sup_semi_our_comparison_u2pl}, we compare the performance of supervised learning, semi-supervised methods and SS-ADA under different ratios of target labeled data. Semi-supervised methods show an improvement in accuracy compared to supervised ones, and as the ratio of labeled data decreases, the absolute improvement becomes more significant, highlighting the benefits of leveraging unlabeled data. However, these methods still fall short of the performance achieved by supervised learning with 100\% labeled data. 
SS-ADA achieves further improvements in accuracy compared to semi-supervised methods. This is made through the use of an acquisition strategy and the inclusion of labeled source domain data. Remarkably, SS-ADA reaches the accuracy level of supervised learning while using only 25\% of the target labeled data, significantly reducing the annotation requirements and associated costs.

\subsubsection{SYNTHIA-to-Cityscapes}

\begin{table*}[!ht]
    \centering
    \setlength{\tabcolsep}{3pt}
    \renewcommand{\arraystretch}{1.05}
    \captionsetup{format=myformat}
    \caption{The performance of SS-ADA on SYNTHIA-to-Cityscapes.} 
    \label{tab:synthia_comparison}
    \resizebox{\linewidth}{!}{
    \begin{tabular}{c|c|c|ccccccccccccccccc}
        \hline
        Methods & Type & Ratio & mIoU & road & sw & build & wall & fence & pole & light & sign & vege & sky & person & rider & car & bus & motor & bike \\ \hline
        Neutral\cite{neutral} & \multirow{2}{*}{UDA} & 0 & 44.9 & 85.7  & 46.6  & 81.6  & 10.7  & 0.7  & 38.1  & 23.9  & 37.2  & 86.1  & 88.1  & 54.7  & 19.5  & 74.1  & 25.9  & 9.5  & 36.5 \\
        TUFL\cite{TUFL} & & 0 & 52.2 & 87.3  & 53.4  & 82.9  & 20.5  & 0.4  & 45.6  & 45.5  & 58.7  & 88.6  & 91.7  & 66.4  & 25.3  & 79.2  & 29.3  & 13.5  & 46.3 \\ 
        \hline \hline
        \multirow{2}{*}{UniMatch\cite{unimatch}} & joint 
        & 100 & \textbf{75.7} & \textbf{97.5}  & \textbf{82.1}  & \underline{91.2}  & \underline{52.4}  & \textbf{53.0}  & \textbf{60.7}  & \textbf{66.3}  & \textbf{75.3}  & \underline{92.3}  & \underline{94.1}  & \textbf{79.9}  & \textbf{57.5}  & \textbf{94.4}  & \textbf{82.1}  & \textbf{57.9}  & \textbf{74.5} \\ 
        & sup & 100 & {73.7} & \textbf{97.5}  & \textbf{82.1}  & {90.9}  & 44.3  & 49.0  & {59.1}  & 64.2  & {73.7}  & {92.2}  & {94.0}  & 77.7  & 55.5  & {93.8}  & \underline{79.6}  & 54.1  & 72.8 \\ \hline
        \multirow{3}{*}{Ours (UniMatch)} & \multirow{3}{*}{ADA} & 5 & 70.4 & 96.7  & 76.3  & 89.7  & 45.2  & 41.8  & 55.9  & 58.3  & 68.3  & 91.5  & {93.4}  & 77.2  & 49.8  & 92.6  & 68.8  & 49.9  & 71.0 \\
        & & 12.5 & {72.0}  & \underline{97.1}  & \underline{79.4}  & {90.2}  & {49.8}  & \underline{49.8}  & 56.9  & 58.2  & {72.2}  & {91.6}  & {93.4}  & {78.1}  & {53.3}  & 92.8  & 69.1  & 48.4  & {72.1} \\
        & & 25 & \underline{75.2}  & \textbf{97.5}  & \textbf{82.1}  & \textbf{91.3}  & \textbf{55.6}  & \textbf{53.0}  & \underline{60.4} & \underline{65.0}  & \underline{75.0}  & \textbf{92.4}  & \textbf{94.3}  & \underline{79.4}  & \underline{56.2}  & \underline{94.2}  & 76.3  & \underline{56.6}  & \underline{73.9} \\
        \hline \hline
        \multirow{2}{*}{U2PL\cite{u2pl}} &  joint & 100 
        & \textbf{71.0}  & \textbf{97.5}  & \textbf{81.7}  & \textbf{90.0}  & 36.9  & \textbf{50.9}  & \underline{56.8}  & 59.9  & \underline{71.7}  & \textbf{91.6}  & \underline{93.1}  & \textbf{76.5}  & 43.5  & \textbf{93.6}  & \textbf{75.4}  & \underline{45.2}  & \textbf{72.1} \\
        &  sup & 100 & \underline{70.1} & \underline{96.6}  & \underline{76.6}  & \underline{89.9}  & \textbf{39.5}  & \underline{49.2}  & 55.2  & \underline{61.5}  & \underline{71.7}  & \underline{91.5}  & \textbf{93.2}  & \underline{76.1}  & \textbf{51.4}  & \underline{93.2}  & 70.5  & 33.9  & \underline{71.2} \\ \hline
        \multirow{3}{*}{Ours (U2PL)} & \multirow{3}{*}{ADA} & 5 & 59.7  & 84.5  & 51.0  & 82.1  & 23.7  & 26.8  & 50.0  & 52.4  & 60.2  & 90.7  & 91.6  & 69.9  & 39.8  & 88.9  & 54.2  & 28.4  & 60.7 \\
        & & 12.5 & 66.4  & 91.0  & 62.0  & 86.7  & 38.9  & 33.4  & 53.6  & 58.9  & 69.0  & 91.0  & 92.5  & 73.9  & 44.6  & 92.3  & 69.3  & 37.3  & 67.2 \\
        & & 25 & \underline{70.1} & 95.4  & 72.0  & 89.4  & \underline{37.8}  & 41.1  & \textbf{58.0}  & \textbf{62.3}  & \textbf{72.0}  & \textbf{91.6}  & 92.9  & \underline{76.1}  & \underline{47.9}  & {93.0}  & \underline{71.0}  & \textbf{50.0}  & 70.6 \\

        \hline
    \end{tabular}}
    \vspace{-0.45cm}
\end{table*}

The experimental results on SYNTHIA-to-Cityscapes are shown in Table~\ref{tab:synthia_comparison}. Similarly, SS-ADA achieves the accuracy of supervised learning using 25\% of target labeled data but falls short of reaching the performance level of joint training. One possible reason for this discrepancy is that SYNTHIA data has higher quality compared to GTA5, and its distribution is closer to that of Cityscapes. Therefore, joint training tends to yield better performance. From the second column of Figure~\ref{fig:sup_semi_our_comparison_u2pl}, it's evident that SS-ADA consistently improves the mIoU compared to semi-supervised learning methods across different ratios. 



\subsubsection{Cityscapes-to-ACDC}


\begin{table*}[!htbp]
    \centering
    \setlength{\tabcolsep}{3pt}
    \renewcommand{\arraystretch}{1.05}
    \captionsetup{format=myformat}
    \caption{The performance of SS-ADA on Cityscapes-to-ACDC.} 
    \label{tab:acdc_comparison}
    \resizebox{\linewidth}{!}{
    \begin{tabular}{c|c|c|cccccccccccccccccccc}
        \hline
        Methods & Type & Ratio & mIoU & road & sw & build & wall & fence & pole & light & sign & vege & terrain & sky & person & rider & car & truck & bus & train & motor & bike \\ \hline
        Neutral\cite{neutral} & \multirow{2}{*}{UDA} & 0 & 40.3 & 81.9  & 47.2  & 66.0  & 18.6  & 16.3  & 28.1  & 50.3  & 41.8  & 64.8  & 32.1  & 79.5  & 33.0  & 9.5  & 69.2  & 36.8  & 24.9  & 33.3  & 16.5  & 16.1 \\
        TUFL\cite{TUFL} & & 0 & 43.6 & 79.1  & 45.1  & 63.8  & 22.9  & 21.6  & 35.2  & 49.4  & 45.0  & 73.1  & 34.4  & 81.1  & 38.0  & 14.6  & 74.0  & 40.2  & 38.5  & 34.9  & 17.0  & 19.9 \\ 
        \hline \hline
        \multirow{2}{*}{UniMatch\cite{unimatch}} & joint & 100 & \underline{62.3}  & \textbf{92.7}  & \textbf{69.9}  & \textbf{83.7}  & \textbf{48.2}  & \textbf{43.4}  & \textbf{55.9}  & \textbf{70.7}  & \underline{56.7}  & \textbf{85.0}  & \underline{42.7}  & \textbf{95.0}  & 47.9  & \underline{28.5}  & \textbf{82.0}  & \textbf{46.4}  & \textbf{80.6}  & \underline{85.4}  & \underline{27.2}  & \underline{41.7} \\
        & sup & 100 & {57.4}  & 91.5  & 67.0  & 82.1  & 46.9  & 39.9  & \underline{56.1}  & 68.9  & 55.2  & 83.9  & 40.8  & 94.3  & 43.2  & 7.2  & 78.7  & 33.5  & \underline{76.3}  & 69.9  & 15.5  & 39.7 \\ \hline
        
        \multirow{3}{*}{Ours (UniMatch)} & \multirow{3}{*}{ADA} & 5 & 54.6 & 87.9  & 58.3  & 79.5  & 33.5  & 32.8  & 51.7  & 63.6  & 55.2  & 81.5  & 36.7  & 93.9  & 50.0  & {27.4}  & 77.1  & 32.1  & 59.3  & 73.5  & 18.2  & 25.6 \\
        & & 12.5 & 58.4  & {90.6}  & {64.9}  & 80.8  & 38.0  & 37.9  & 52.2  & 65.5  & 55.4  & 83.6  & 39.2  & {94.5}  & \underline{53.8}  & 23.9  & 78.6  & 35.0  & {69.4}  & 79.6  & 25.5  & 41.3 \\
        & & 25 & \textbf{62.4}  & \underline{91.8}  & \underline{67.8}  & \underline{83.0}  & \underline{48.0}  & \underline{42.5}  & 54.4  & \underline{69.2}  & \textbf{59.2}  & \underline{84.3}  & \textbf{43.5}  & \underline{94.6}  & \textbf{56.1}  & \textbf{33.1}  & \underline{81.3}  & \underline{42.0}  & 71.8  & \textbf{86.0}  & \textbf{27.8}  & \textbf{49.2} \\
        \hline \hline

        \multirow{2}{*}{U2PL\cite{u2pl}} &  joint & 100 & \textbf{61.5} & \textbf{93.0}  & \textbf{72.9}  & \underline{82.2}  & \textbf{50.2}  & \underline{38.9}  & 51.9  & 66.1  & \underline{55.8}  & 82.9  & \textbf{43.7}  & 93.6  & \underline{53.0}  & \underline{24.2}  & \textbf{81.2}  & \textbf{56.8}  & \textbf{83.9}  & \textbf{81.8}  & 27.2  & 28.8 \\ 
        &  sup & 100 & {56.4} & \underline{92.3}  & \underline{71.7}  & 81.1  & 45.2  & 36.9  & 53.6  & 66.4  & 51.0  & \underline{83.8}  & \underline{41.8}  & \underline{94.5}  & 40.0  & 0.8  & {78.0}  & 45.0  & \underline{77.5}  & 67.9  & 17.3  & 26.4 \\ \hline
        \multirow{3}{*}{Ours (U2PL)} & \multirow{3}{*}{ADA} & 5 & 51.8  & 87.0  & 57.6  & 77.2  & 32.1  & 33.9  & 53.9  & 66.6  & 50.9  & 79.6  & 39.0  & 91.3  & 39.6  & 12.0  & 67.5  & 40.0  & 51.2  & 64.0  & 16.7  & 24.8 \\
        & & 12.5 & 57.2  & 87.8  & 59.0  & 80.4  & {38.3}  & 34.2  & \underline{54.6}  & \underline{67.0}  & 55.5  & 82.5  & 36.3  & \textbf{94.6}  & 44.9  & 20.9  & 76.5  & 40.9  & 69.0  & {80.2}  & \textbf{29.1}  & \underline{34.8} \\
        & & 25 & \underline{61.4} & {91.6}  & {69.1}  & \textbf{82.6}  & \textbf{48.2}  & \textbf{39.2}  & \textbf{56.8}  & \textbf{69.8}  & \textbf{59.1}  & \textbf{84.4}  & \underline{41.8}  & {94.3}  & \textbf{56.7}  & \textbf{26.5}  & \underline{80.2}  & \underline{44.5}  & 68.5  & \underline{80.6}  & \underline{27.3}  & \textbf{45.0} \\
        \hline
    \end{tabular}}
    \vspace{-0.2cm}
\end{table*}

Table~\ref{tab:acdc_comparison} presents the quantitative results on Cityscapes-to-ACDC. When using 12.5\% of target labeled data, SS-ADA achieves the accuracy of supervised learning. With 25\% of target labeled data, the performance of SS-ADA is comparable to the joint training. From the third column of Figure~\ref{fig:sup_semi_our_comparison_u2pl}, SS-ADA shows steady mIoU improvements over semi-supervised methods U2PL and UniMatch across different ratios of target labeled data. These results highlight the effectiveness of SS-ADA in the normal-to-adverse weather domain adaptation setting.  

\subsubsection{Cityscapes-to-FishEyeCampus}
\begin{table*}[!ht]
    \centering
    \setlength{\tabcolsep}{3pt}
    \renewcommand{\arraystretch}{1.05}
    \captionsetup{format=myformat}
    \caption{The performance of SS-ADA on Cityscapes-to-FishEyeCampus.} 
    \label{tab:fisheye_comparison}
    \resizebox{\linewidth}{!}{
    \begin{tabular}{c|c|c|cccccccccccccccccc}
        \hline
        Methods & Type & Ratio & mIoU & road & sw & build & guard & pole & light & sign & vege & terrain & sky & person & rider & car & truck & bus & motor & bike \\ \hline
        \multirow{2}{*}{UniMatch\cite{unimatch}} & joint & 100 & \underline{67.2}  & \underline{98.3}  & \textbf{70.9}  & \underline{74.6}  & \textbf{56.6}  & 28.0  & \underline{43.3}  & \underline{47.0}  & \underline{88.8}  & {65.2}  & 94.8  & 57.2  & 50.4  & \textbf{87.2}  & \textbf{87.4}  & \underline{77.6}  & 52.1  & 63.1 \\
        & sup & 100 & {66.8}  & \underline{98.3}  & {67.9}  & \textbf{74.7}  & \underline{56.4}  & 31.4  & 40.0  & 44.2  & \textbf{89.2}  & \textbf{66.1}  & \textbf{95.2}  & \textbf{59.6}  & \underline{54.8}  & 86.9  & \underline{86.4}  & 70.5  & 50.5  & 63.1 \\ \hline
        \multirow{3}{*}{Ours (UniMatch)} & \multirow{3}{*}{ADA} & 5 & 63.9  & 98.0  & 57.0  & 72.3  & 46.8  & {32.6}  & 34.2  & 33.1  & 86.5  & 51.8  & 93.9  & 57.2  & {54.0}  & \underline{87.1}  & 84.5  & \textbf{79.0}  & \underline{54.1}  & \underline{64.4} \\
        & & 12.5 & 66.3  & {98.1}  & 61.4  & 73.8  & 51.9  & \underline{33.4}  & \textbf{44.2}  & \textbf{47.2}  & 88.2  & 62.2  & 95.0  & 58.7  & 53.0  & 86.0  & 85.3  & 75.4  & 52.5  & 61.5 \\
        & & 25 & \textbf{67.8}  & \textbf{98.4}  & \underline{68.2}  & \underline{74.6}  & 52.8  & \textbf{37.7}  & 43.0  & 45.5  & \textbf{89.2}  & \underline{65.9}  & \underline{95.1}  & \underline{59.5}  & \textbf{56.3}  & 86.3  & 84.3  & 76.1  & \textbf{55.4}  & \textbf{64.9} \\
        \hline \hline
        \multirow{2}{*}{U2PL\cite{u2pl}} & joint & 100 & \underline{66.2}  & \textbf{98.2}  & \textbf{67.2}  & \underline{73.3}  & \underline{55.7}  & 30.4  & \textbf{43.0}  & \textbf{46.0}  & \underline{88.3}  & \textbf{65.1}  & \textbf{94.9}  & \textbf{59.0}  & \underline{54.2}  & \textbf{87.2}  & 83.0  & 67.4  & 50.0 & \underline{63.0} \\
        & sup & 100 & {65.1}  & \underline{98.0}  & 65.6  & 73.2  & 55.3  & 27.1  & 36.4  & 39.7  & \underline{88.3}  & \underline{62.4}  & 94.7  & 56.5  & 52.8  & \underline{86.6}  & \textbf{85.3}  & \underline{77.1}  & 47.8  & 60.6 \\ \hline
        \multirow{4}{*}{Ours (U2PL)} & \multirow{4}{*}{ADA}  & 5 & 55.1  & 96.4  & 46.9  & 68.7  & 33.5  & 25.3  & 37.3  & 31.8  & 86.9  & 40.5  & 94.0  & 51.6  & 46.5  & 84.0  & 67.1  & 27.4  & 41.5  & 58.0 \\ 
        & & 12.5 & 64.5  & 97.9  & 60.4  & 72.5  & 45.4  & \underline{32.2}  & 39.7  & 42.6  & 88.0  & 55.9  & \underline{94.8}  & \textbf{58.9}  & \textbf{54.6}  & 84.3  & 81.4  & 74.2  & \underline{50.6}  & 62.6 \\
        & & 25 & \textbf{66.5}  & \underline{98.0}  & \underline{66.0}  & \textbf{73.8}  & \textbf{56.0}  & \textbf{35.5}  & \underline{41.2}  & \underline{43.9}  & \textbf{88.5}  & \underline{62.4}  & 94.7  & 55.5  & 47.7  & 85.7  & \underline{84.7}  & \textbf{78.3}  & \textbf{53.3}  & \textbf{64.8} \\
        \hline
    \end{tabular}}
    \vspace{-0.3cm}
\end{table*}
Table~\ref{tab:fisheye_comparison} shows the results on Cityscapes-to-FishEyeCampus. With 25\% of target labeled data, SS-ADA reaches the mIoU of both supervised learning and joint training when employing either UniMatch or U2PL. In the last column of Figure~\ref{fig:sup_semi_our_comparison_u2pl}, SS-ADA consistently improves the accuracy of semi-supervised methods across different annotation ratios. This demonstrates the effectiveness of it in transferring from pinhole to fisheye camera images. \vspace{-0.2cm}

\begin{figure*}[htbp]
    \centering
    \vspace{-0.2cm}
    \captionsetup[subfloat]{font=scriptsize,labelfont=scriptsize,labelformat=empty}
    \subfloat[All classes]{\includegraphics[height=0.244\linewidth]{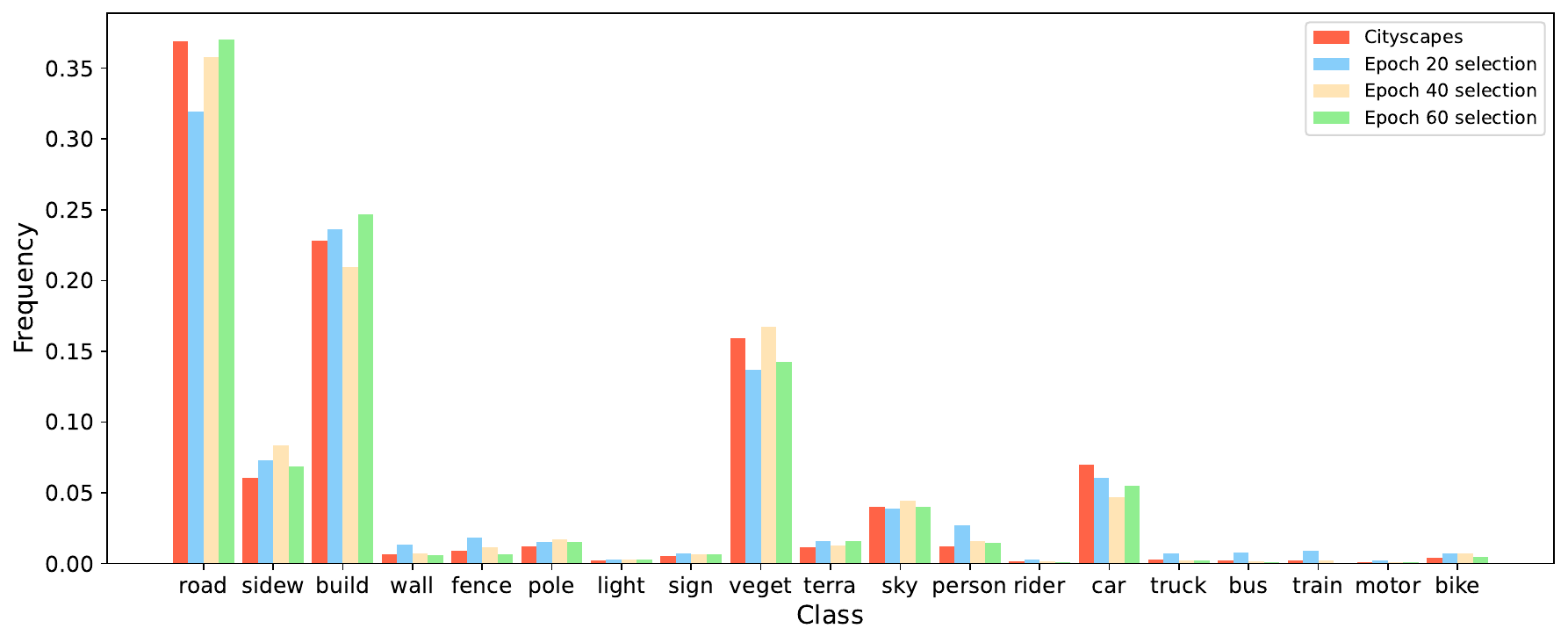}}
    \subfloat[Selected classes]{\includegraphics[height=0.244\linewidth]{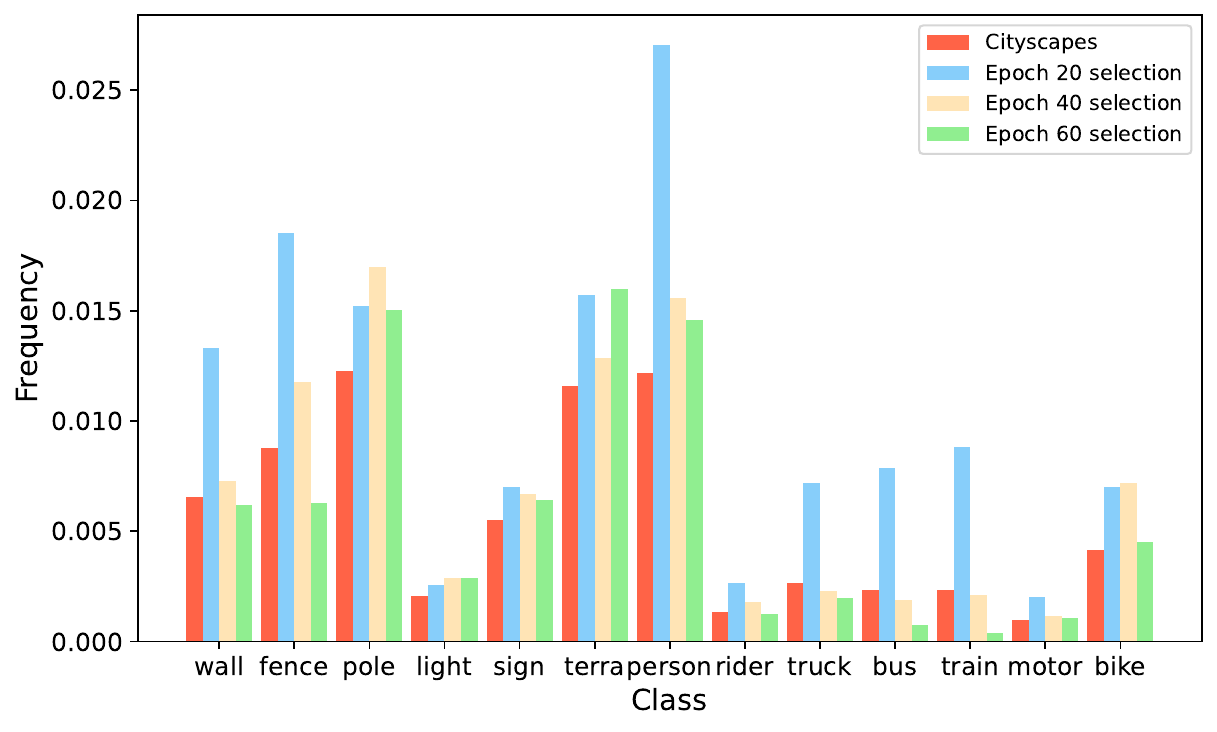}}
    \vspace{-0.2cm}
    \caption{Class frequency distribution of samples selected by the active learning module in the GTA-to-Cityscapes experiment using 50\% of the target annotated data. the red bars represent the original class frequency distribution in Cityscapes.}
    \label{fig:fre_compare_city}
    \vspace{-0.4cm}
\end{figure*} 

\subsection{Demonstration of the active learning selection process}
\begin{figure}[htbp]
    \centering
    \vspace{-0.2cm}
    \captionsetup[subfloat]{font=scriptsize,labelfont=scriptsize,labelformat=empty}
    \subfloat{\includegraphics[width=0.33\linewidth]{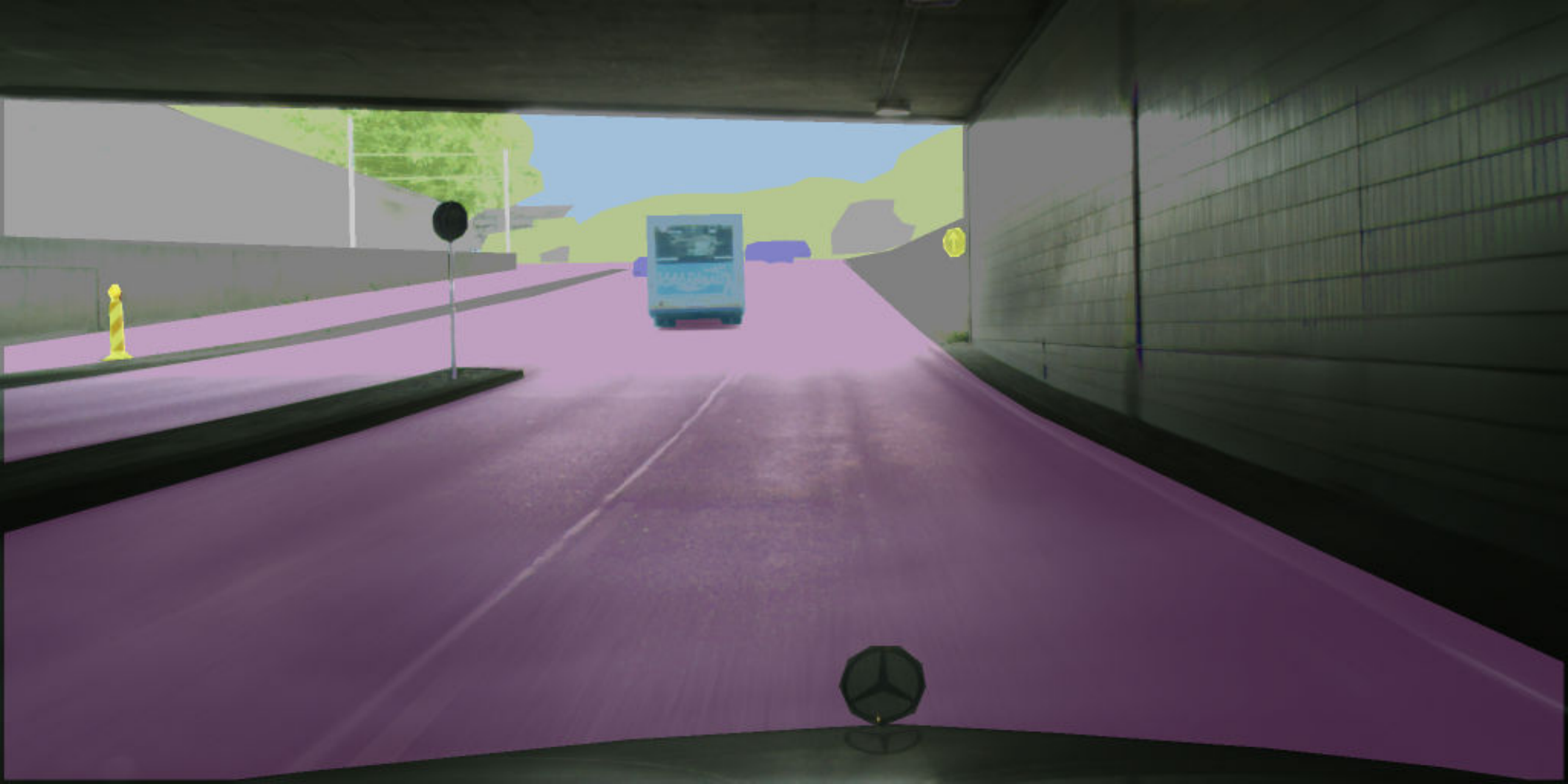}}\hfill
    \subfloat{\includegraphics[width=0.33\linewidth]{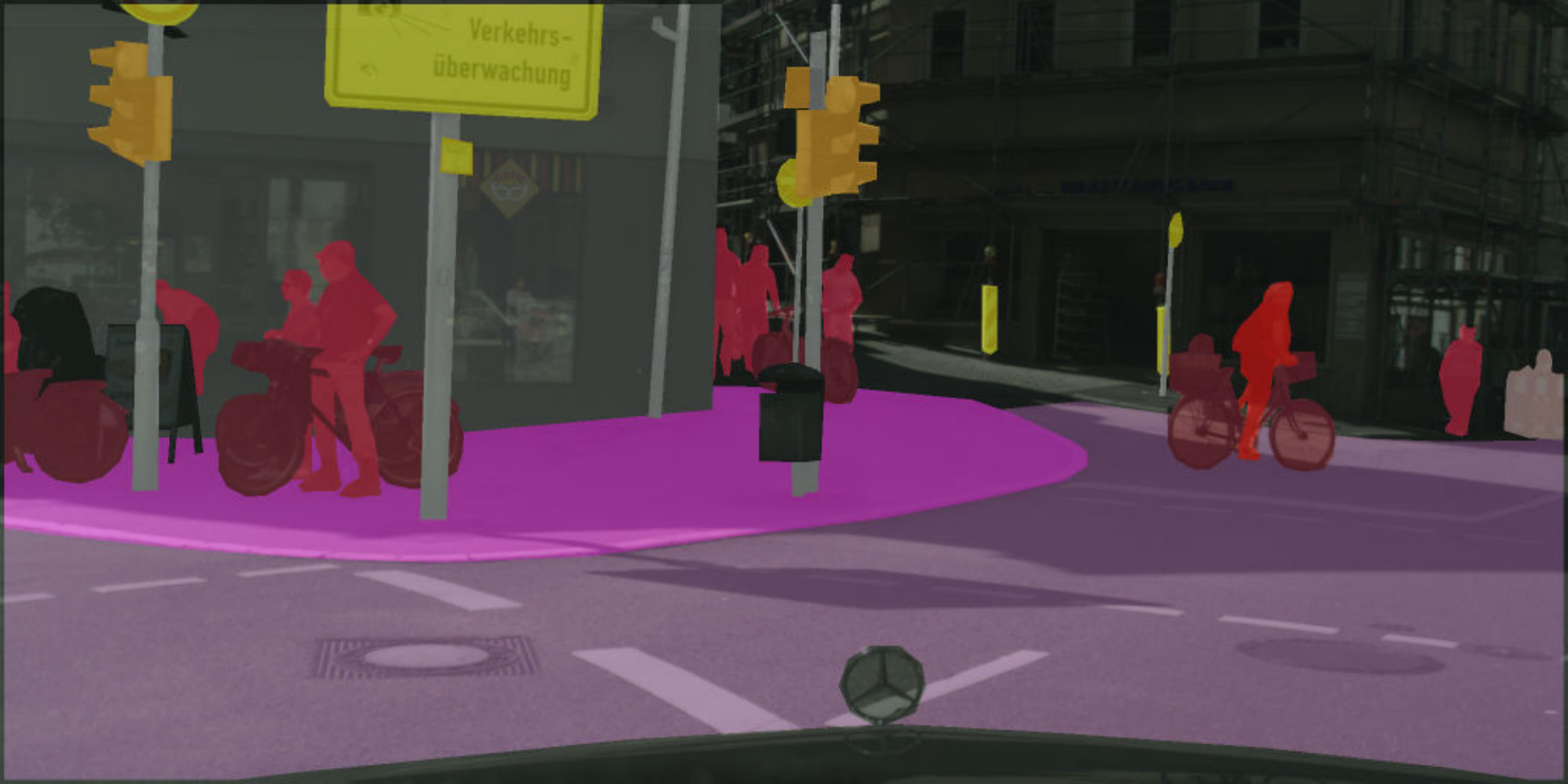}}\hfill
    \subfloat{\includegraphics[width=0.33\linewidth]{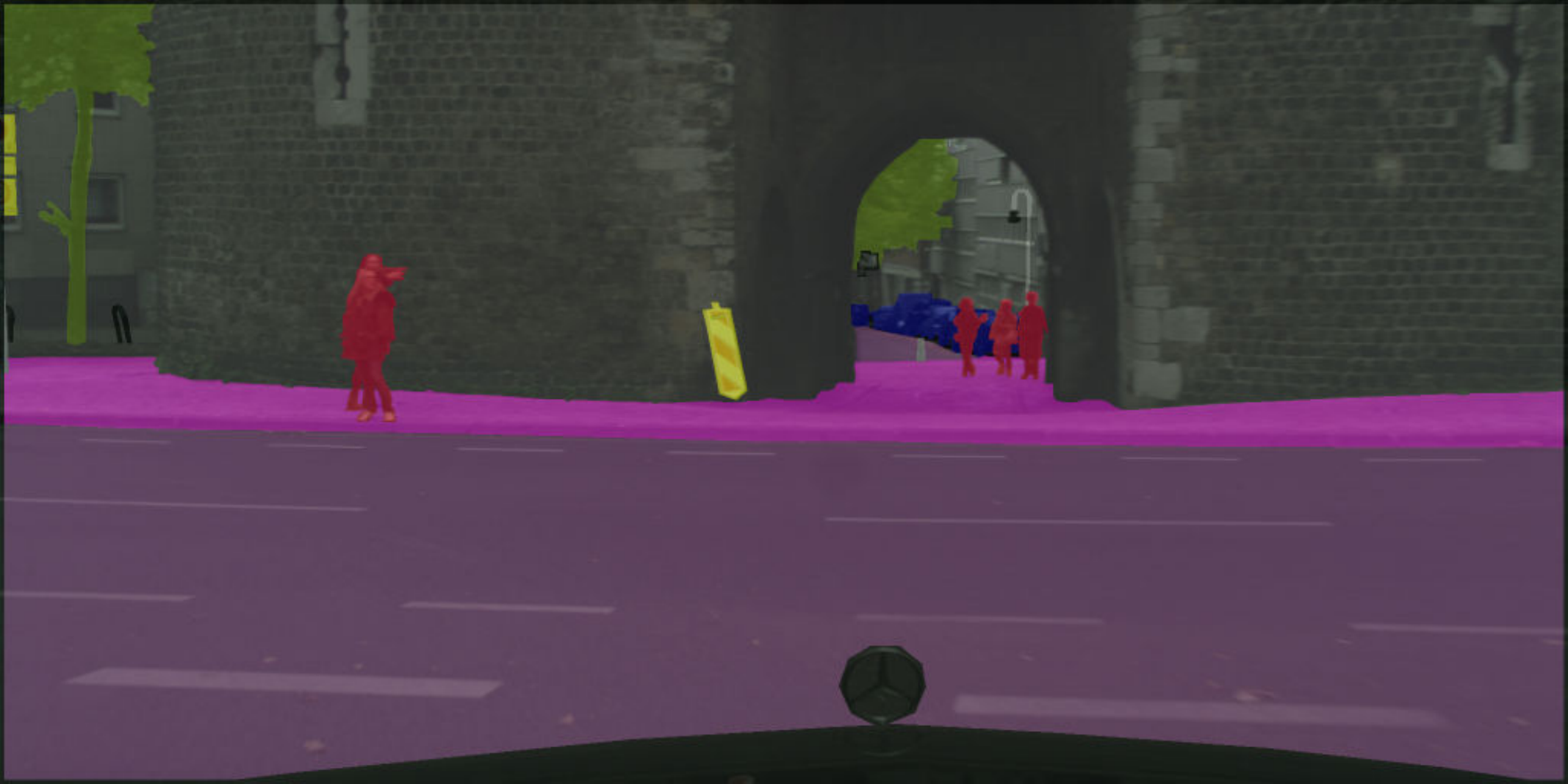}}\\ \vspace{-0.3cm}
    \subfloat{\includegraphics[width=0.33\linewidth]{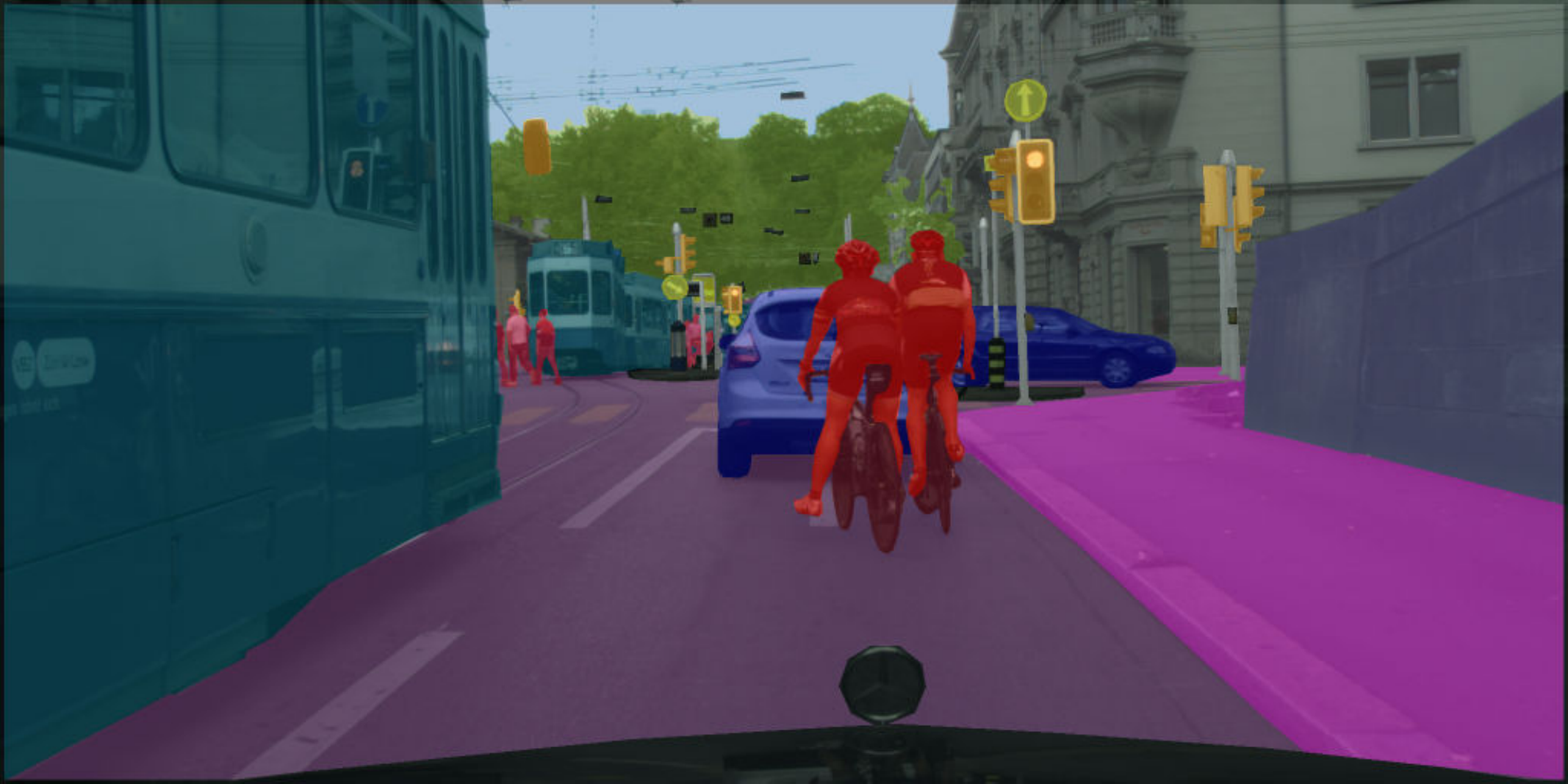}}\hfill
    \subfloat{\includegraphics[width=0.33\linewidth]{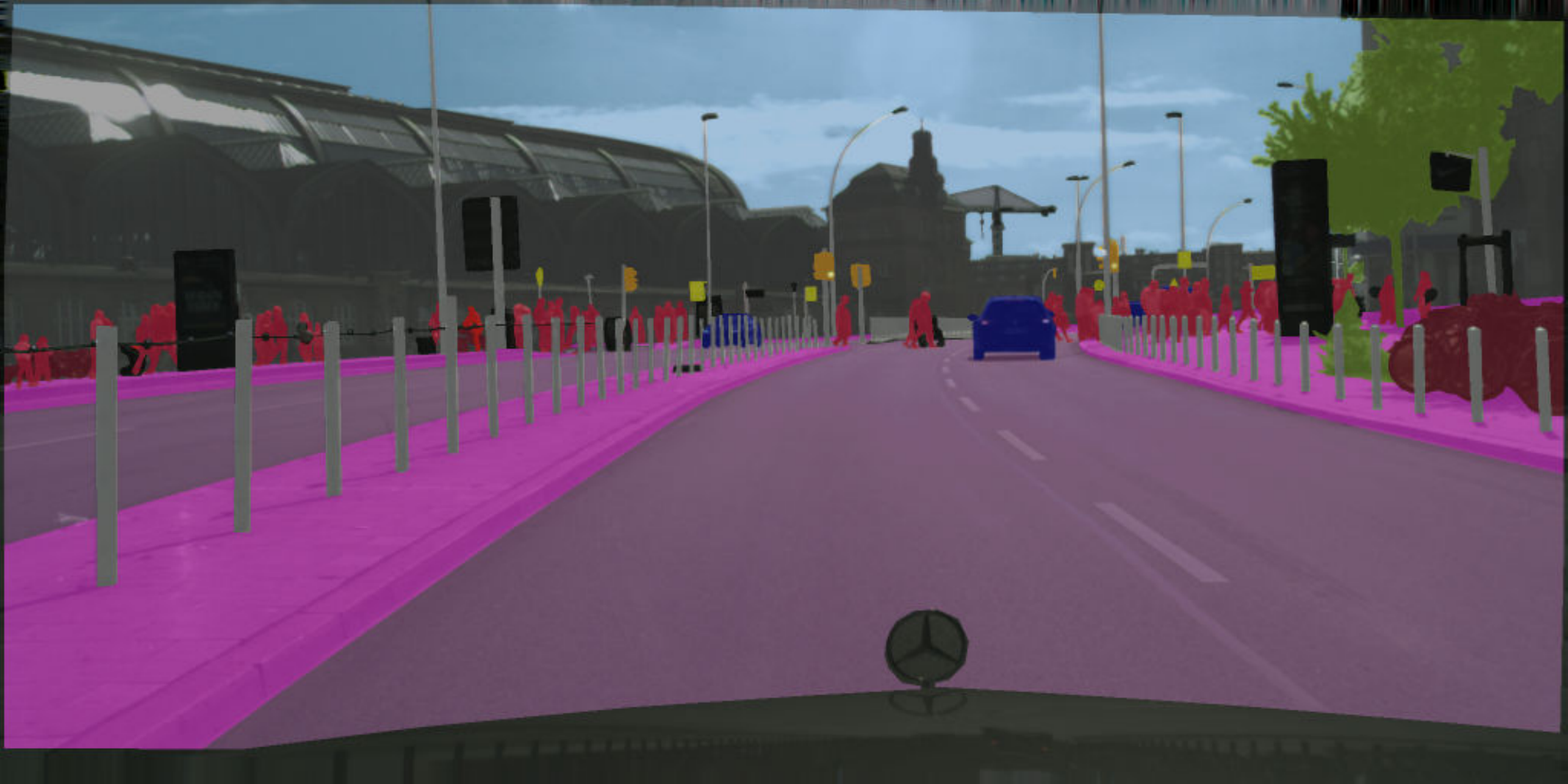}}\hfill
    \subfloat{\includegraphics[width=0.33\linewidth]{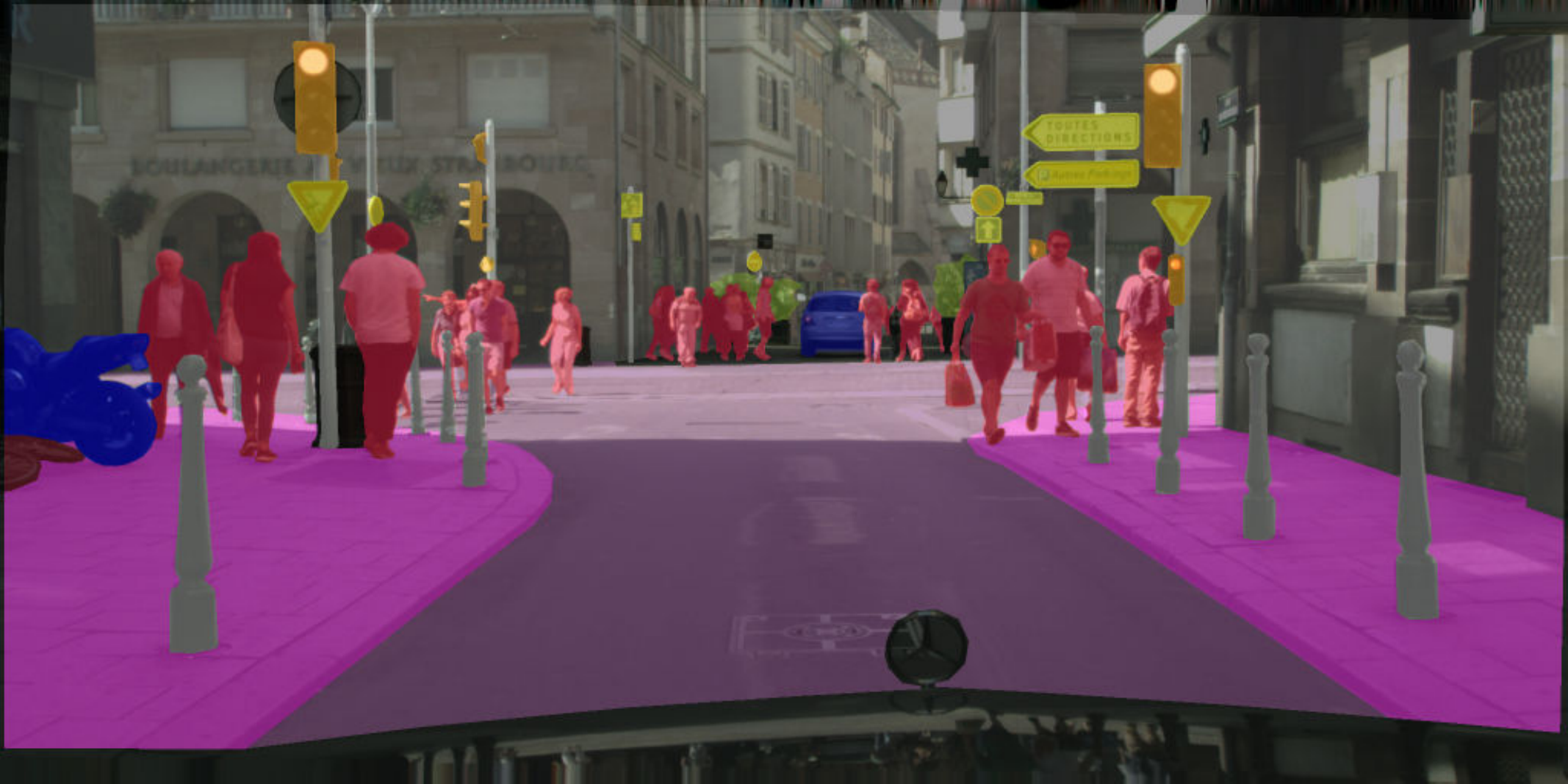}}\\ \vspace{-0.3cm}
    \subfloat{\includegraphics[width=0.33\linewidth]{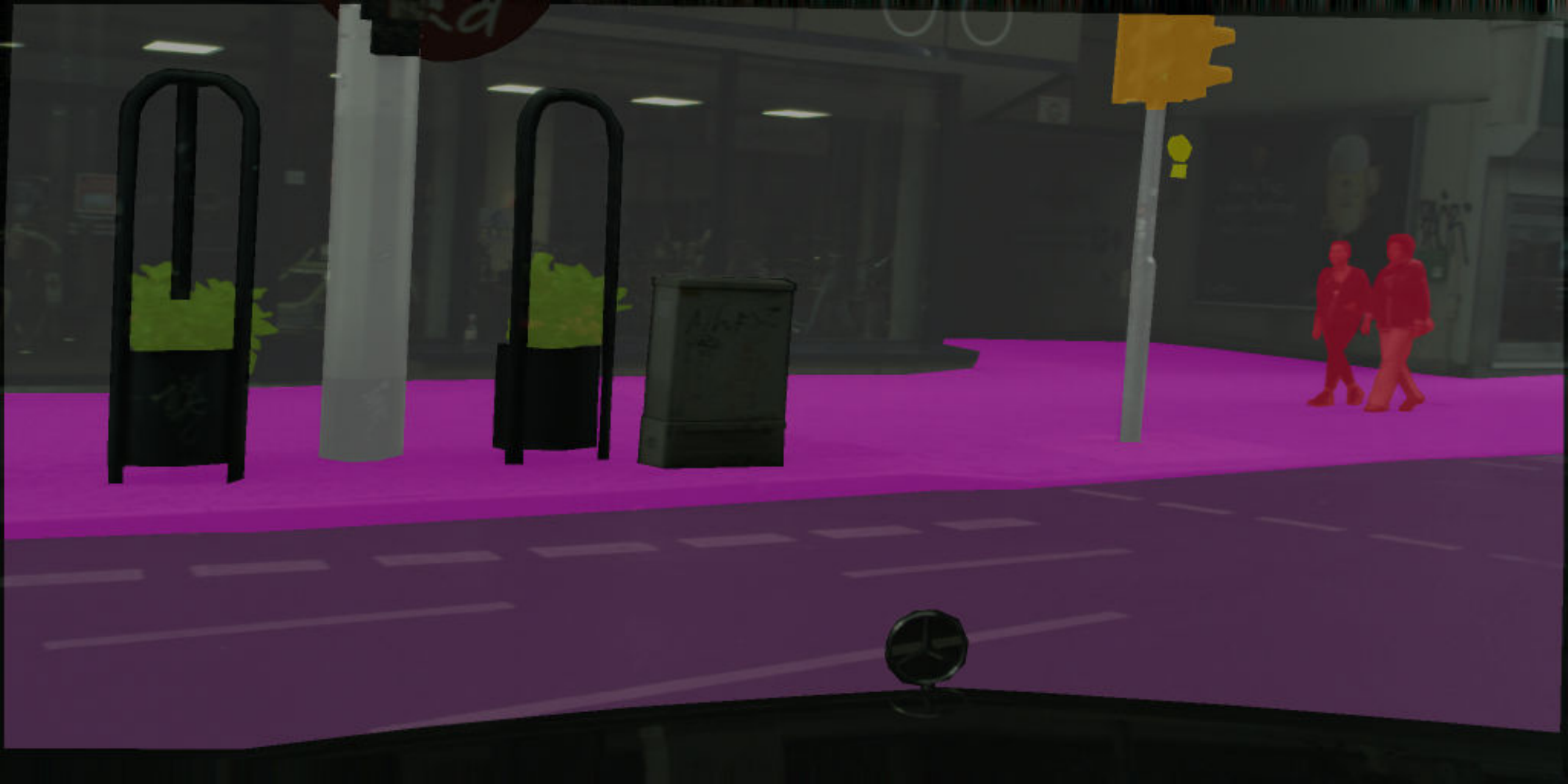}}\hfill
    \subfloat{\includegraphics[width=0.33\linewidth]{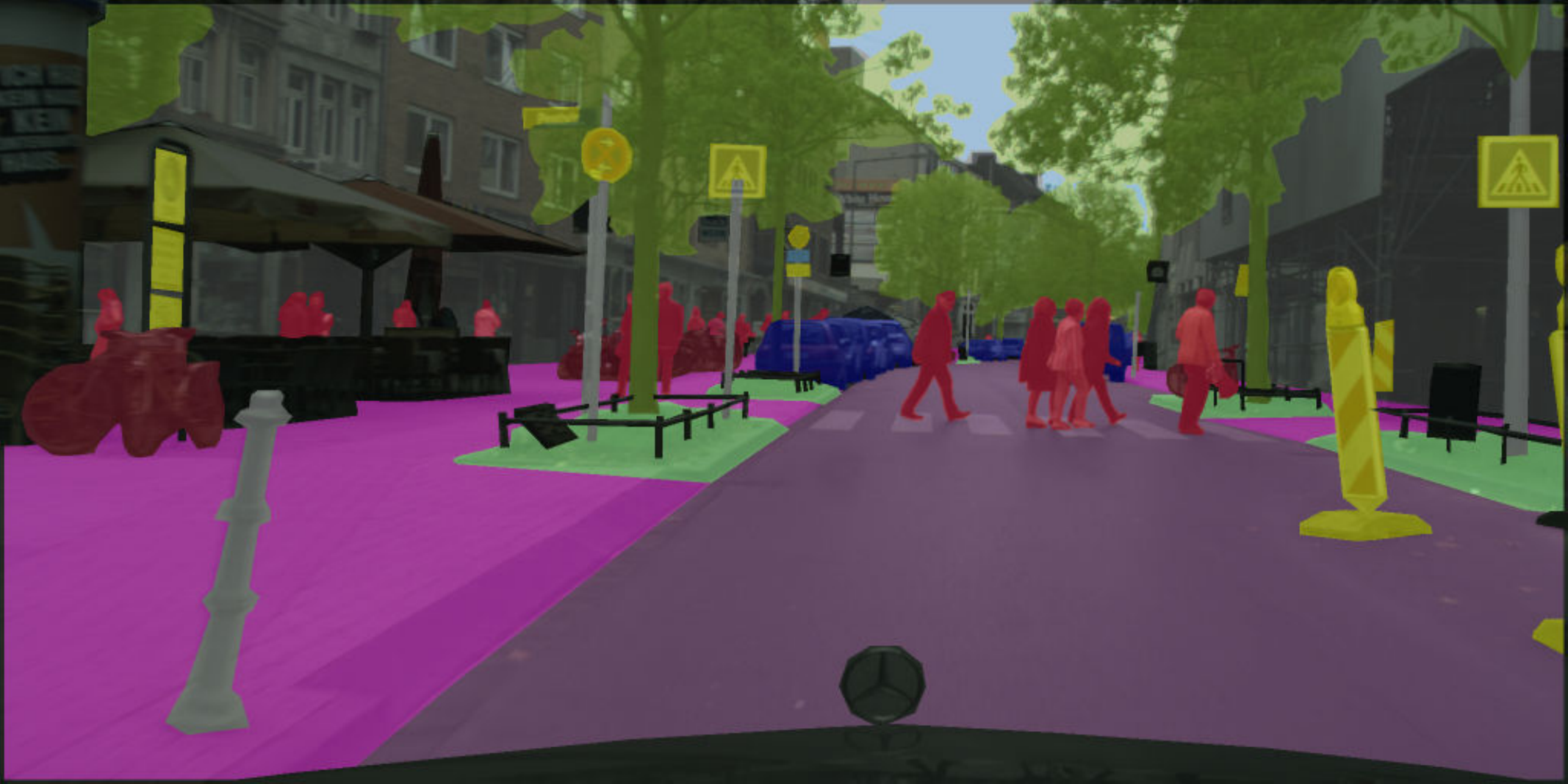}}\hfill
    \subfloat{\includegraphics[width=0.33\linewidth]{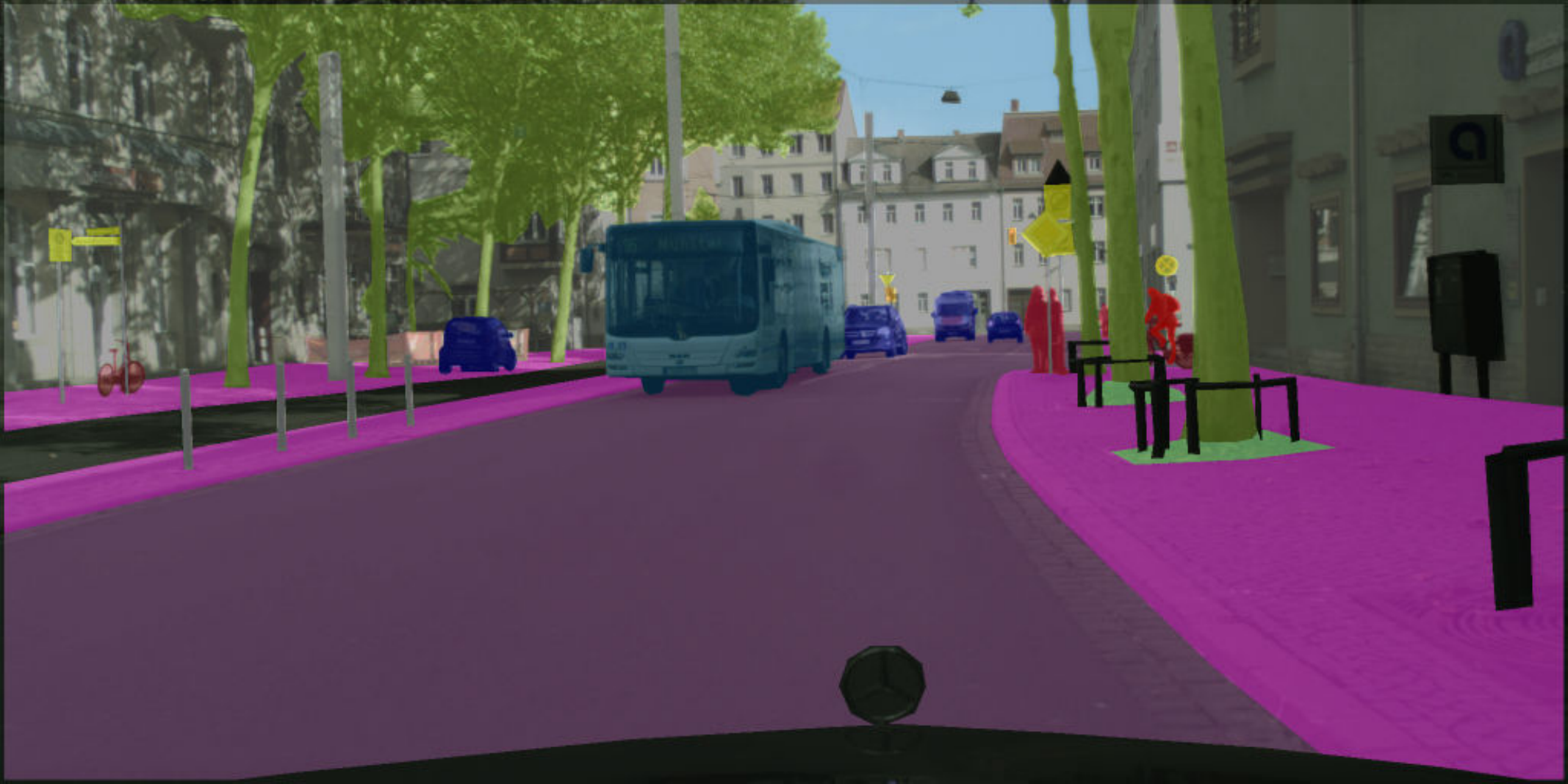}}\\ \vspace{-0.3cm}
    \subfloat[Epoch 20]{\includegraphics[width=0.33\linewidth]{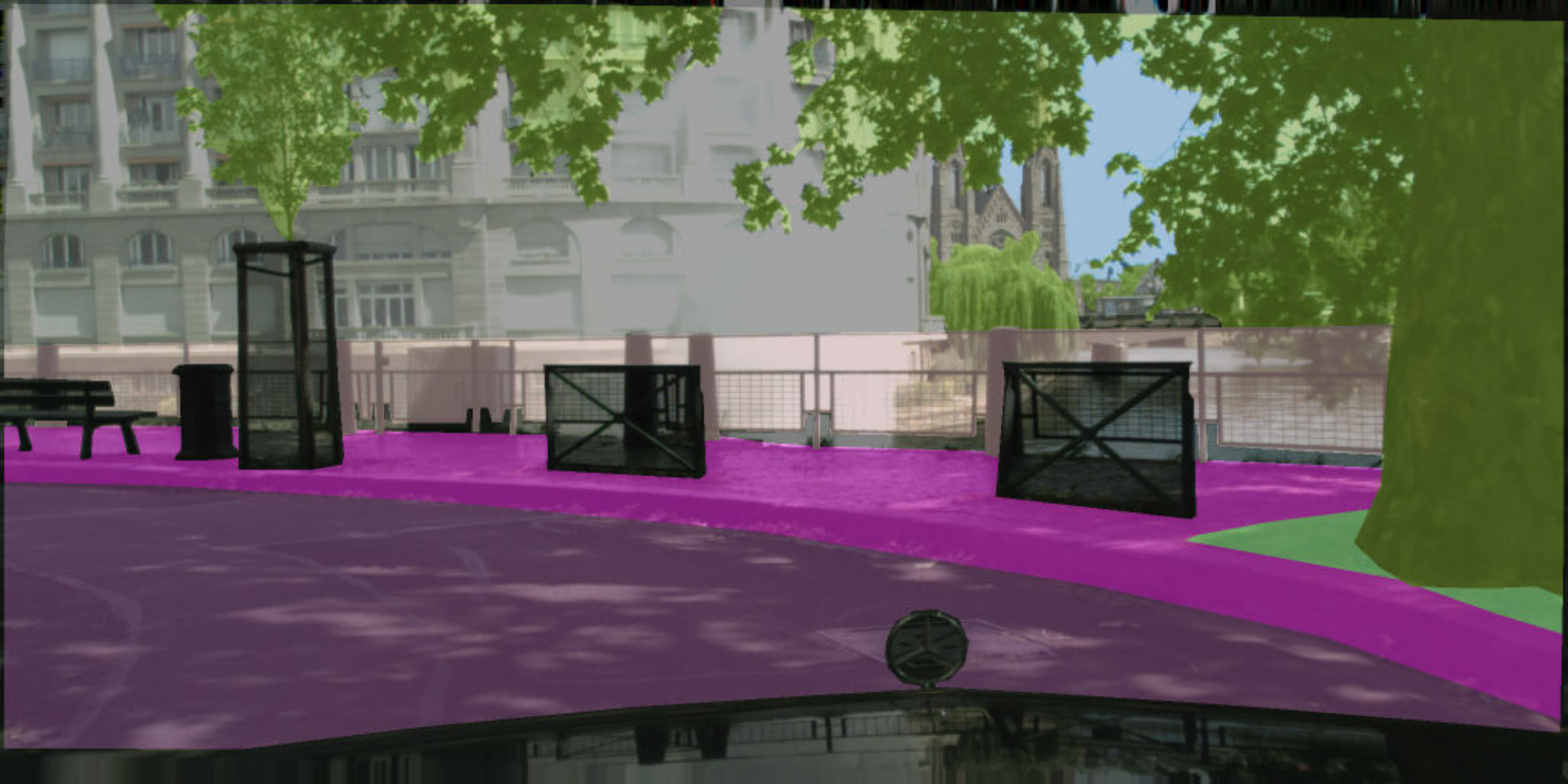}}\hfill
    \subfloat[Epoch 40]{\includegraphics[width=0.33\linewidth]{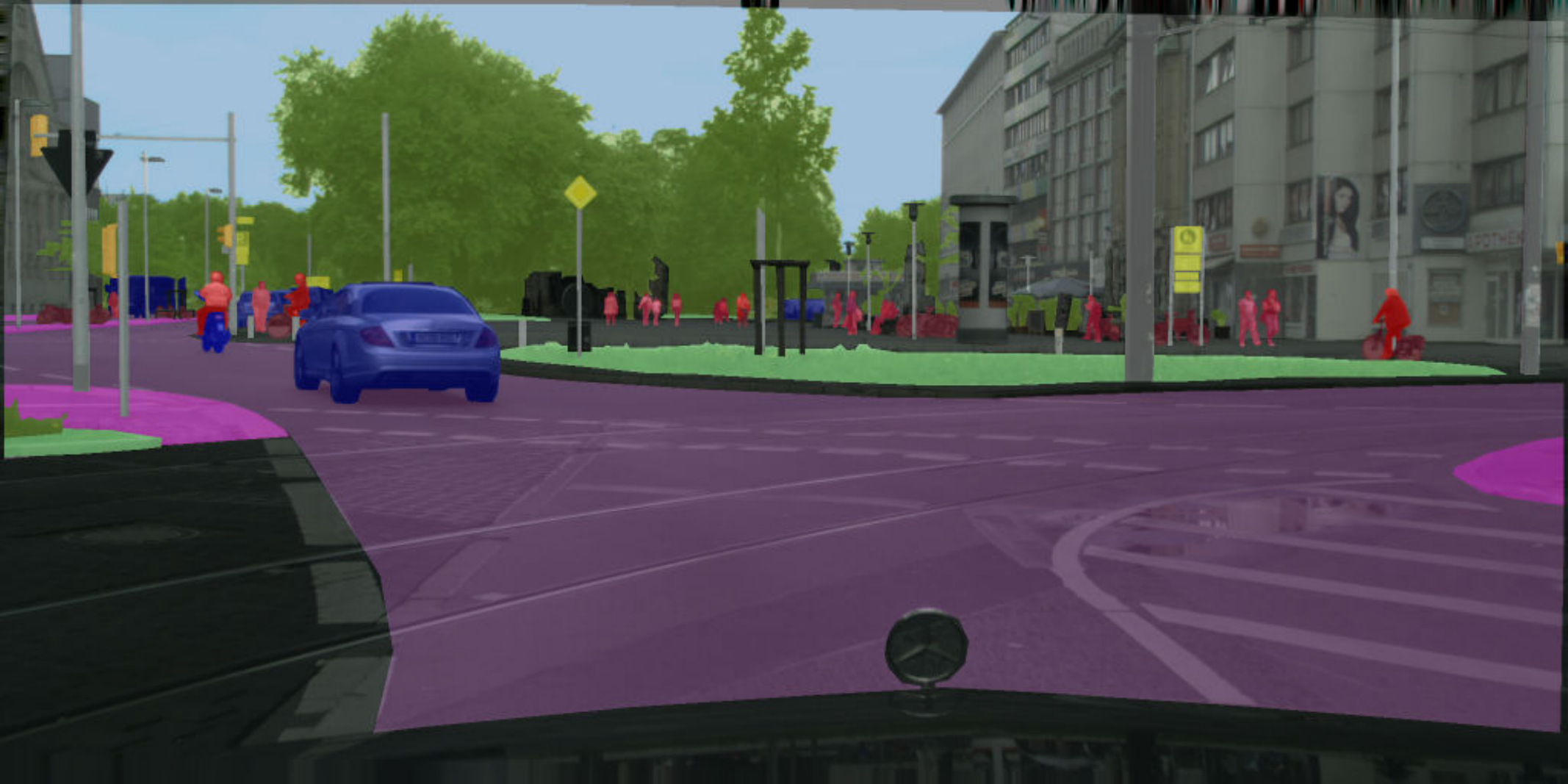}}\hfill
    \subfloat[Epoch 60]{\includegraphics[width=0.33\linewidth]{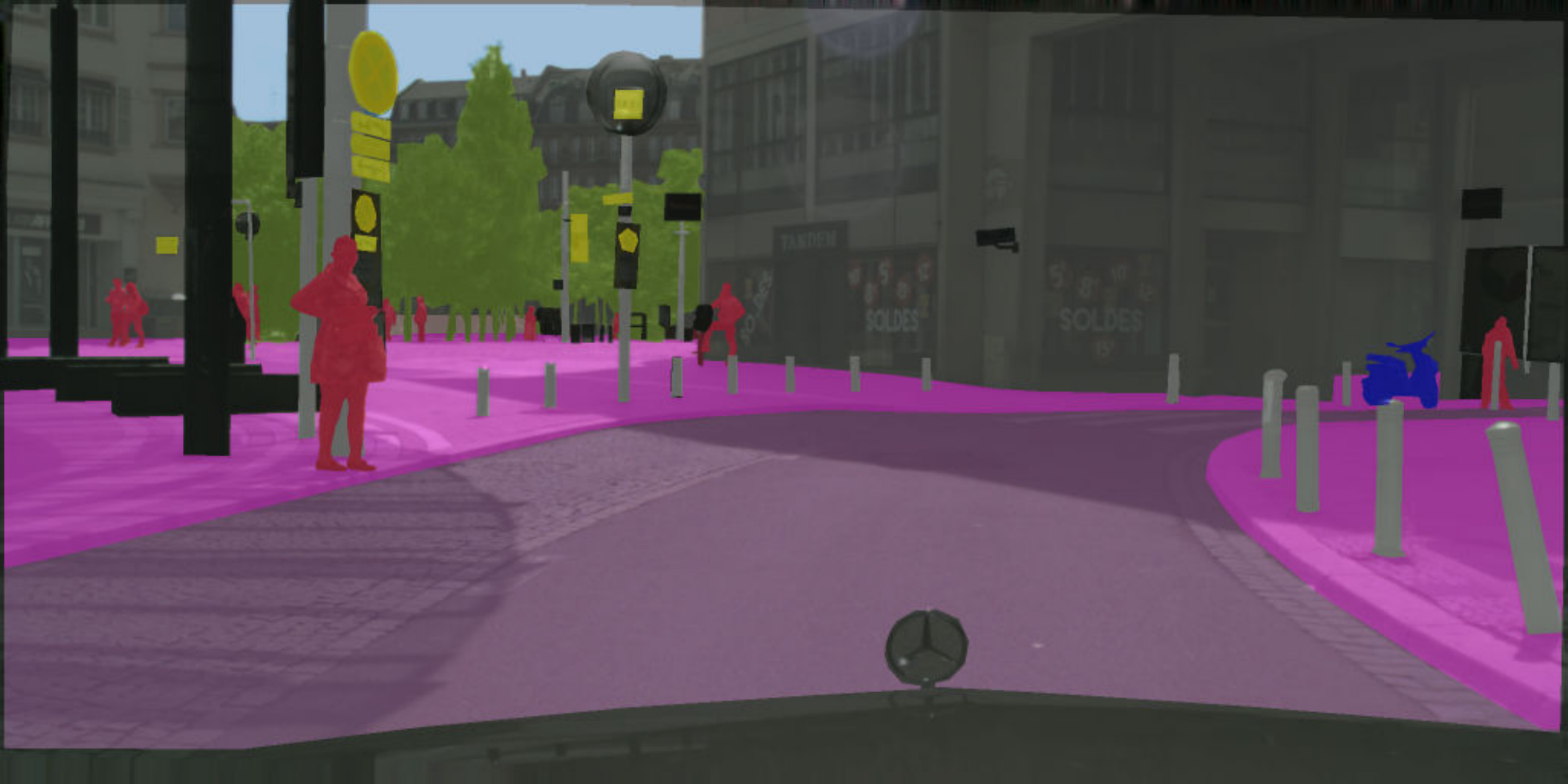}}\\ \vspace{-0.15cm}
    \caption{The top 4 images and annotated labels selected by the active learning module at epoches 20, 40 and 60 in the GTA-to-Cityscapes experiment using 50\% of the target labeled data.}
    \label{fig:ac_selection}
    \vspace{-0.3cm}
\end{figure}

We analyze the selection results of active learning module in GTA5-to-Cityscapes with 50\% of the target labeled data. The module is triggered at training rounds 20, 40, and 60, each time selecting 486 images for manual annotation. 
The class frequency distribution of the annotated labels for the images selected in these three rounds is shown in Figure~\ref{fig:fre_compare_city}. It can be observed that the selected classes have relatively low class frequencies in Cityscapes. After the active learning module, the frequency proportion of these classes is significantly increased, especially during the first selection at epoch 20. 
The top 4 images with corresponding labels selected by the active learning module are presented in Figure~\ref{fig:ac_selection}.
It can be seen that the active learning module selected some rare scenes like backlit and turning, and images with buses and person.

\subsection{Class reweighting analysis}
\begin{figure}[htbp]
    \centering
    \includegraphics[width=0.95\linewidth]{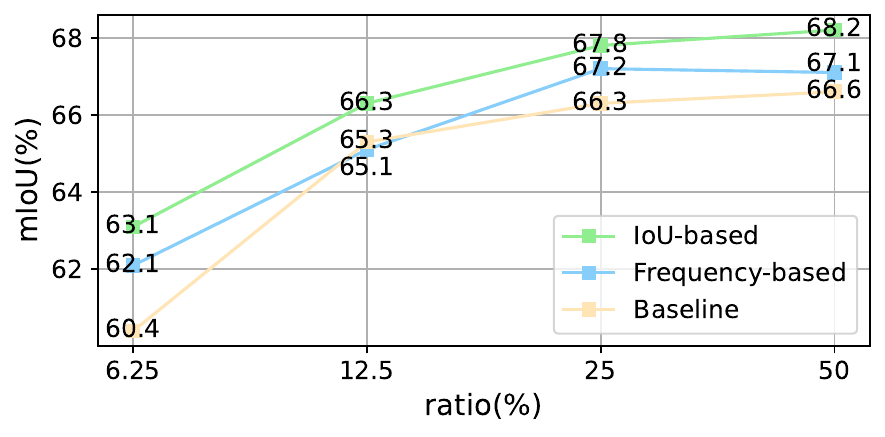}
    \vspace{-0.1cm}
    \caption{Comparison of IoU-based and frequency-based class weighting strategies on Cityscapes-to-FishEyeCampus. The upper bound $u$ is set as 5 for both of them.}
    \label{fig:weighting_comparison}
    \vspace{-0.3cm}
\end{figure}

The IoU-based class weighting is compared with widely used frequency-based class weighting\cite{class_weight}, where the weights are inversely proportional to the class frequency. Figure~\ref{fig:weighting_comparison} shows the results on Cityscapes-to-FishEyeCampus under different ratios of target labeled data. The IoU-based class weighting strategy outperforms the frequency-based one in all settings. This indicates that the IoU-based approach better captures the segmentation performance of the model on different classes and assigns more appropriate class weights.  

\subsection{Comparison with ADA methods}
\begin{figure*}[!tbp]
    \centering
    \captionsetup[subfloat]{font=scriptsize,labelfont=scriptsize,labelformat=empty}
    \subfloat[GTA5-to-Cityscapes]{\includegraphics[width=0.33\linewidth]{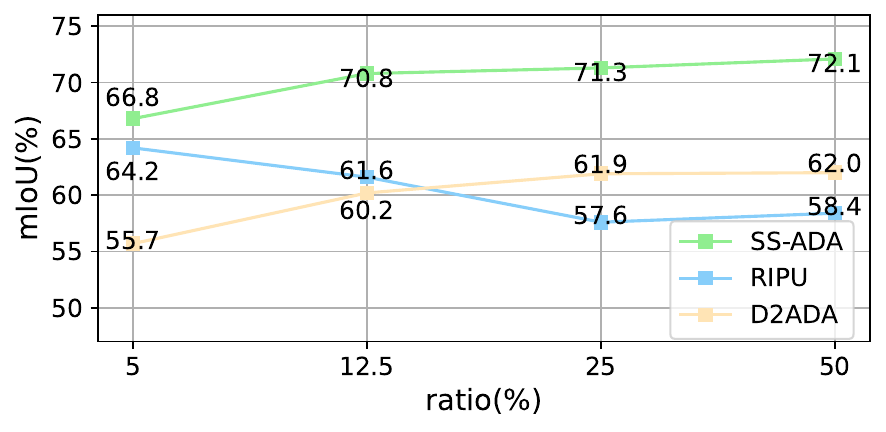}} \hfill
    \subfloat[SYNTHIA-to-Cityscapes]{\includegraphics[width=0.33\linewidth]{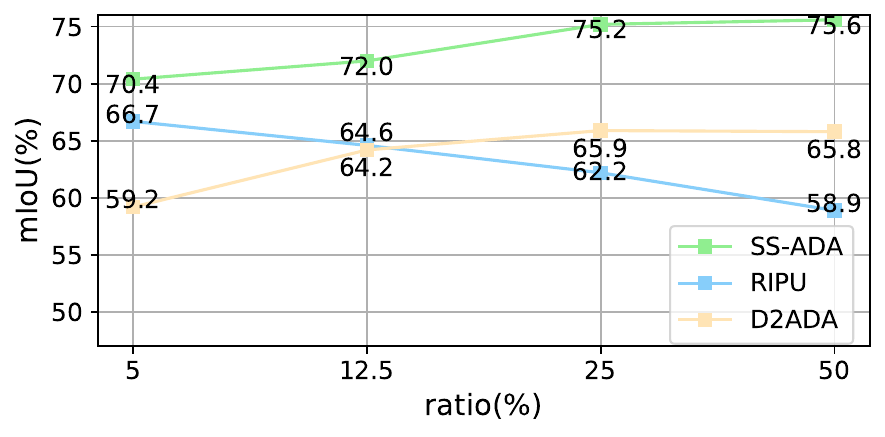}}  \hfill
    \subfloat[Cityscapes-to-ACDC]{\includegraphics[width=0.33\linewidth]{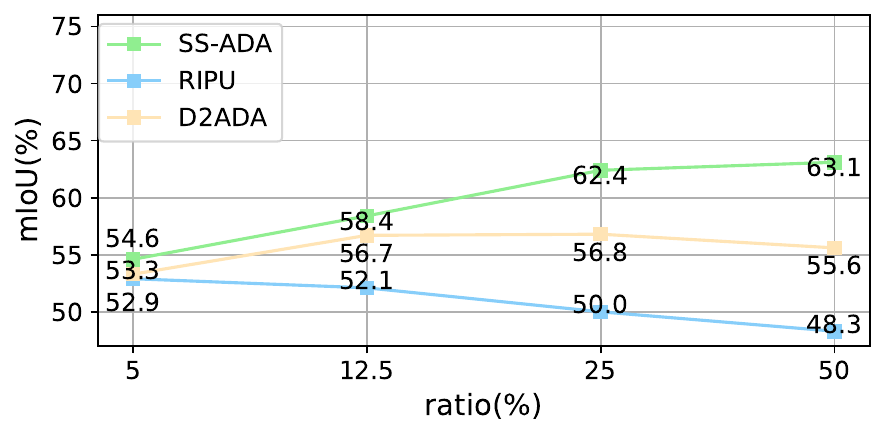}}
    \caption{The comparison between SS-ADA and ADA methods for semantic segmentation.}
    \label{fig:ada_compare}
    \vspace{-0.4cm}
\end{figure*}
Figure~\ref{fig:ada_compare} presents the performance of ADA methods RIPU, D2ADA, and our SS-ADA across different annotation ratios. SS-ADA consistently outperforms RIPU and D2ADA in all settings. As the annotation ratio increases, SS-ADA's accuracy also improves gradually. The performance of D2ADA increases initially but then plateaus or even decreases. While RIPU exhibits a noticeable decrease in accuracy, a trend also observed in the literature\cite{halo}. This suggests that image-level acquisition strategies are less sensitive to the class imbalance problem, possibly because they preserve rich contextual information within the images. Additionally, as the ratio of annotated data increases, the improvement in mIoU slows down, illustrating the characteristic of active learning in selecting valuable samples. More valuable samples are chosen when the annotation ratio is low, whereas adding redundant samples does not significantly enhance model accuracy.

\subsection{Qualitative Results and Analysis}

\begin{figure*}
    \centering
    \captionsetup[subfloat]{font=scriptsize,labelfont=scriptsize,labelformat=empty}
    \subfloat[]{\includegraphics[width=0.142\linewidth]{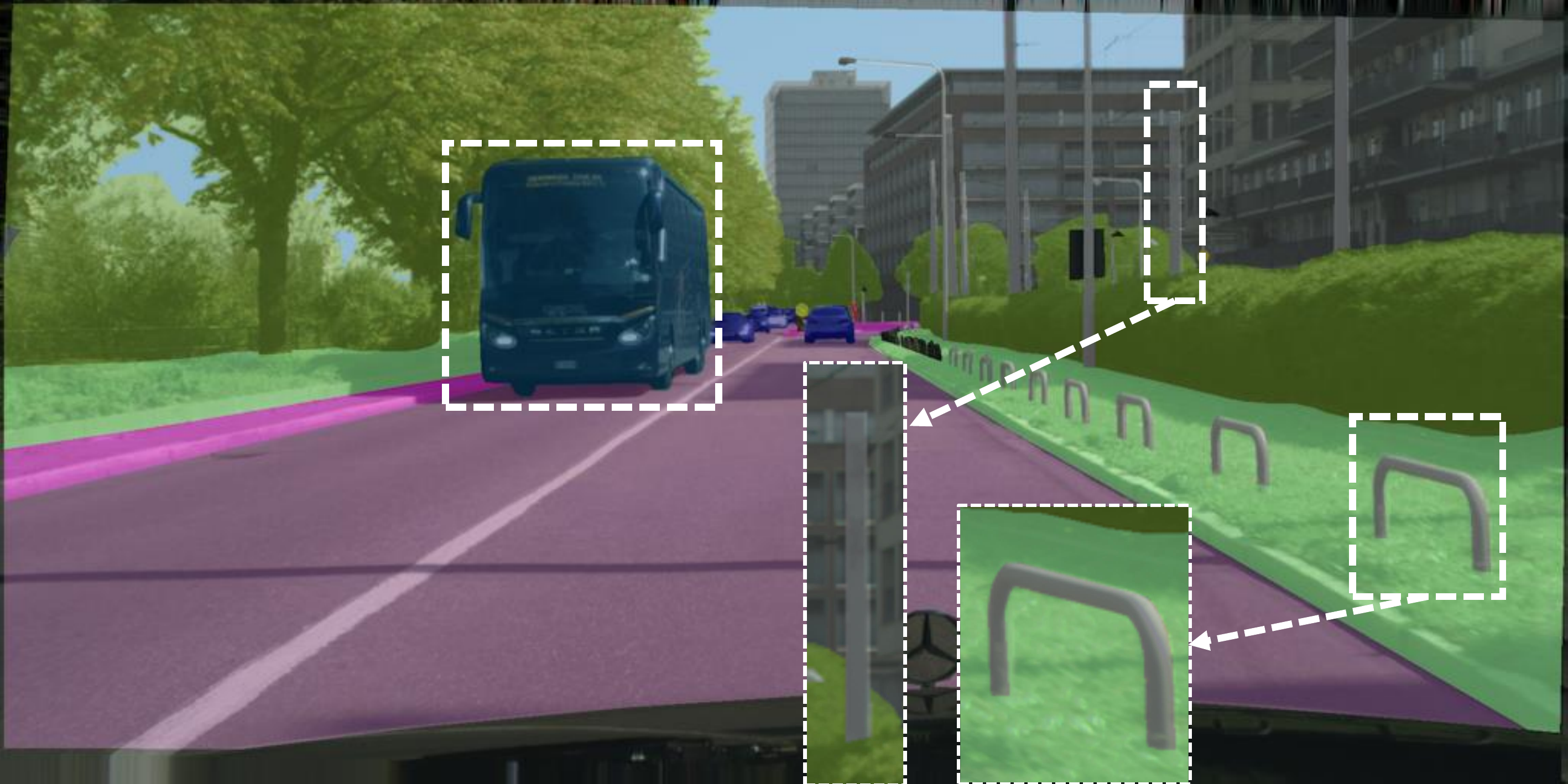}} \hfill
    \subfloat[]{\includegraphics[width=0.142\linewidth]{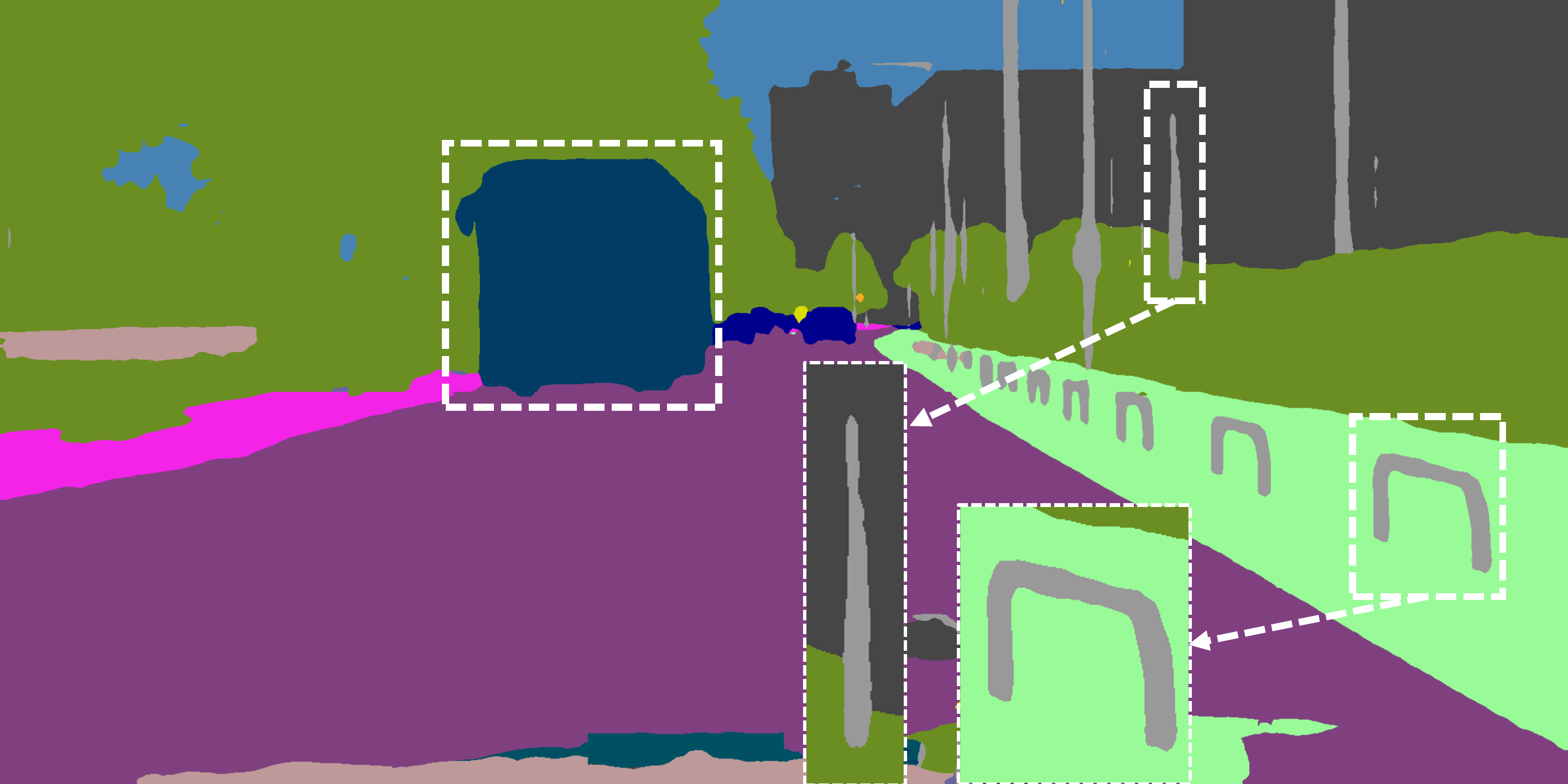}} \hfill
    \subfloat[]{\includegraphics[width=0.142\linewidth]{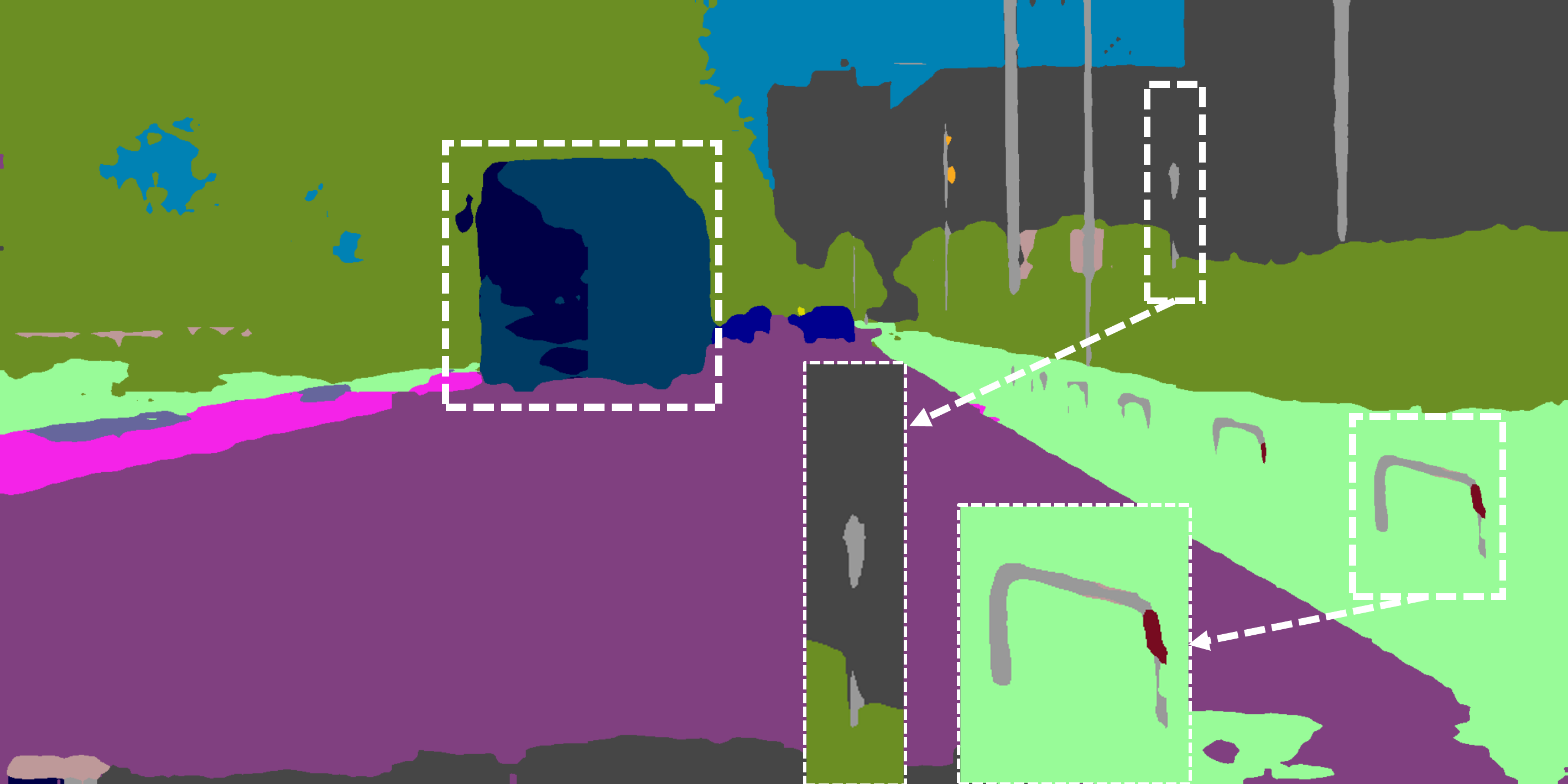}} \hfill
    \subfloat[]{\includegraphics[width=0.142\linewidth]{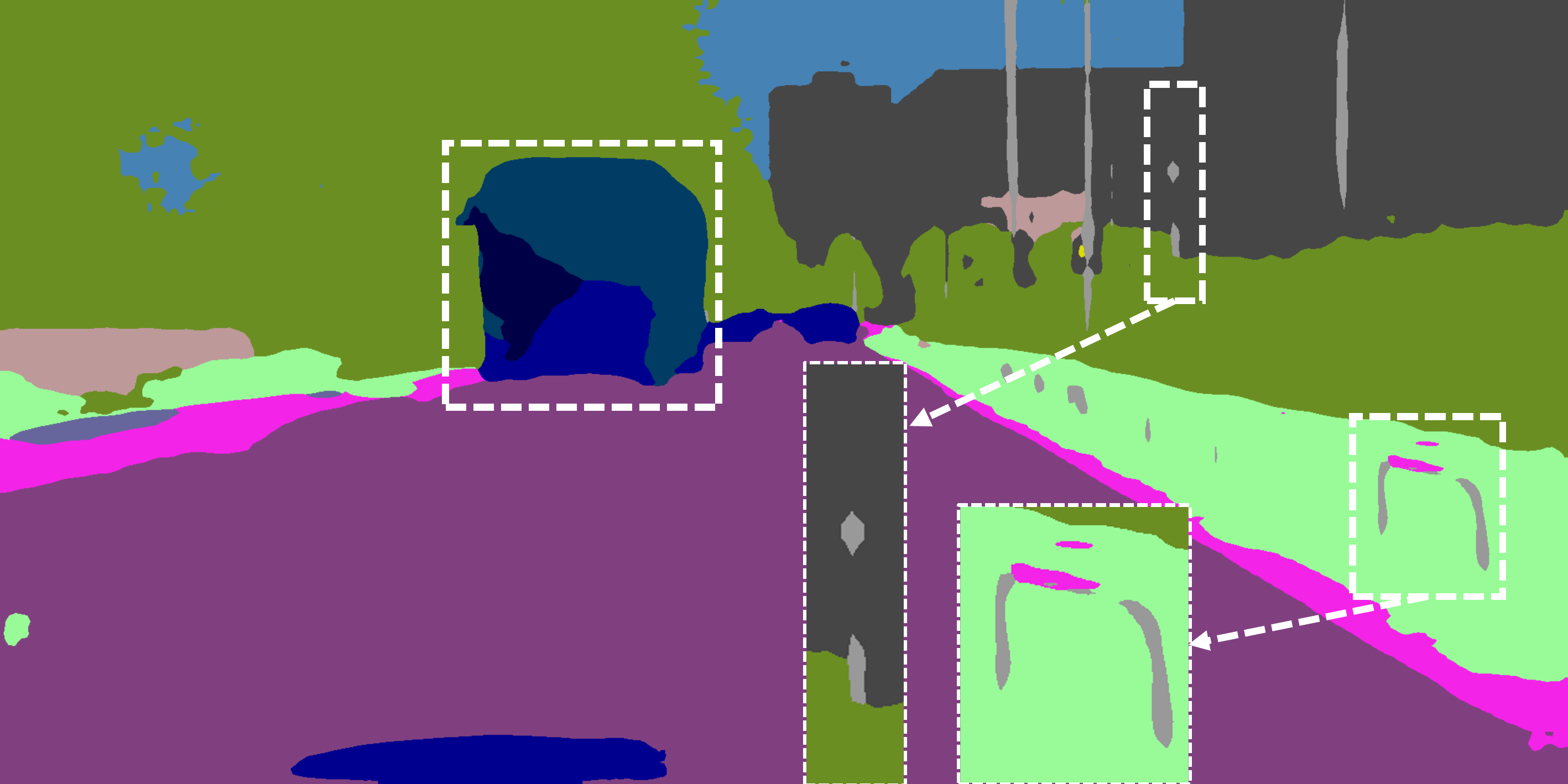}} \hfill
    \subfloat[]{\includegraphics[width=0.142\linewidth]{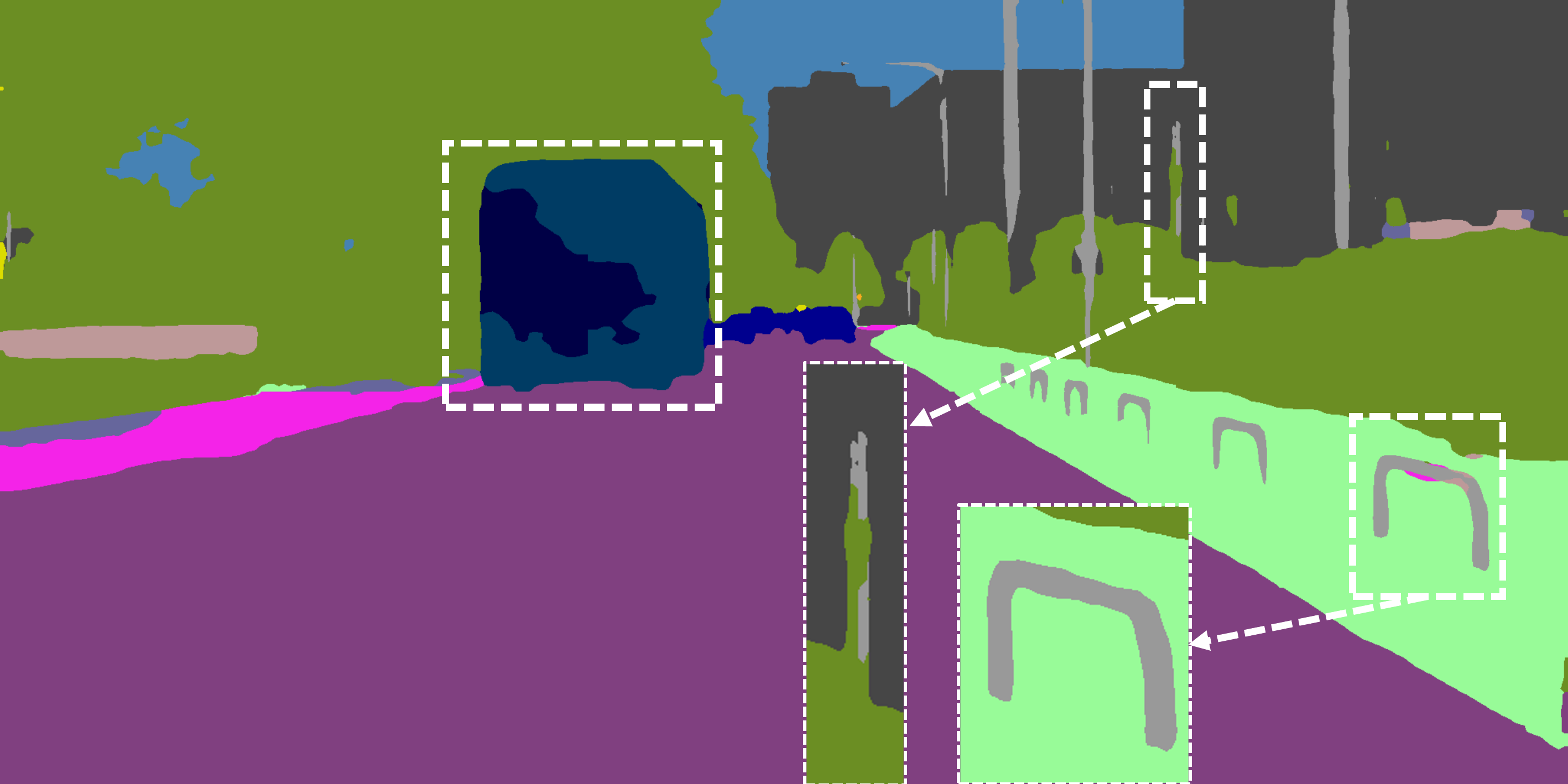}} \hfill
    \subfloat[]{\includegraphics[width=0.142\linewidth]{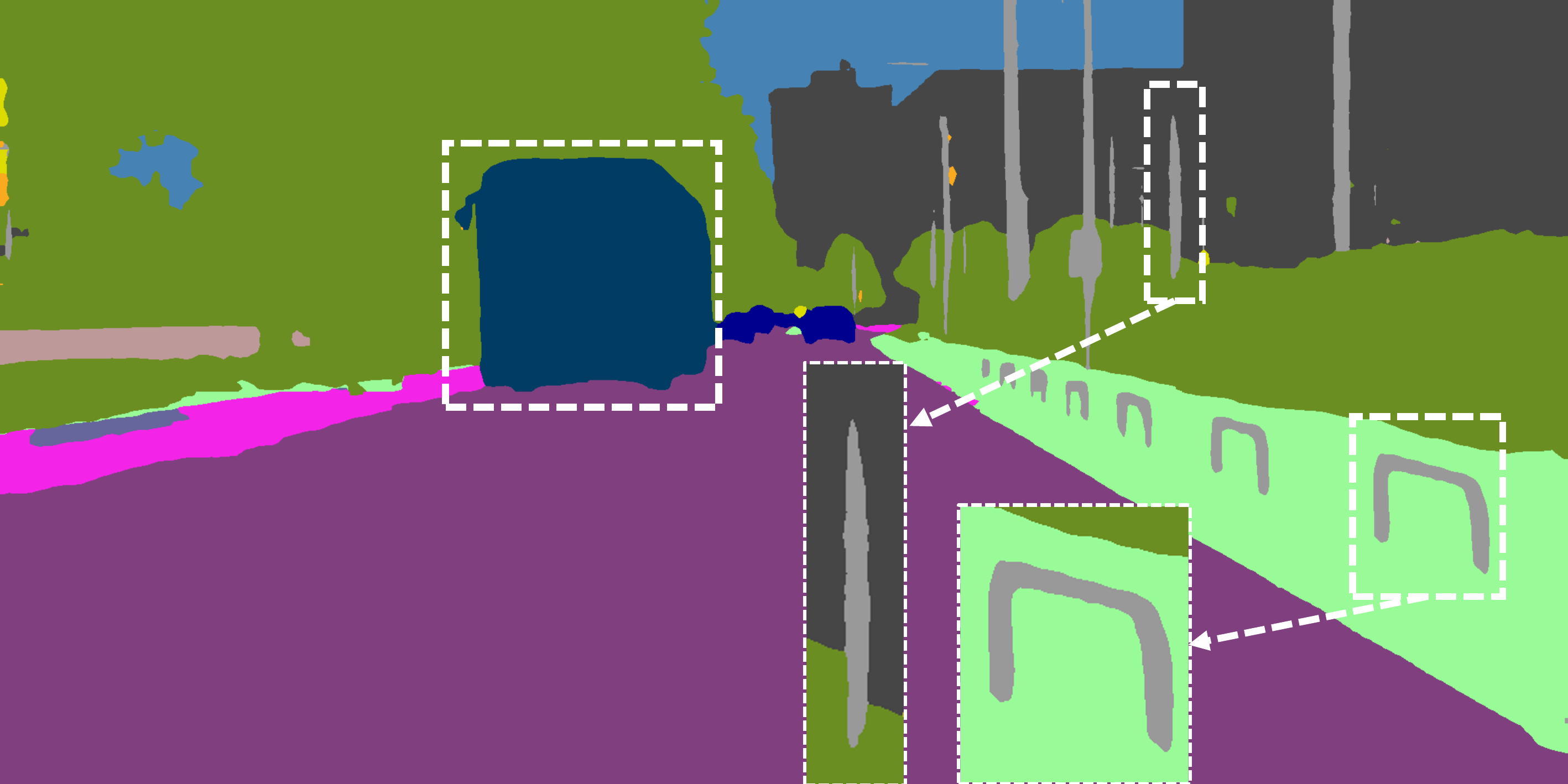}} \hfill
    \subfloat[]{\includegraphics[width=0.142\linewidth]{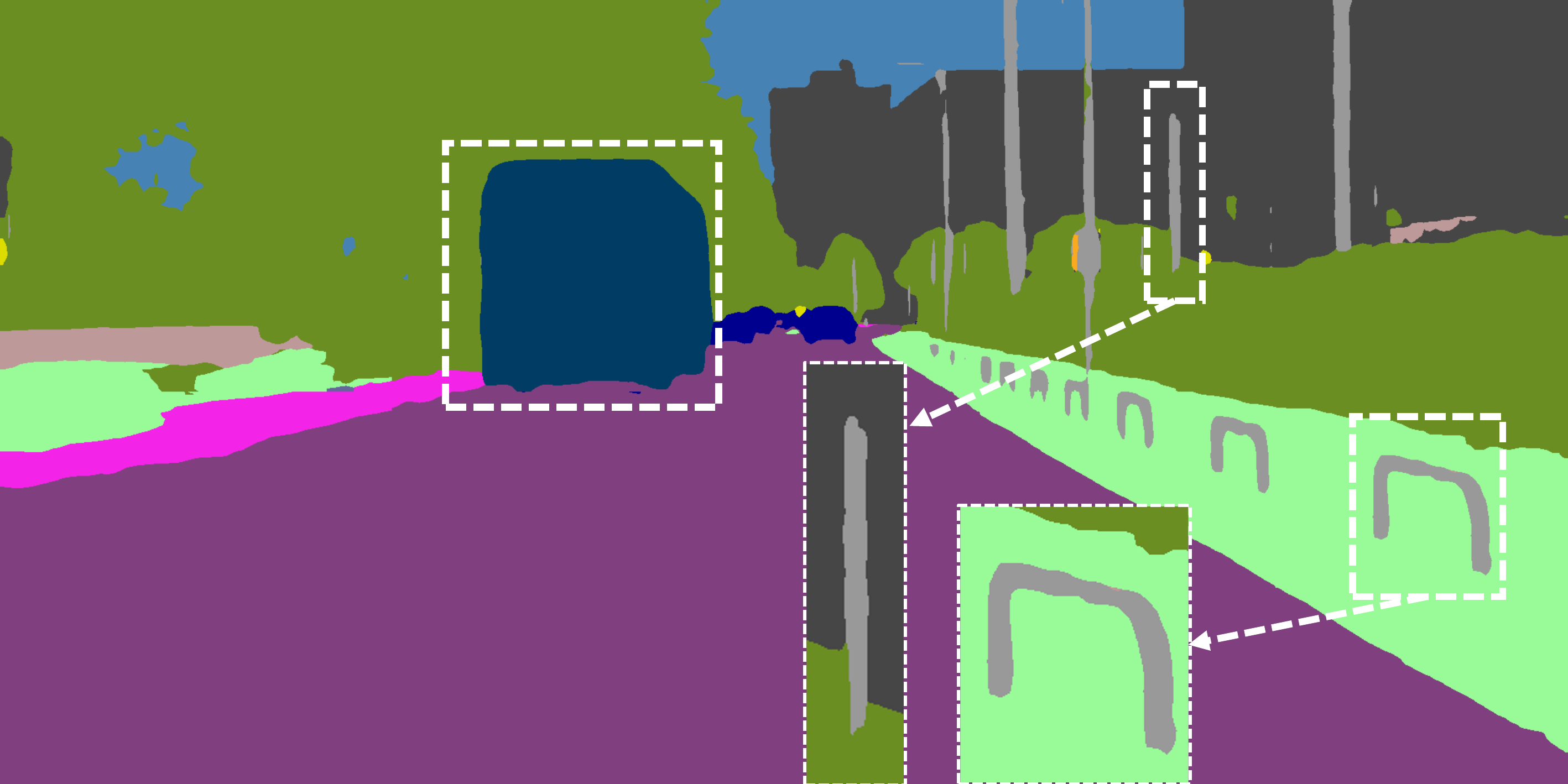}} \\ \vspace{-0.75cm}
    \subfloat[]{\includegraphics[width=0.142\linewidth]{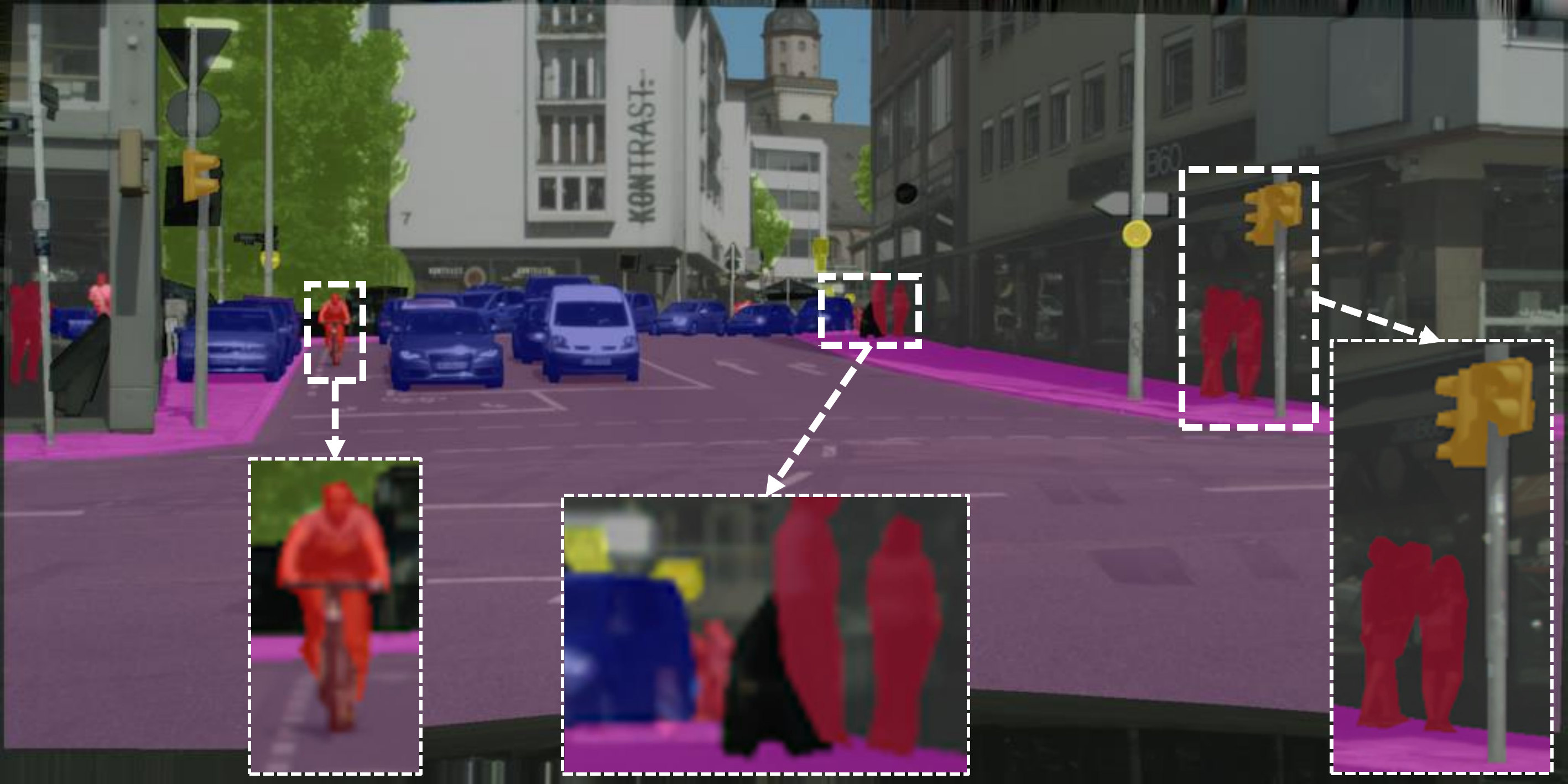}} \hfill
    \subfloat[]{\includegraphics[width=0.142\linewidth]{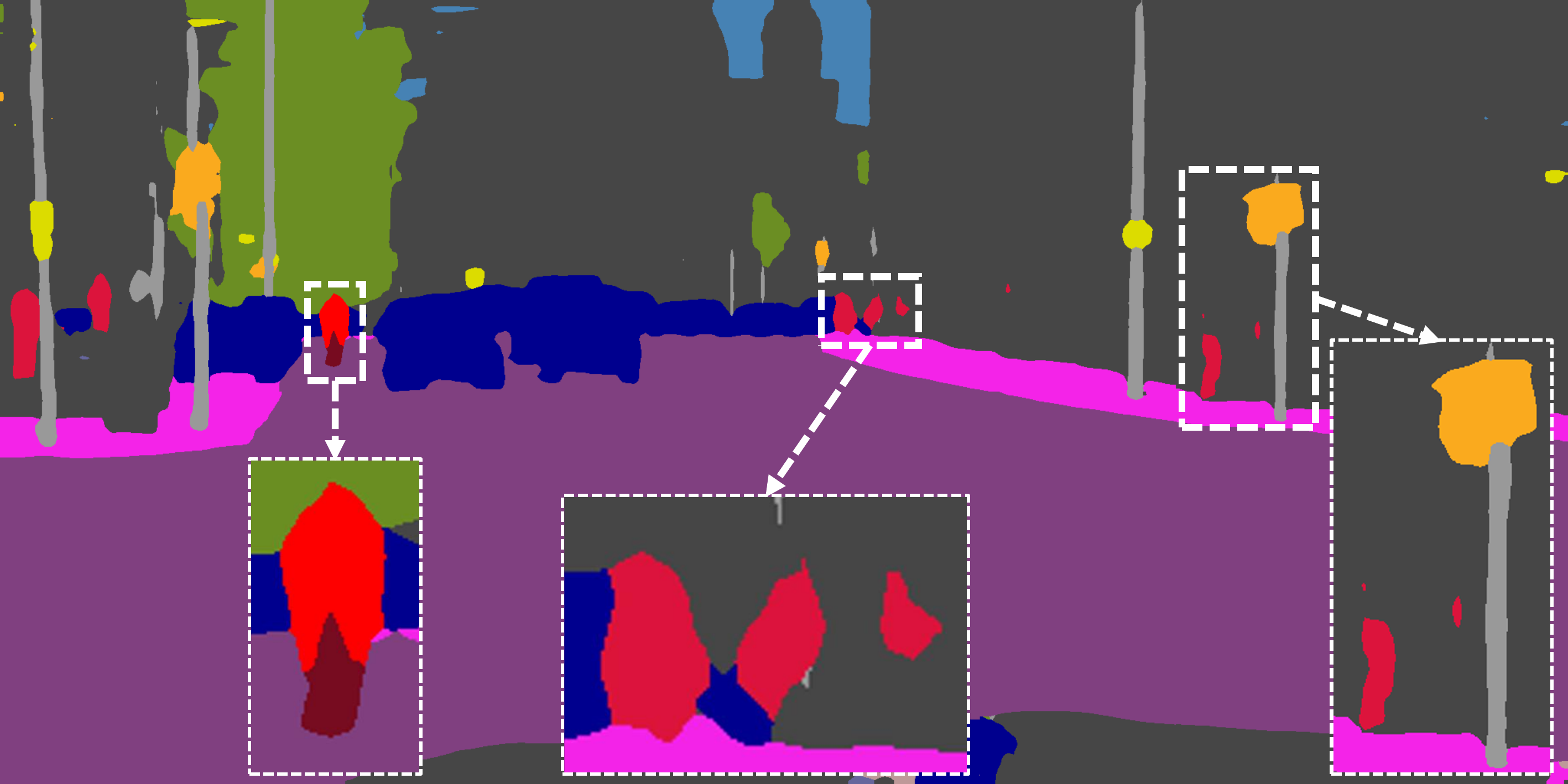}} \hfill
    \subfloat[]{\includegraphics[width=0.142\linewidth]{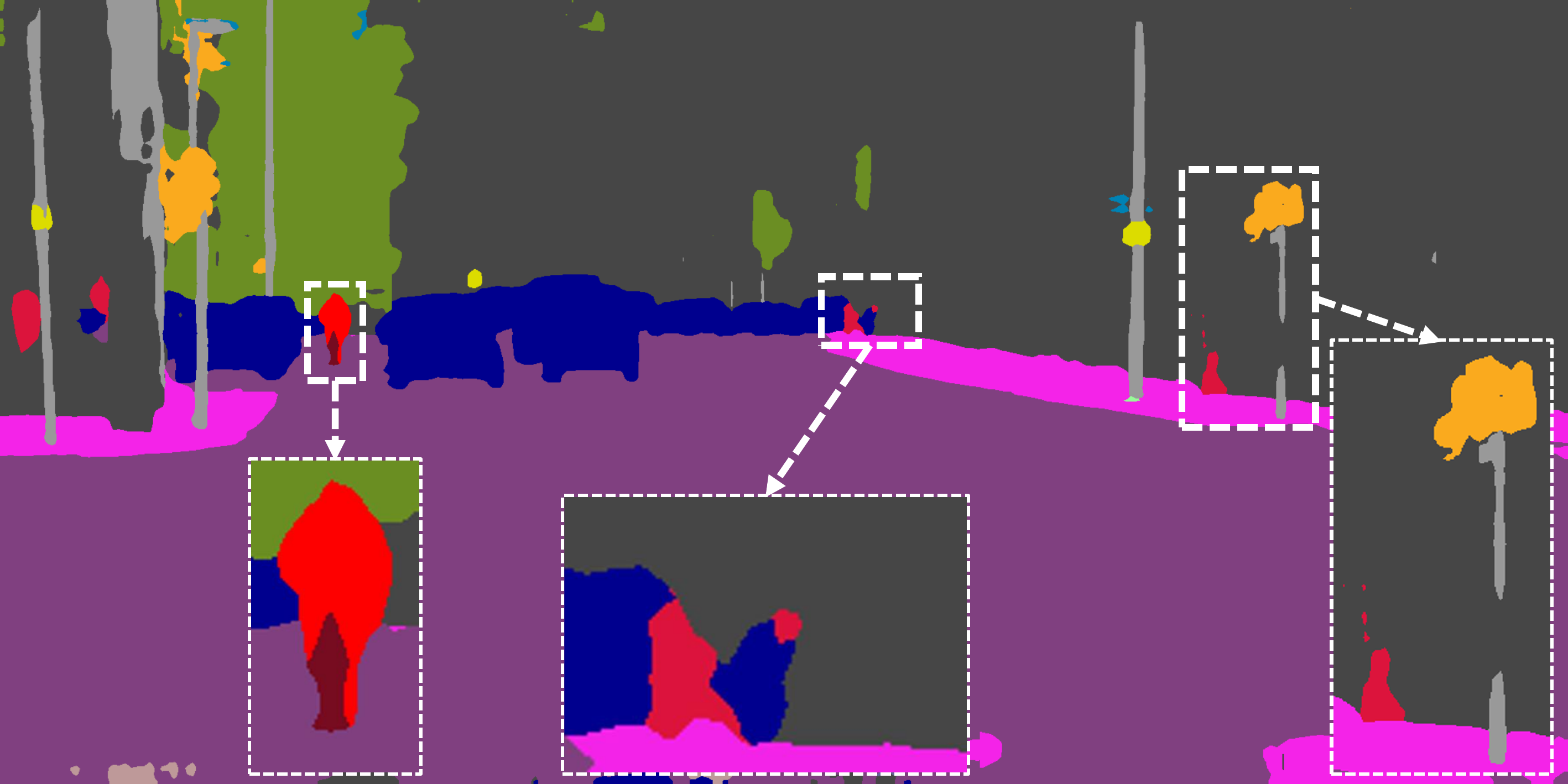}} \hfill
    \subfloat[]{\includegraphics[width=0.142\linewidth]{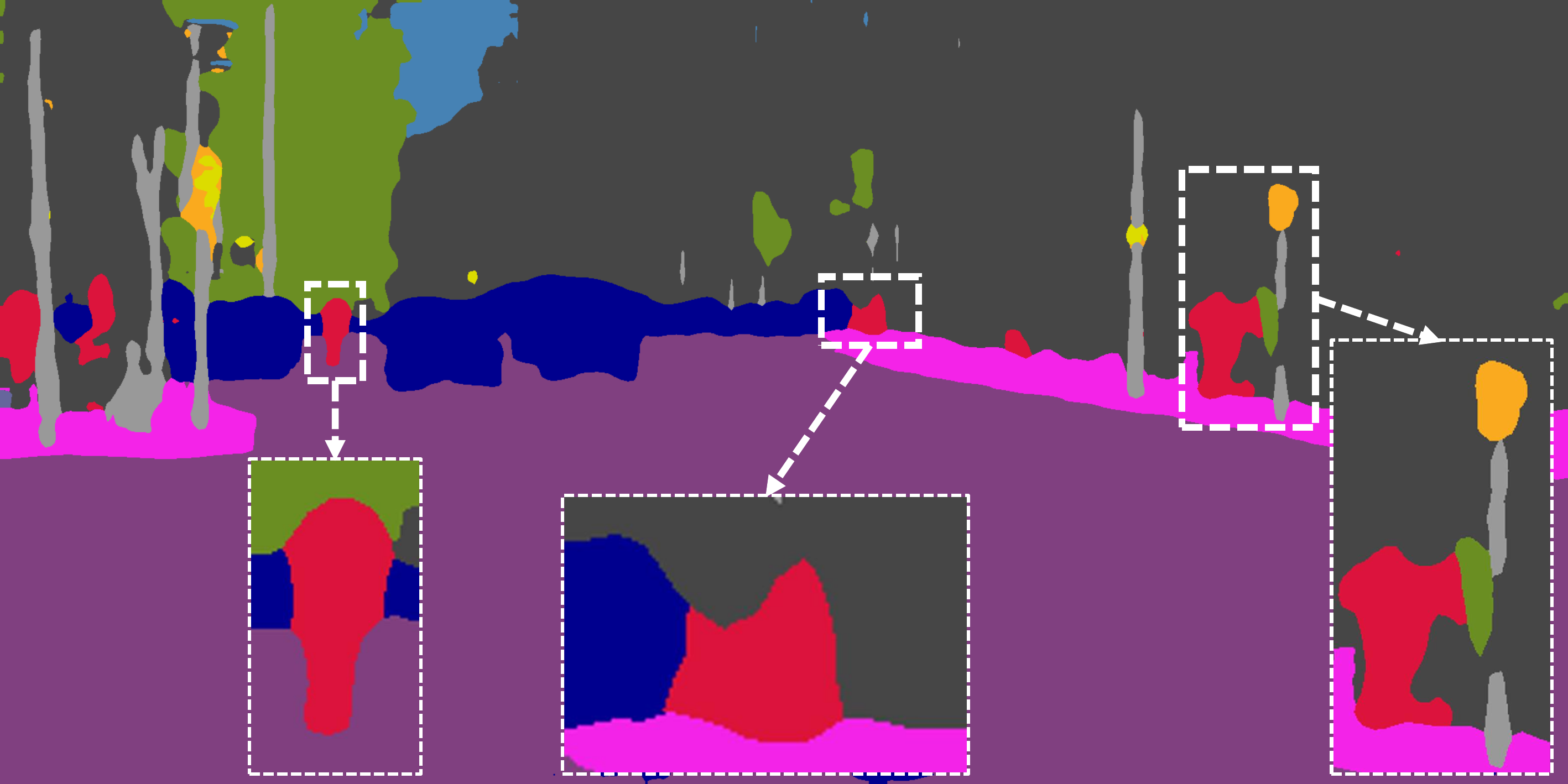}} \hfill
    \subfloat[]{\includegraphics[width=0.142\linewidth]{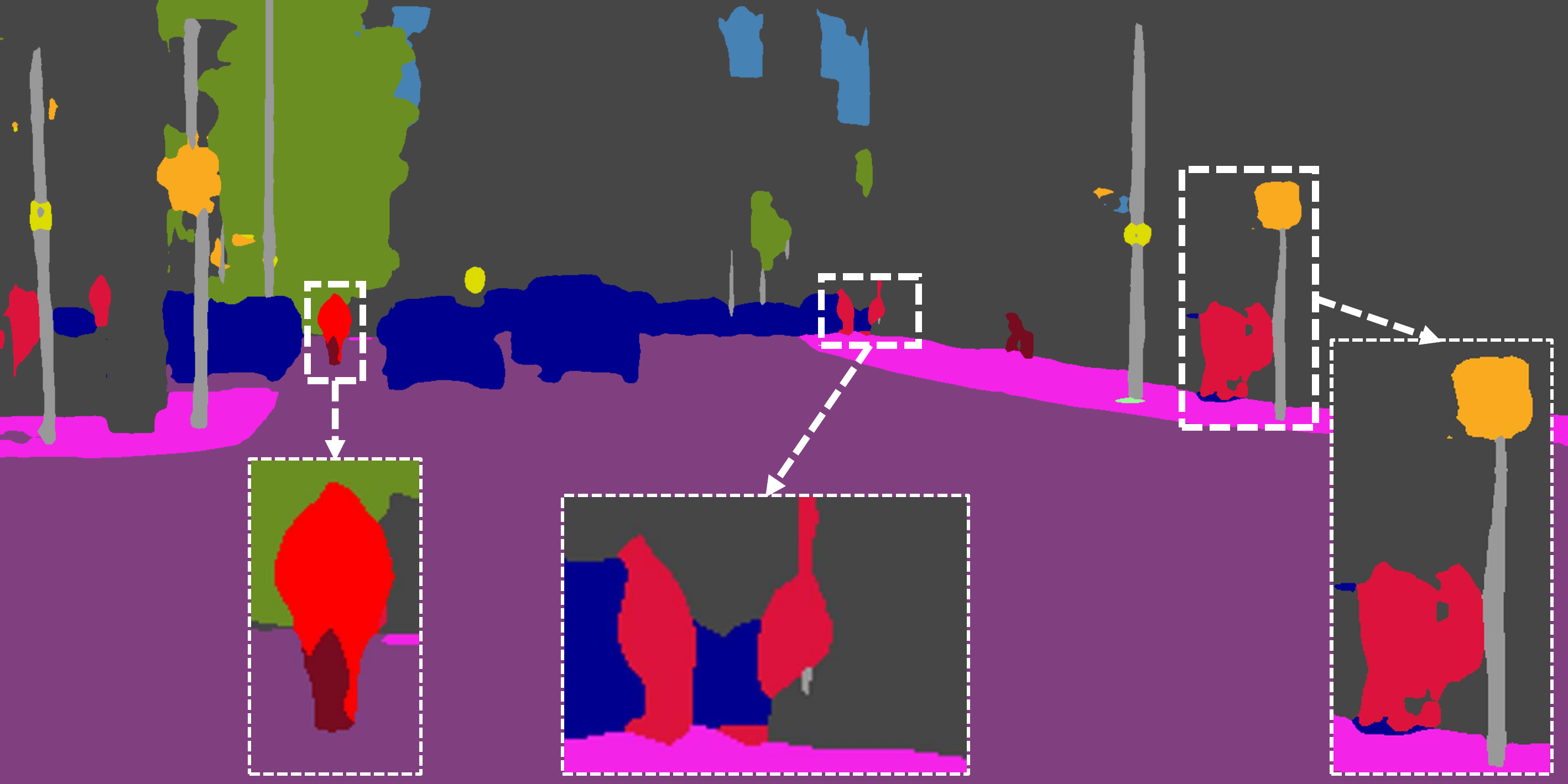}} \hfill
    \subfloat[]{\includegraphics[width=0.142\linewidth]{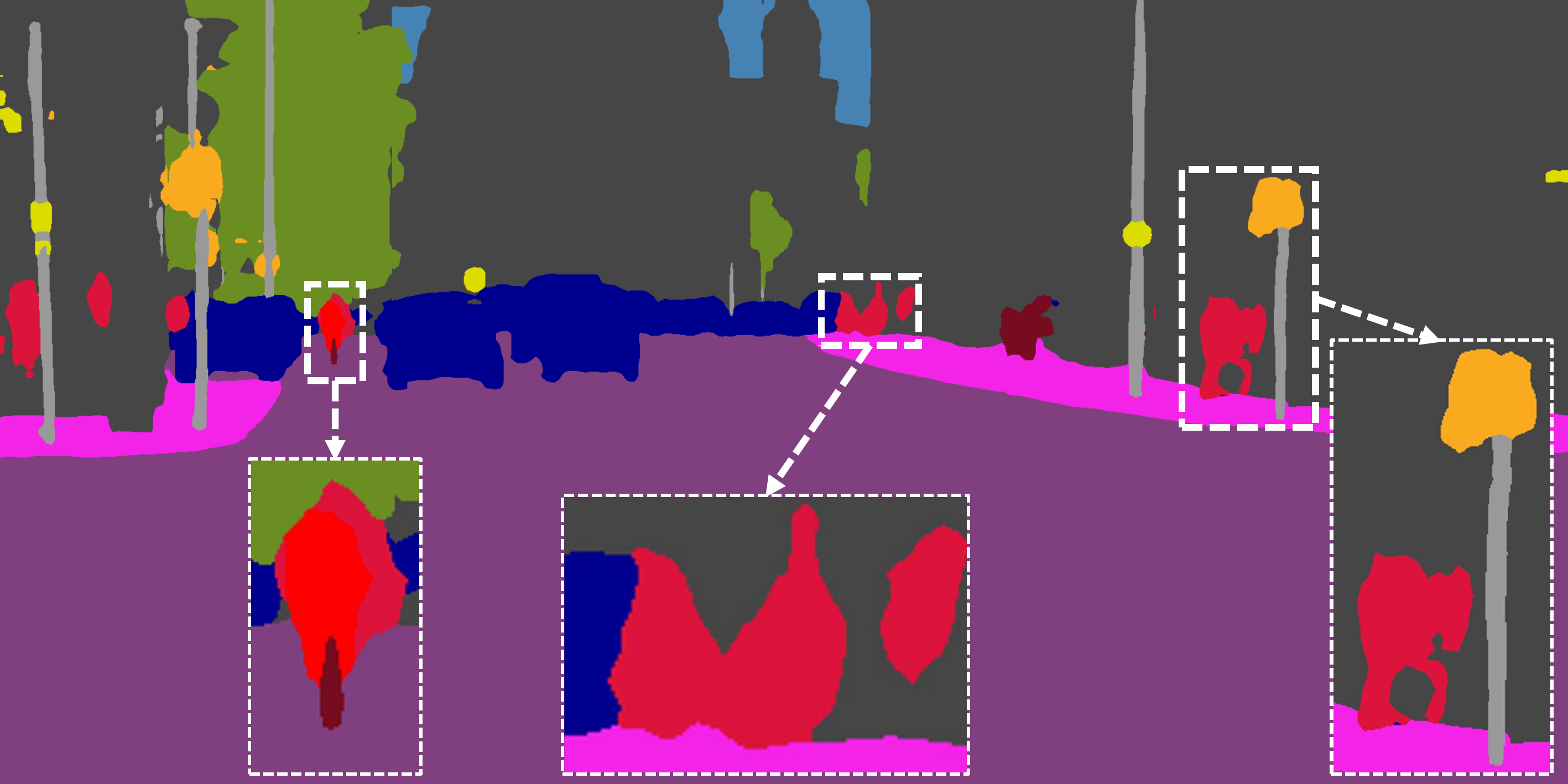}} \hfill
    \subfloat[]{\includegraphics[width=0.142\linewidth]{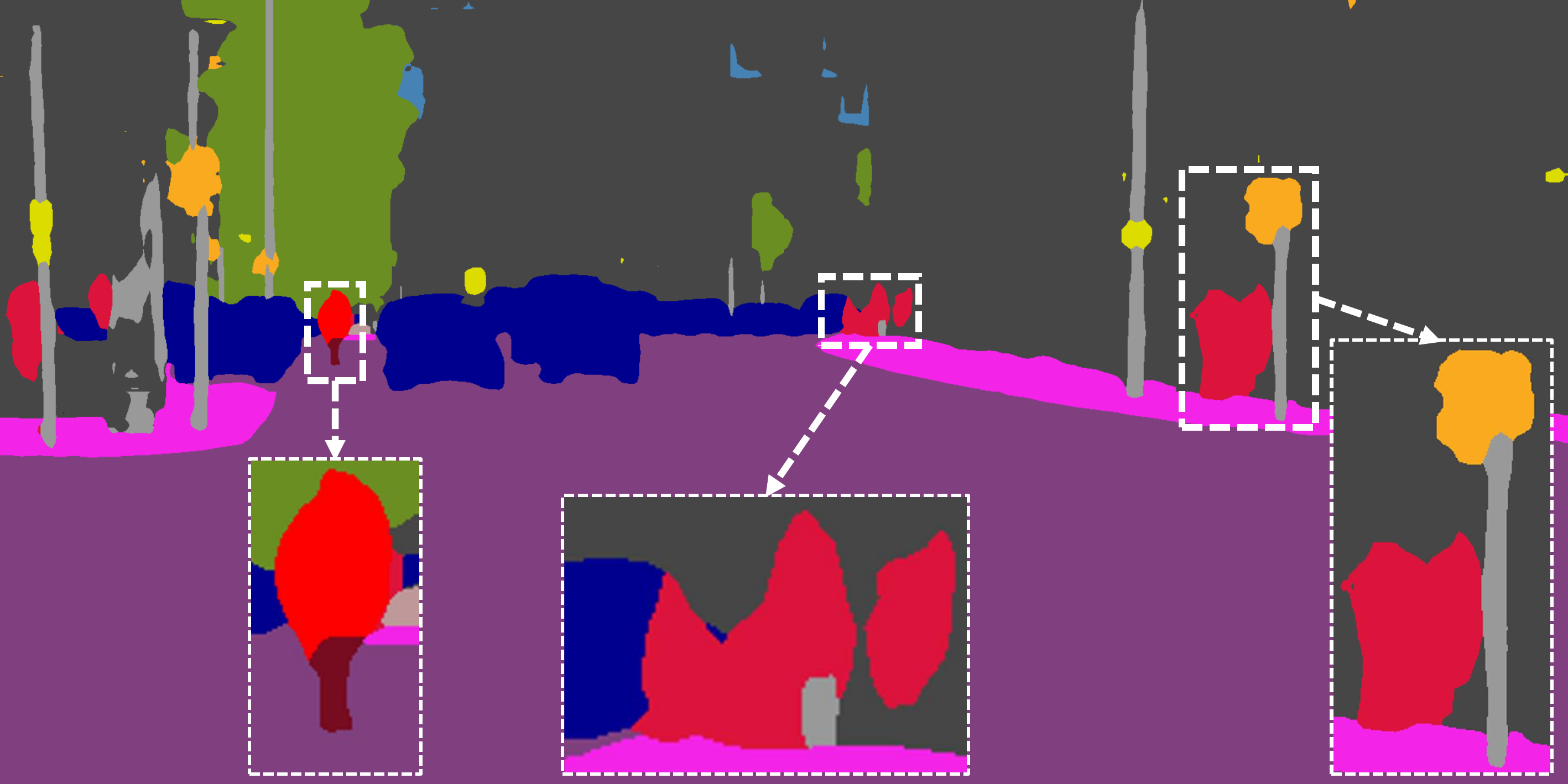}} \\ \vspace{-0.75cm}
    \subfloat[]{\includegraphics[width=0.142\linewidth]{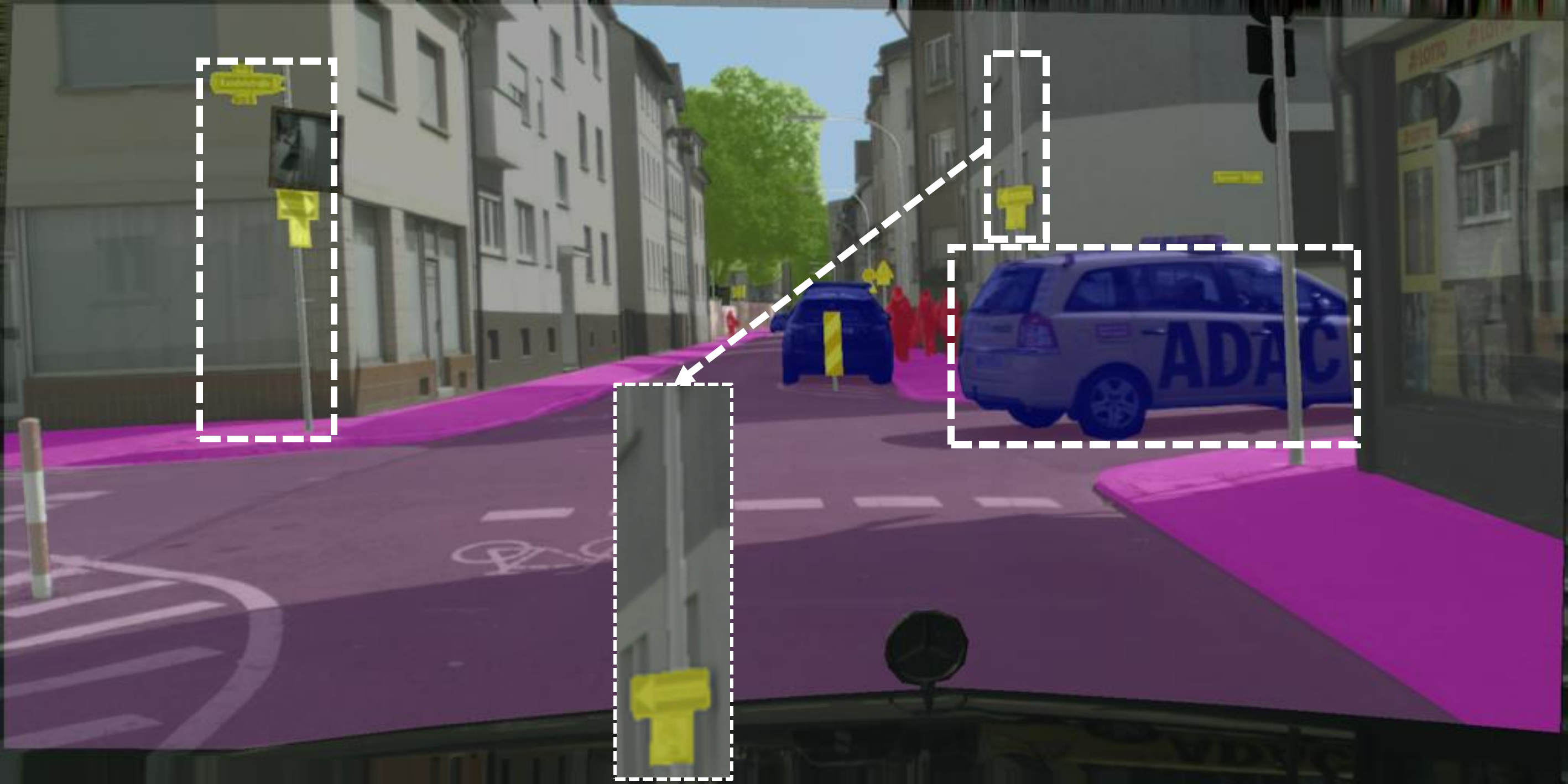}} \hfill
    \subfloat[]{\includegraphics[width=0.142\linewidth]{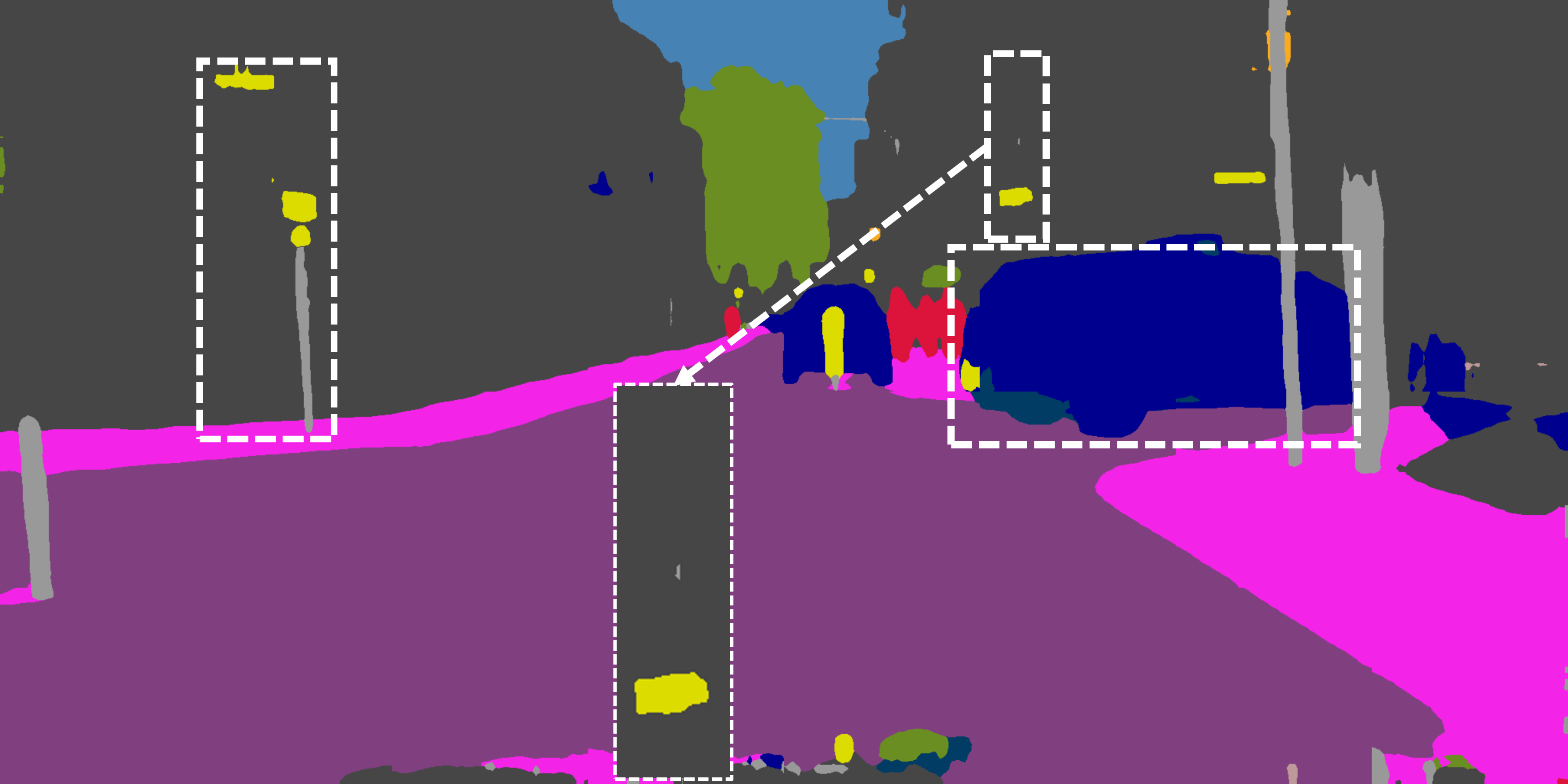}} \hfill
    \subfloat[]{\includegraphics[width=0.142\linewidth]{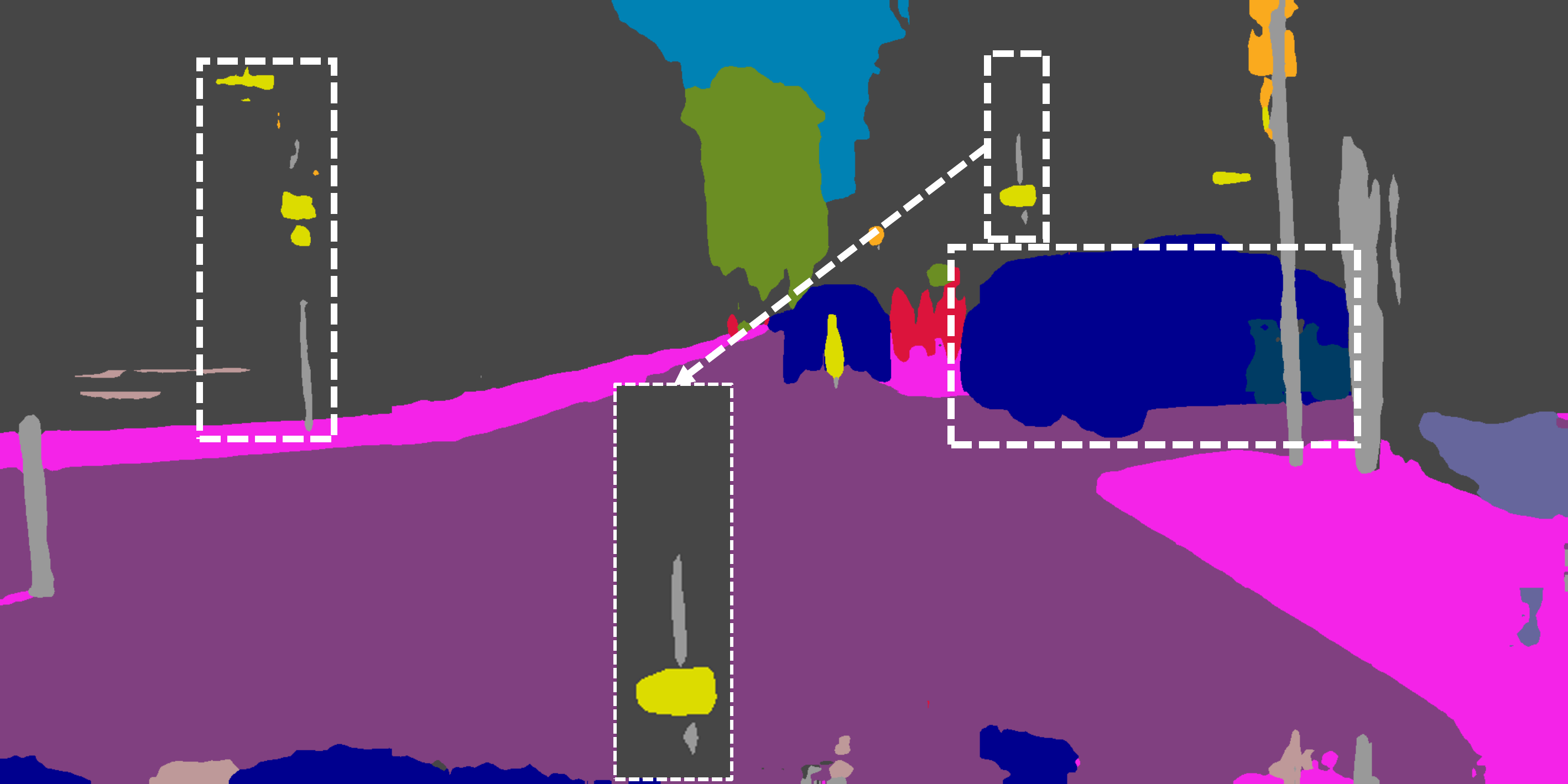}} \hfill
    \subfloat[]{\includegraphics[width=0.142\linewidth]{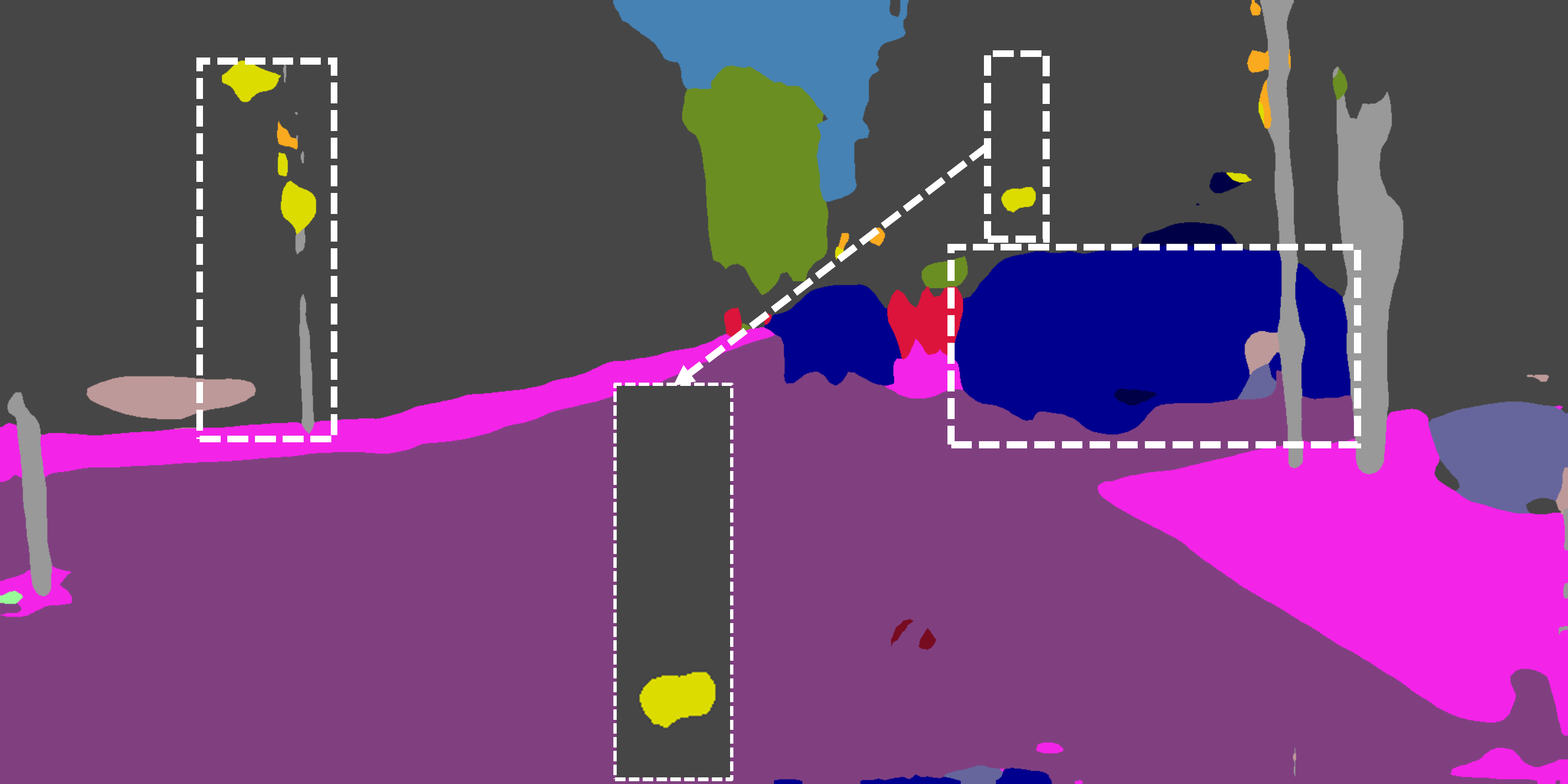}} \hfill
    \subfloat[]{\includegraphics[width=0.142\linewidth]{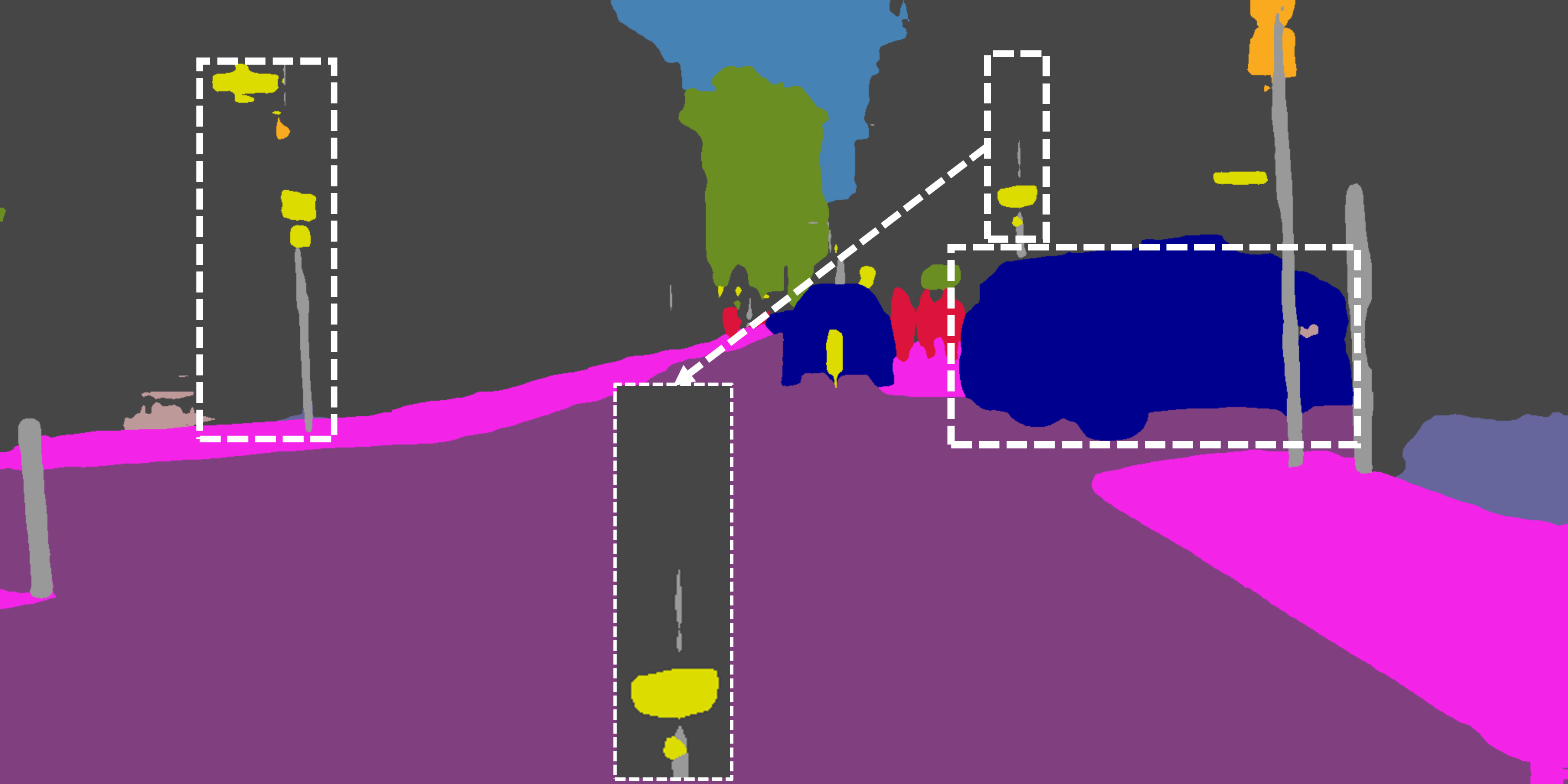}} \hfill
    \subfloat[]{\includegraphics[width=0.142\linewidth]{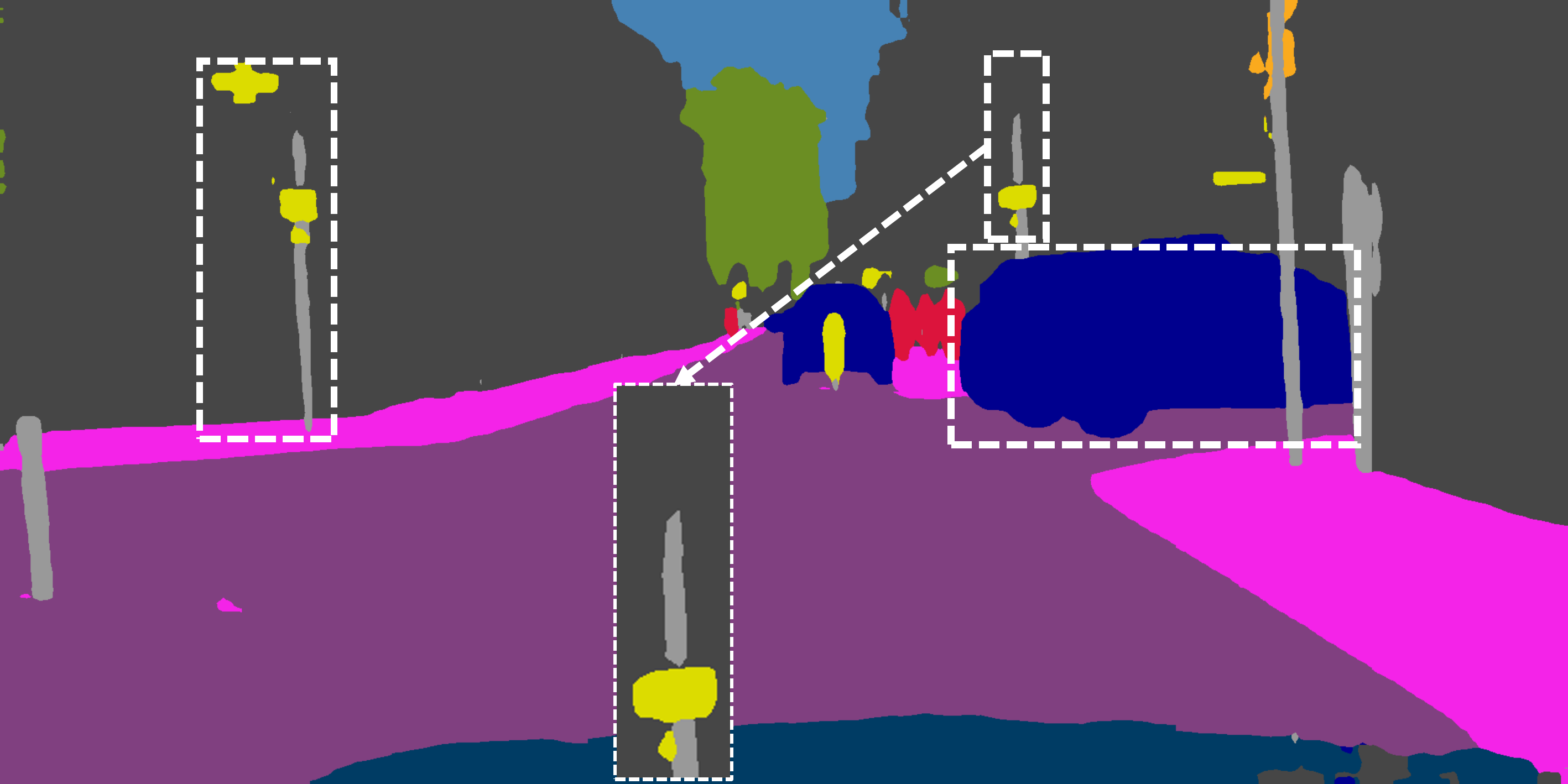}} \hfill
    \subfloat[]{\includegraphics[width=0.142\linewidth]{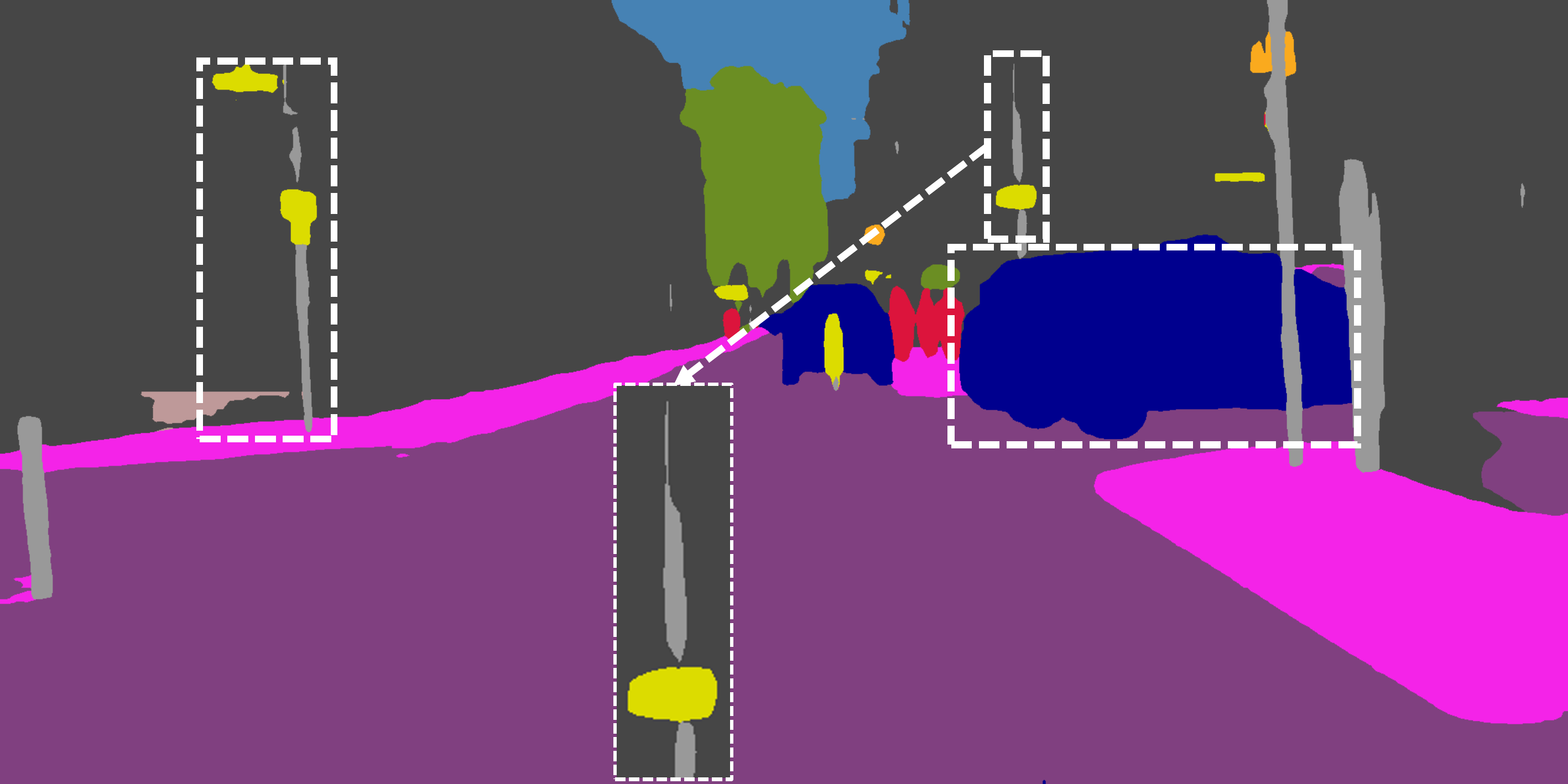}} \\ \vspace{-0.75cm}
    \subfloat[]{\includegraphics[width=0.142\linewidth]{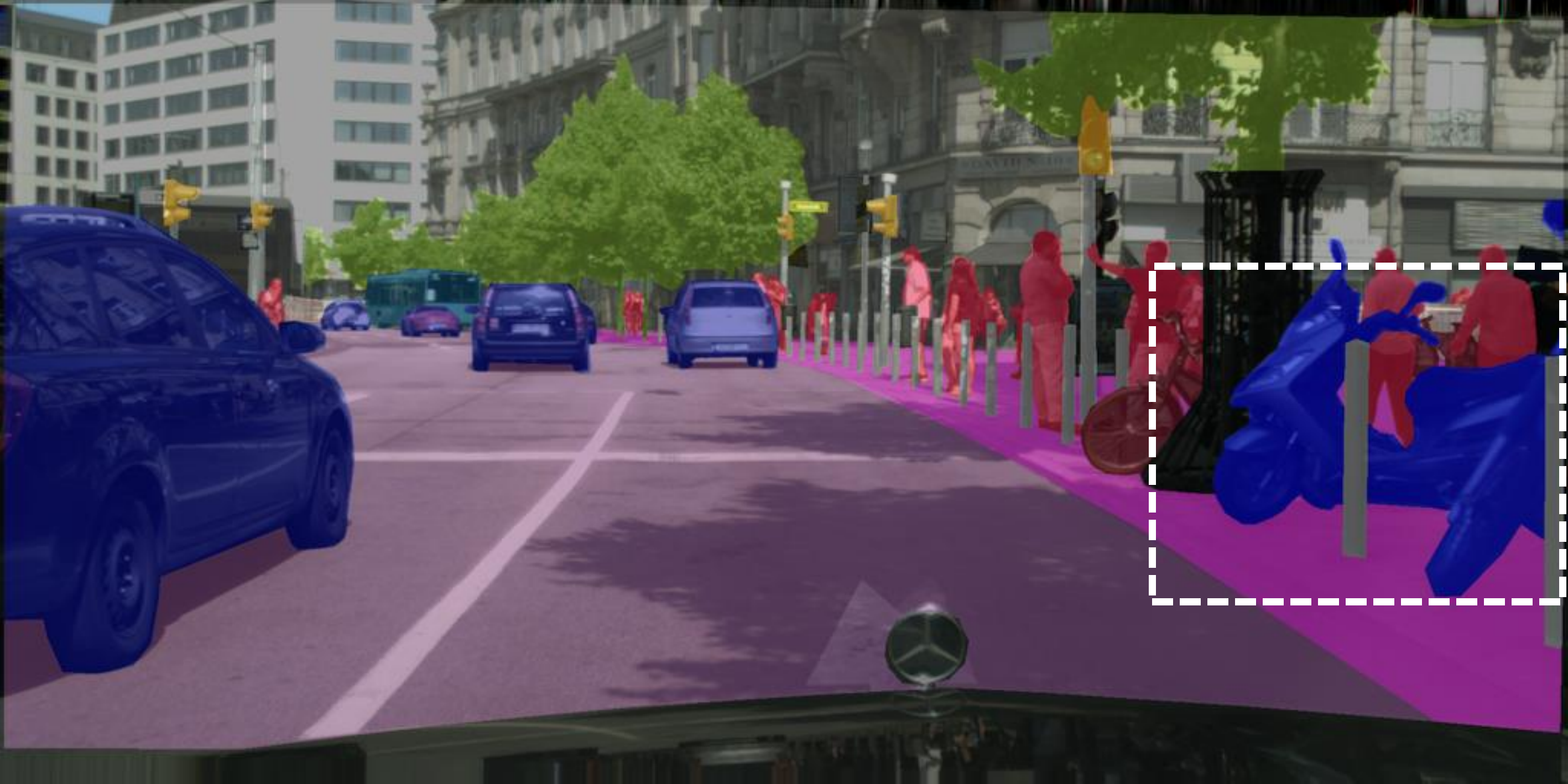}} \hfill
    \subfloat[]{\includegraphics[width=0.142\linewidth]{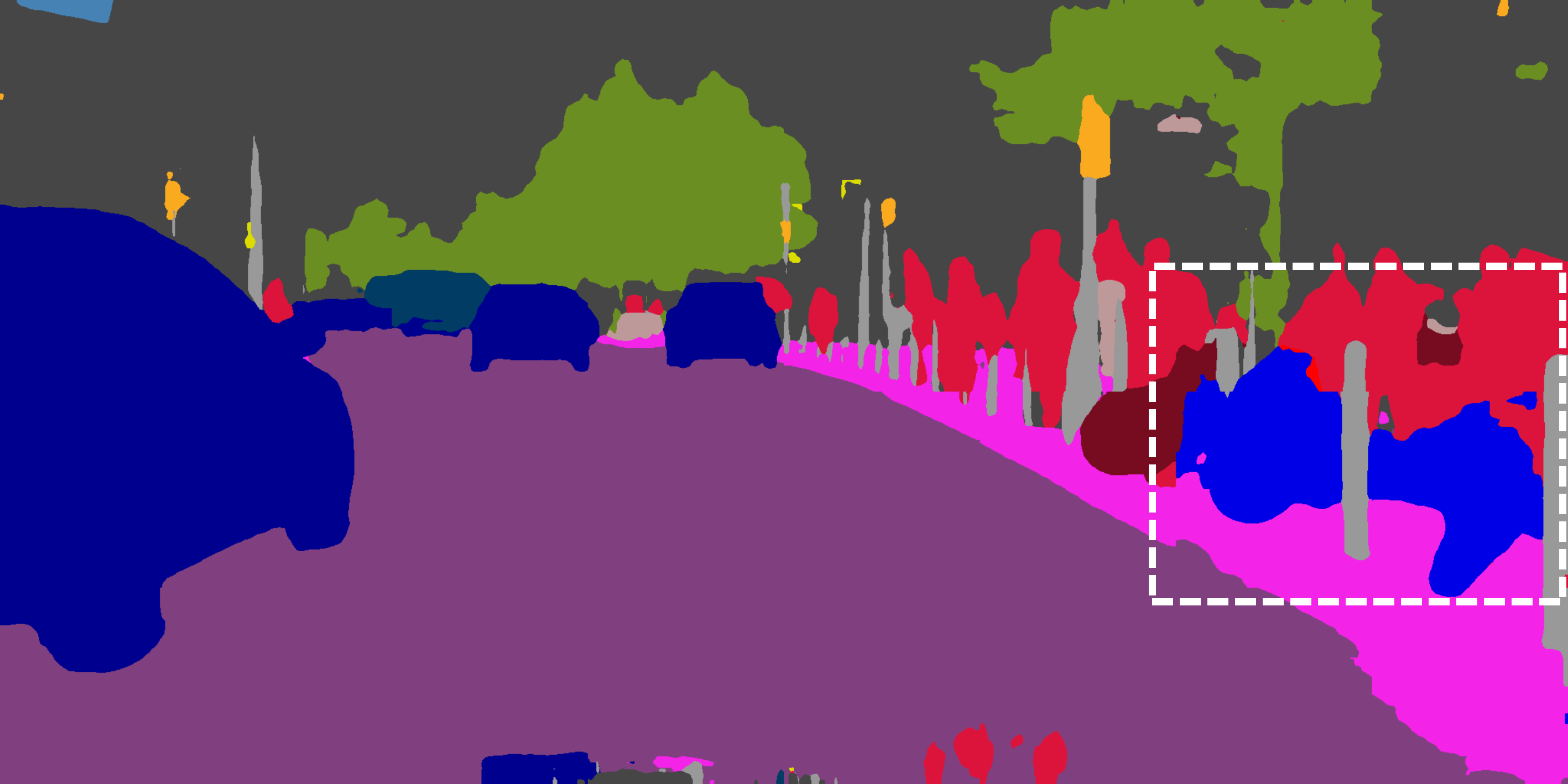}} \hfill
    \subfloat[]{\includegraphics[width=0.142\linewidth]{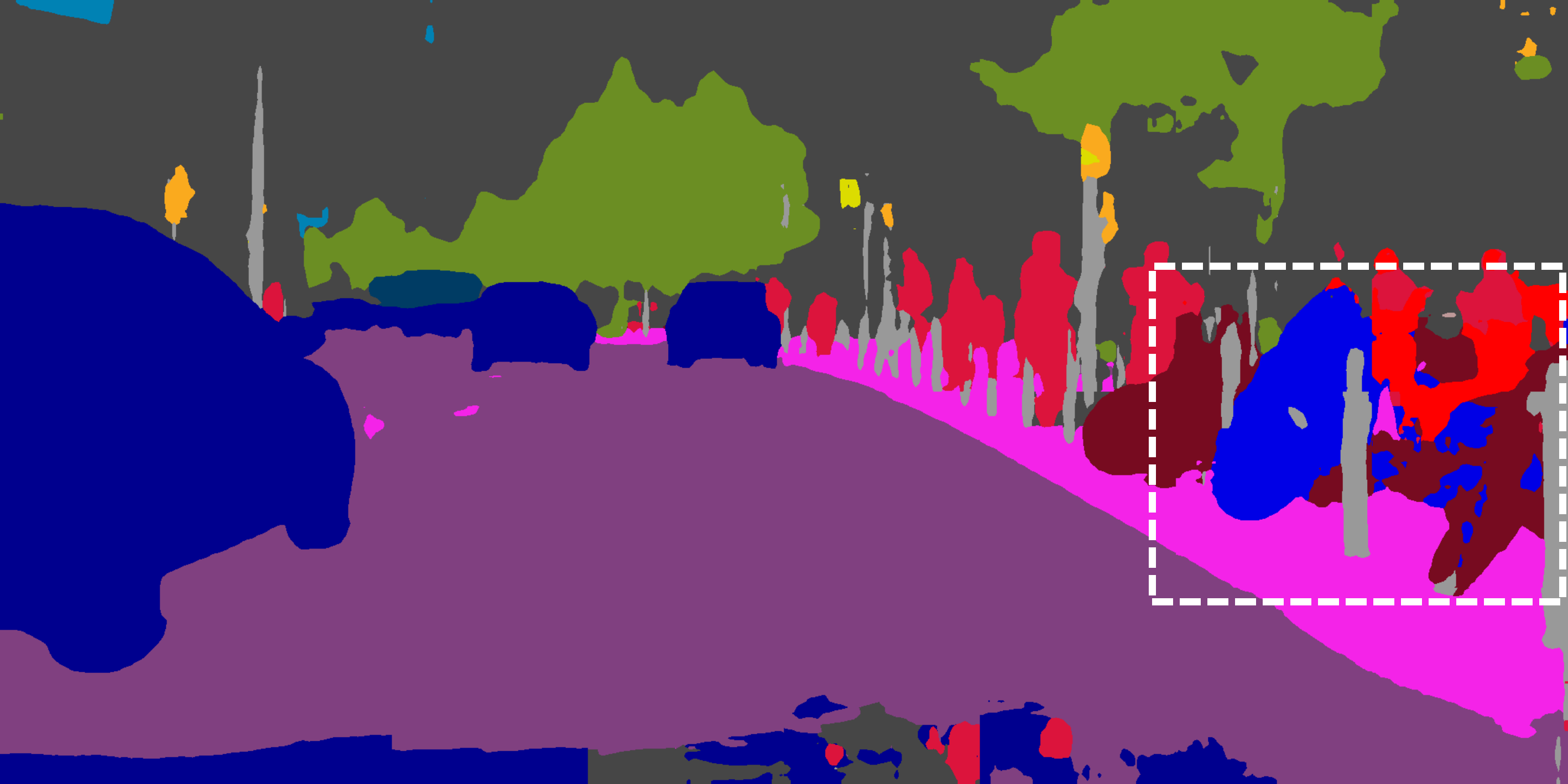}} \hfill
    \subfloat[]{\includegraphics[width=0.142\linewidth]{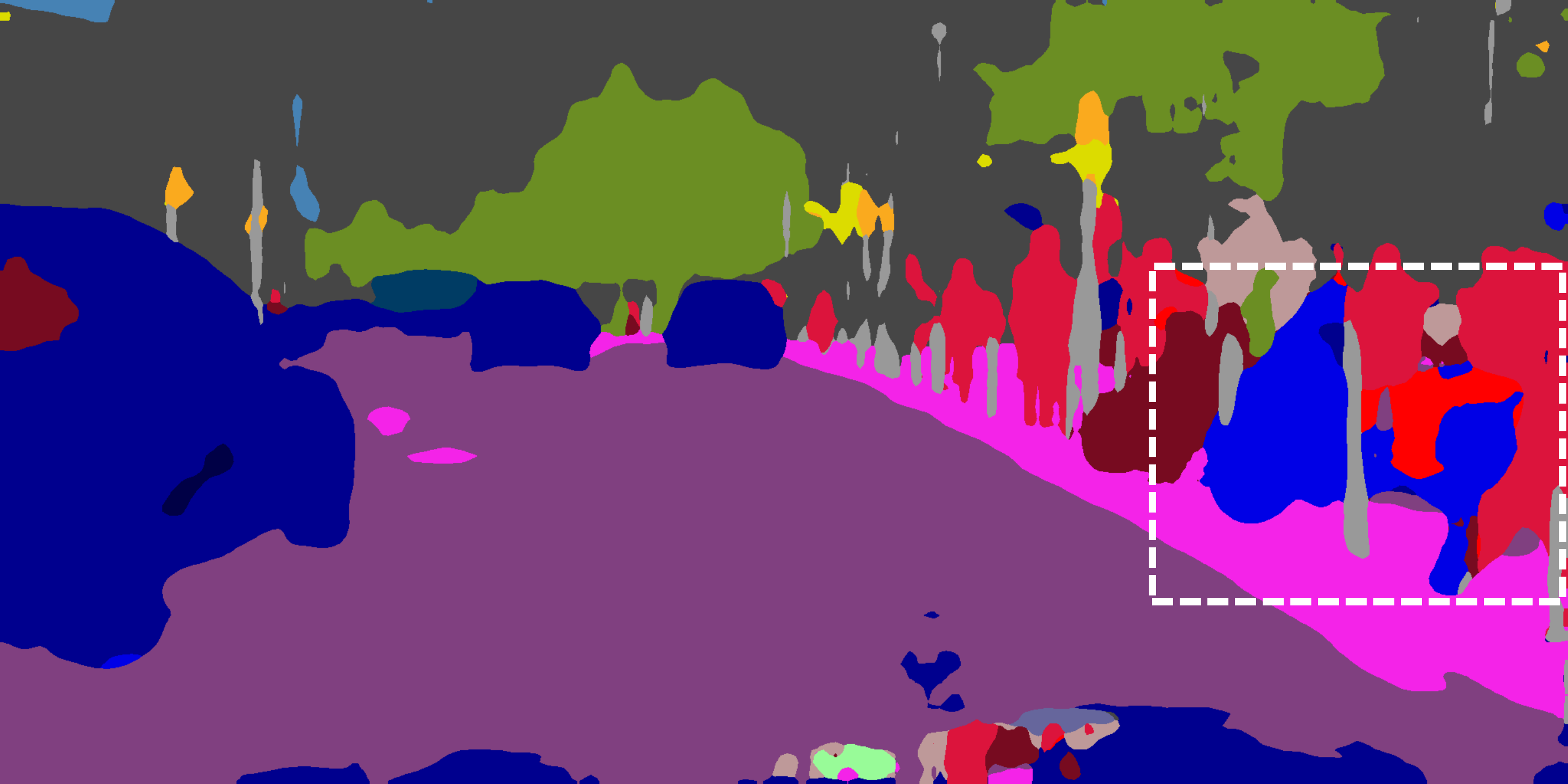}} \hfill
    \subfloat[]{\includegraphics[width=0.142\linewidth]{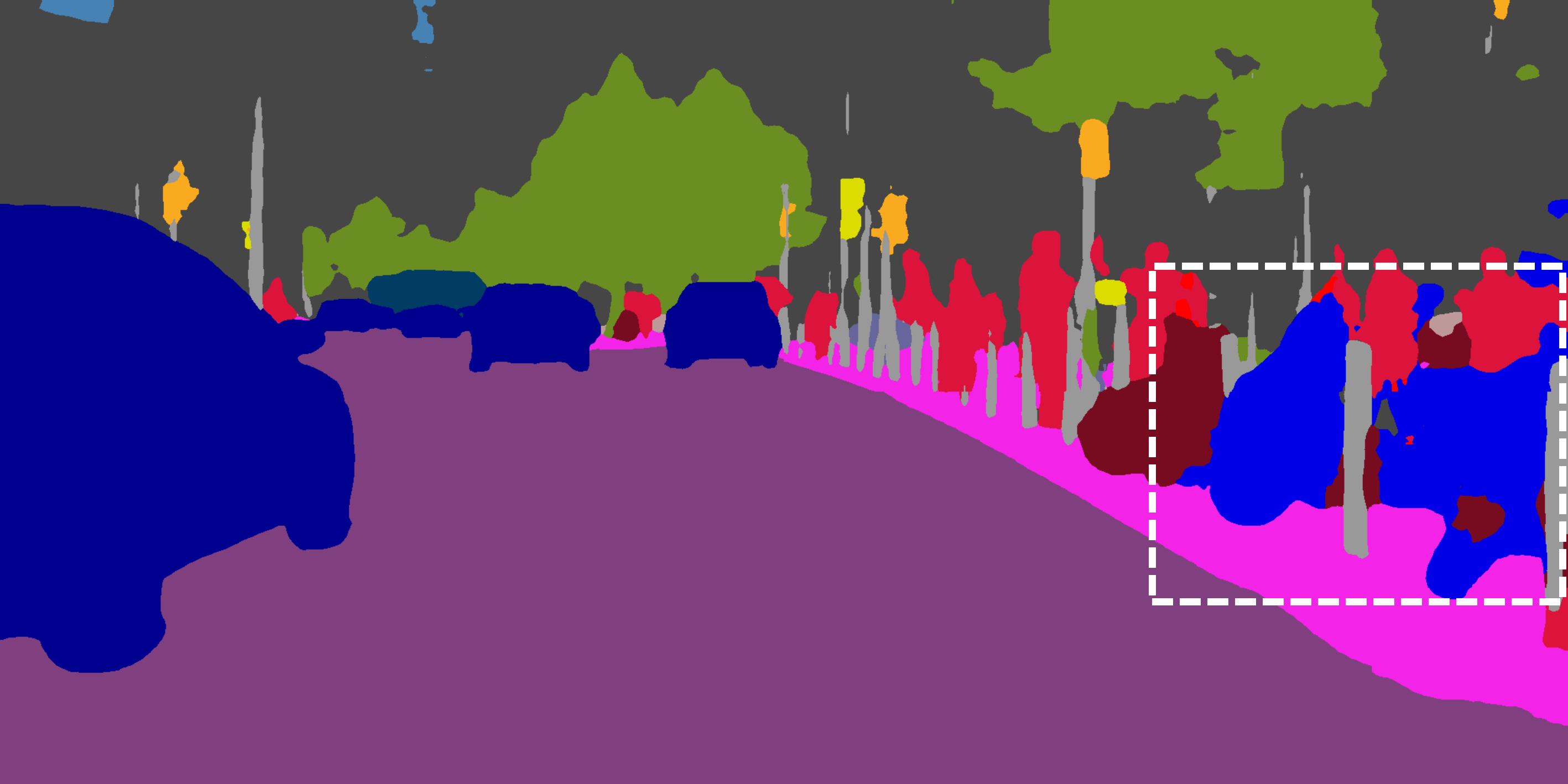}} \hfill
    \subfloat[]{\includegraphics[width=0.142\linewidth]{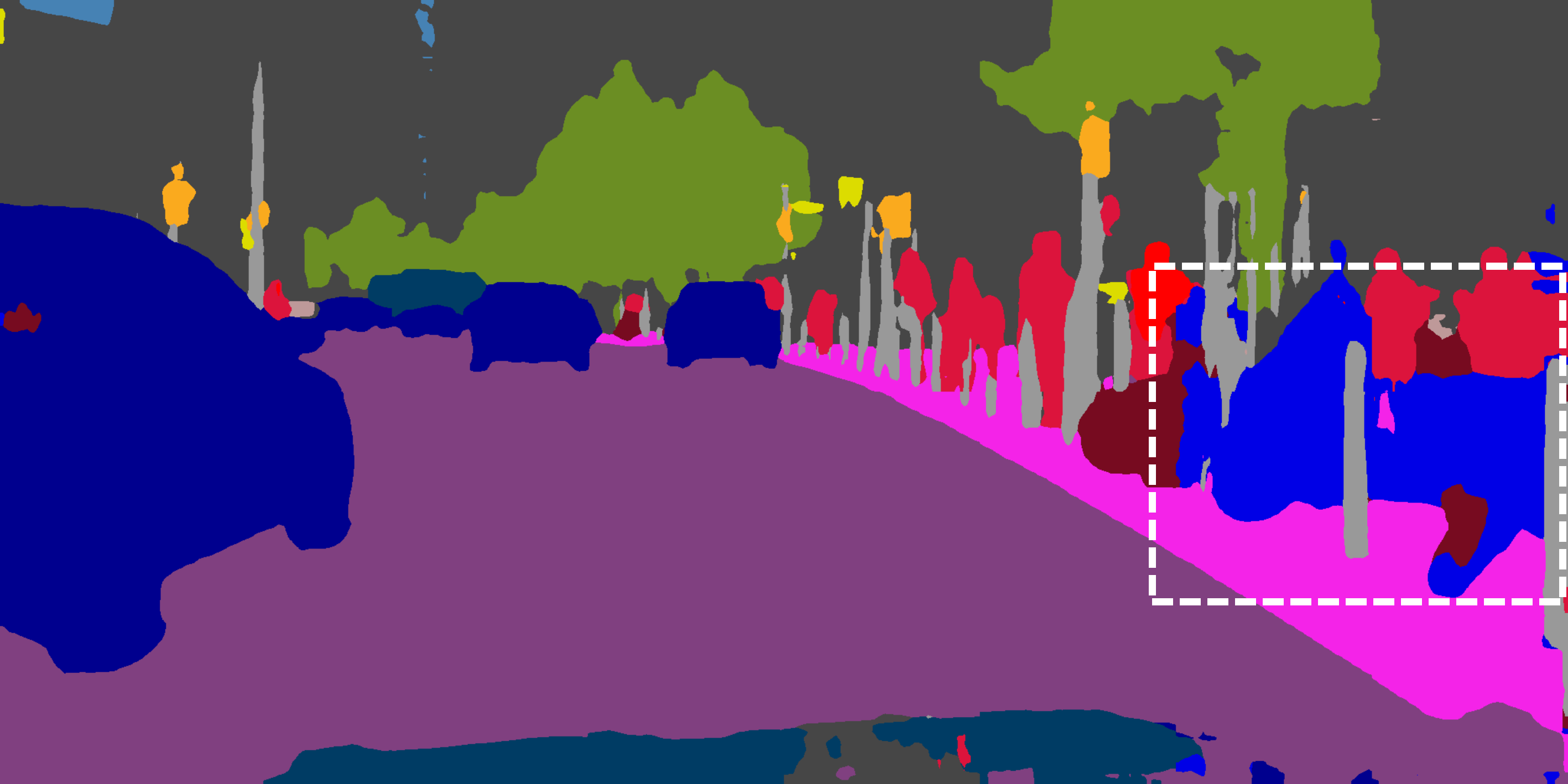}} \hfill
    \subfloat[]{\includegraphics[width=0.142\linewidth]{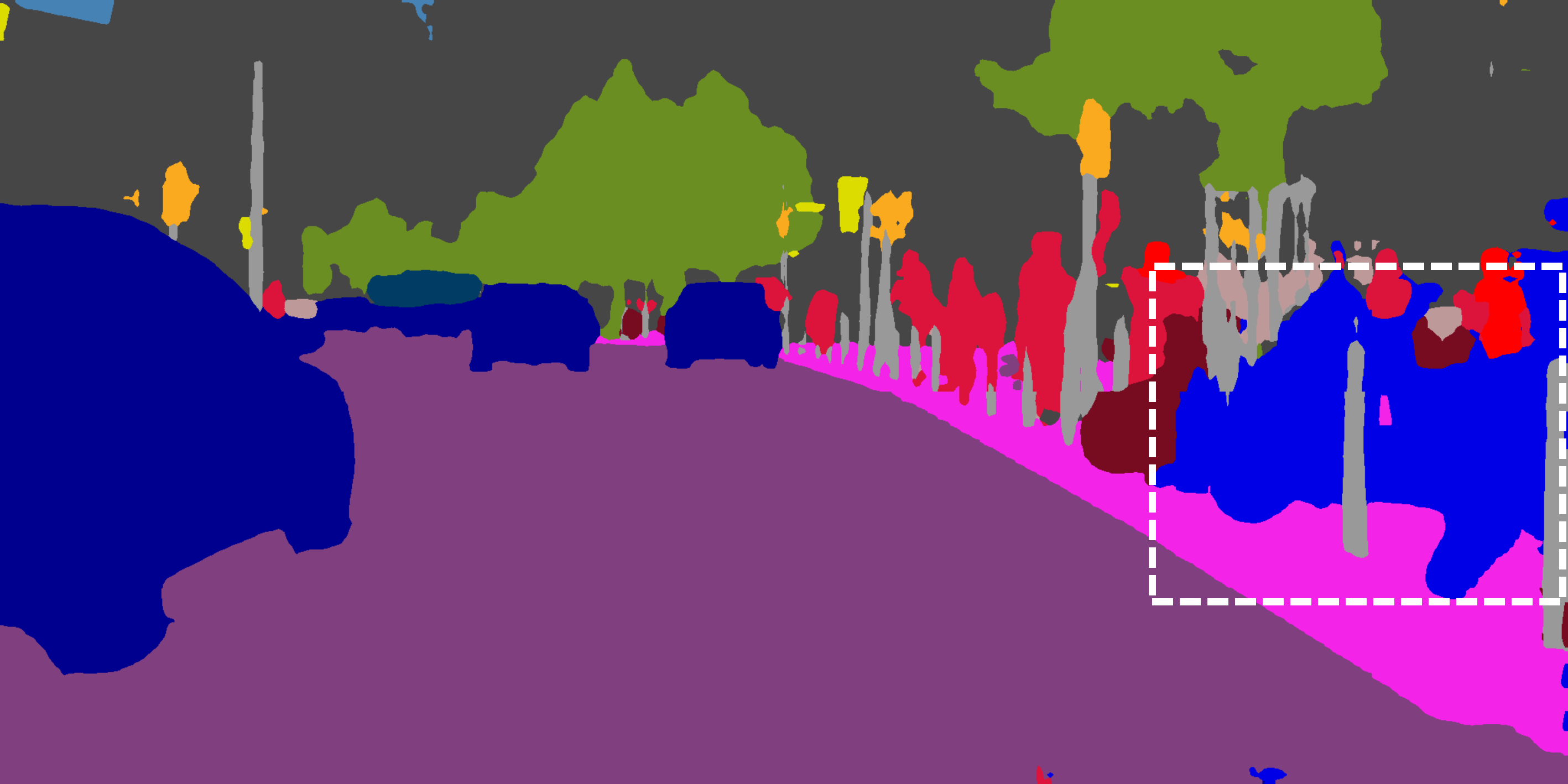}} \\ \vspace{-0.75cm}

    \subfloat[]{\includegraphics[width=0.142\linewidth]{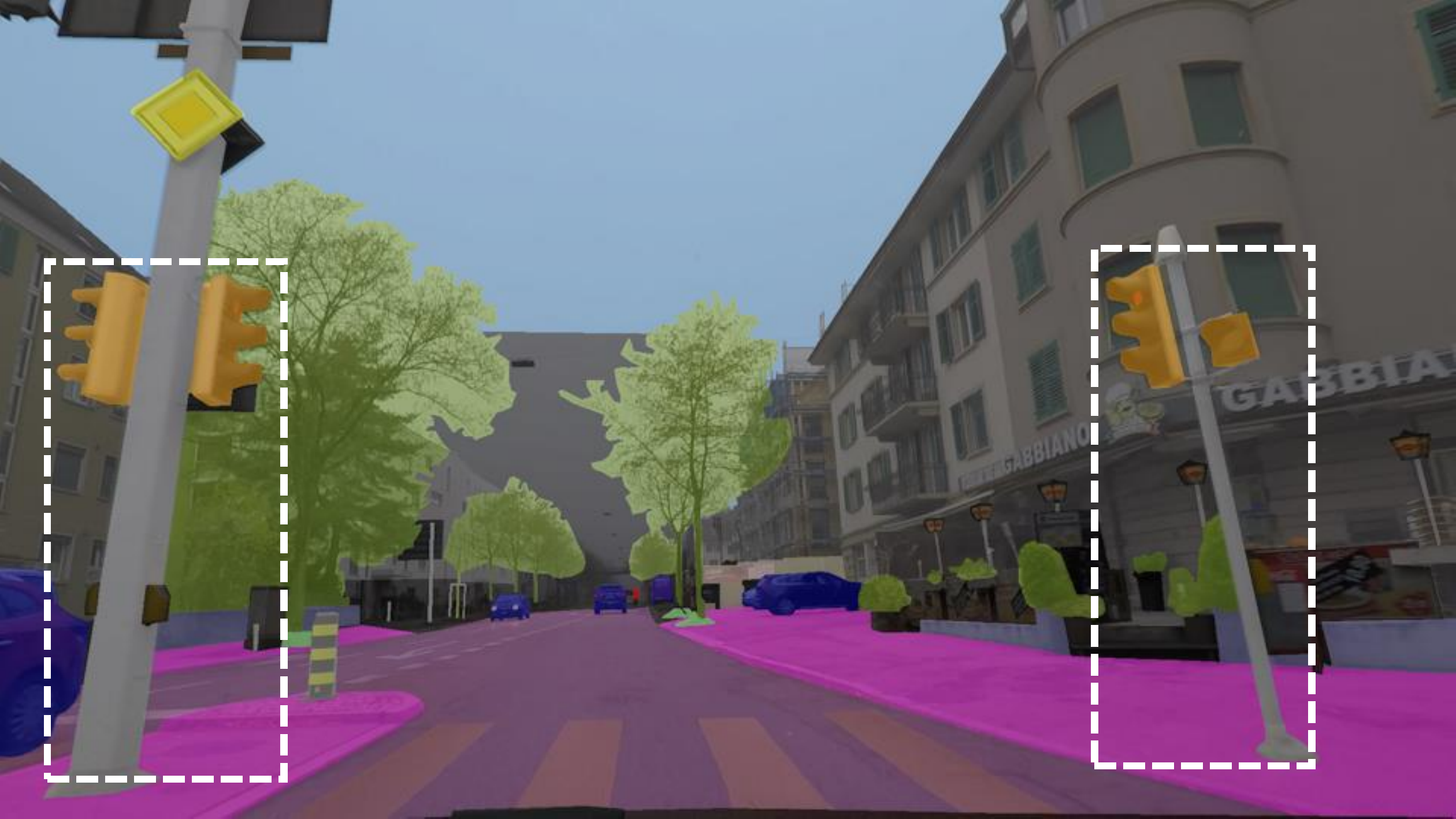}} \hfill
    \subfloat[]{\includegraphics[width=0.142\linewidth]{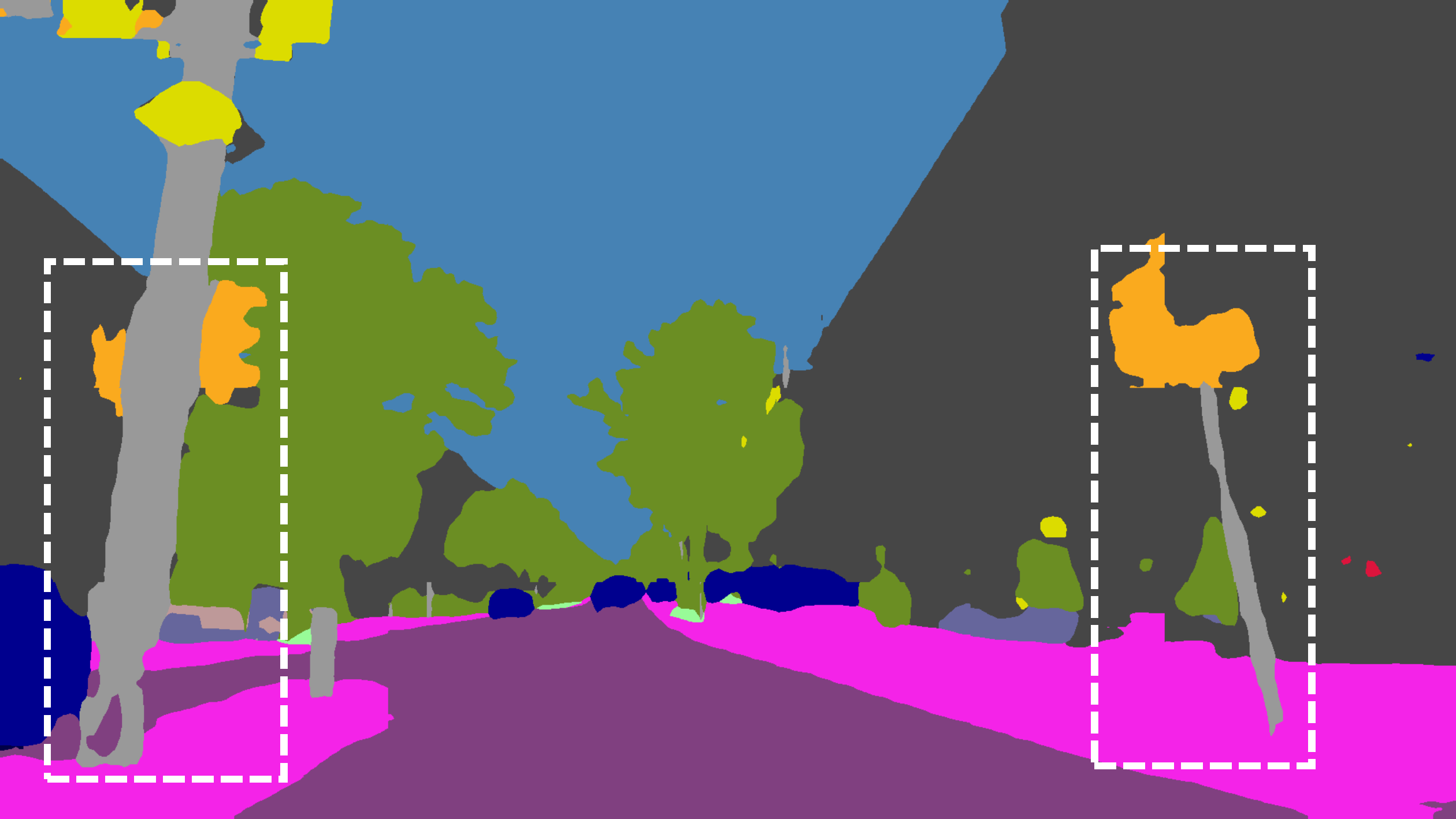}} \hfill
    \subfloat[]{\includegraphics[width=0.142\linewidth]{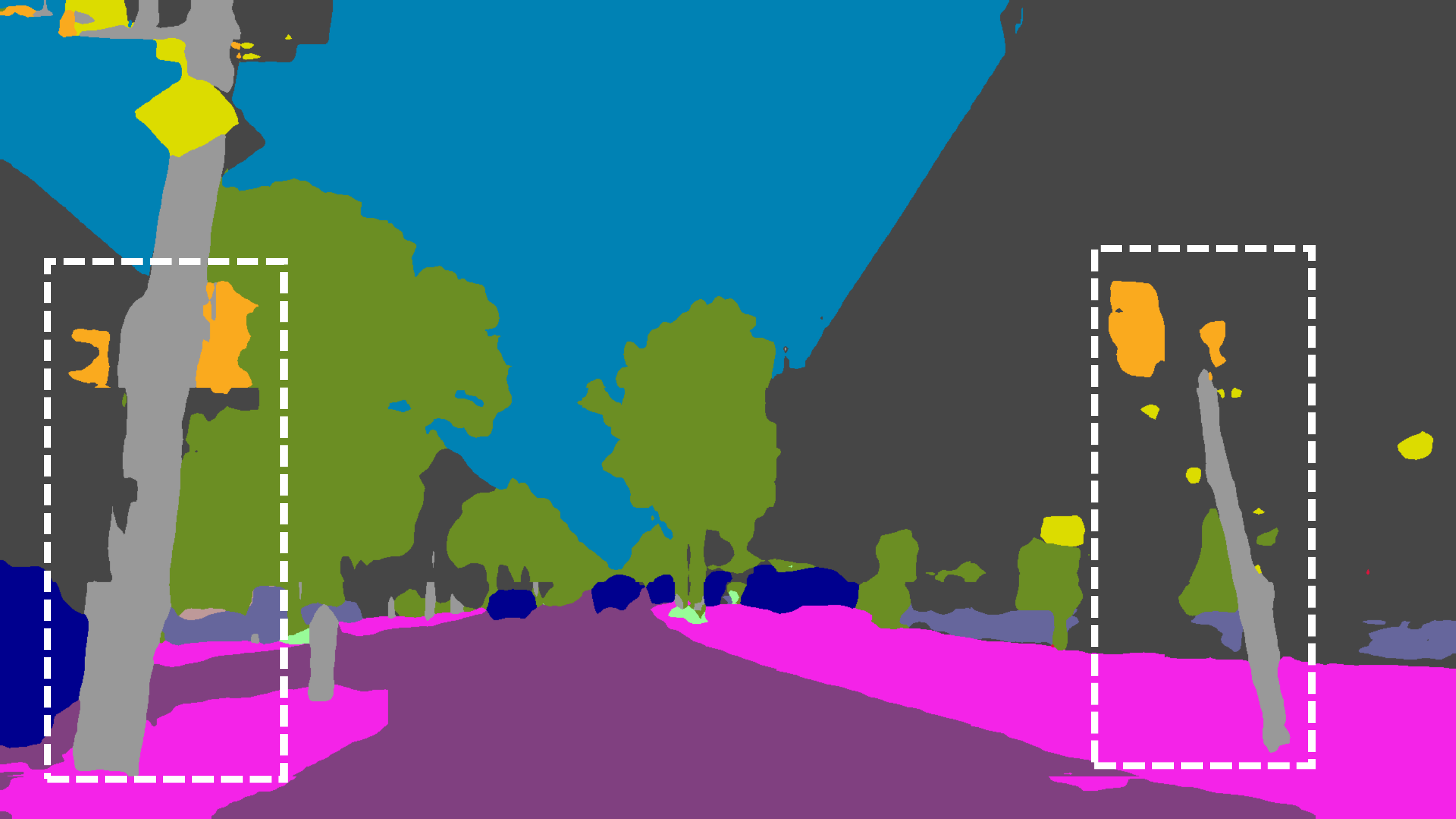}} \hfill
    \subfloat[]{\includegraphics[width=0.142\linewidth]{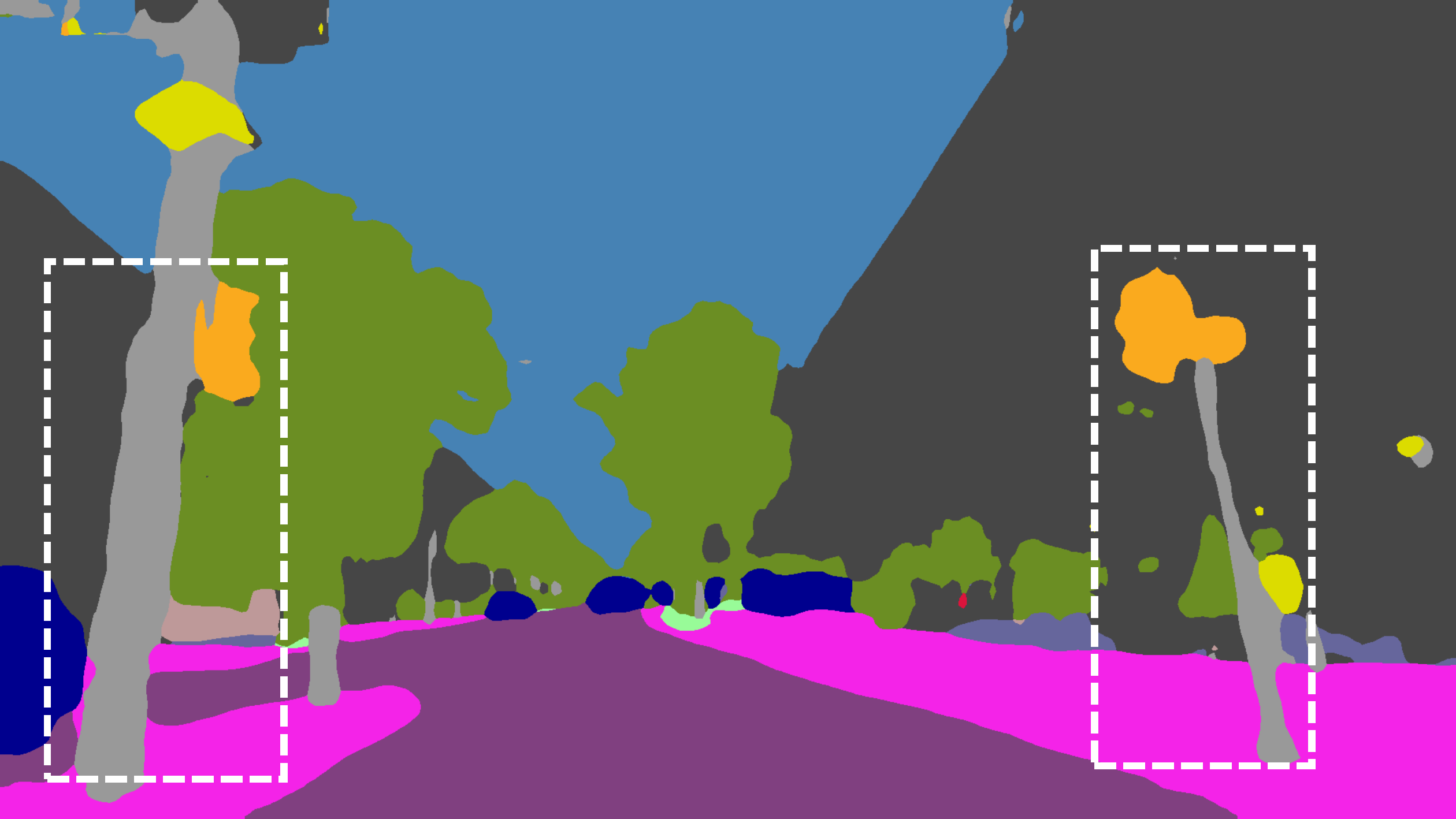}} \hfill
    \subfloat[]{\includegraphics[width=0.142\linewidth]{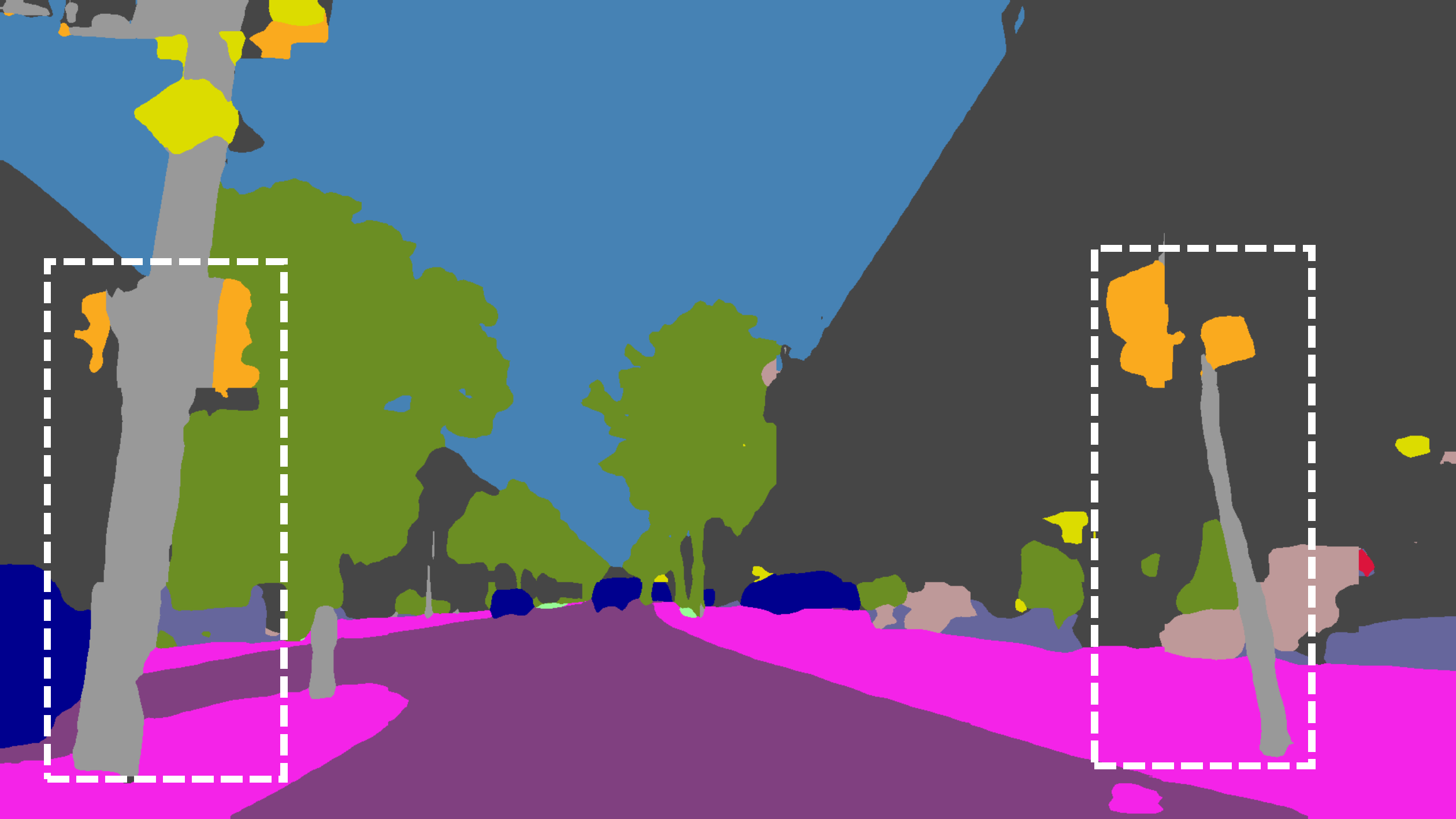}} \hfill
    \subfloat[]{\includegraphics[width=0.142\linewidth]{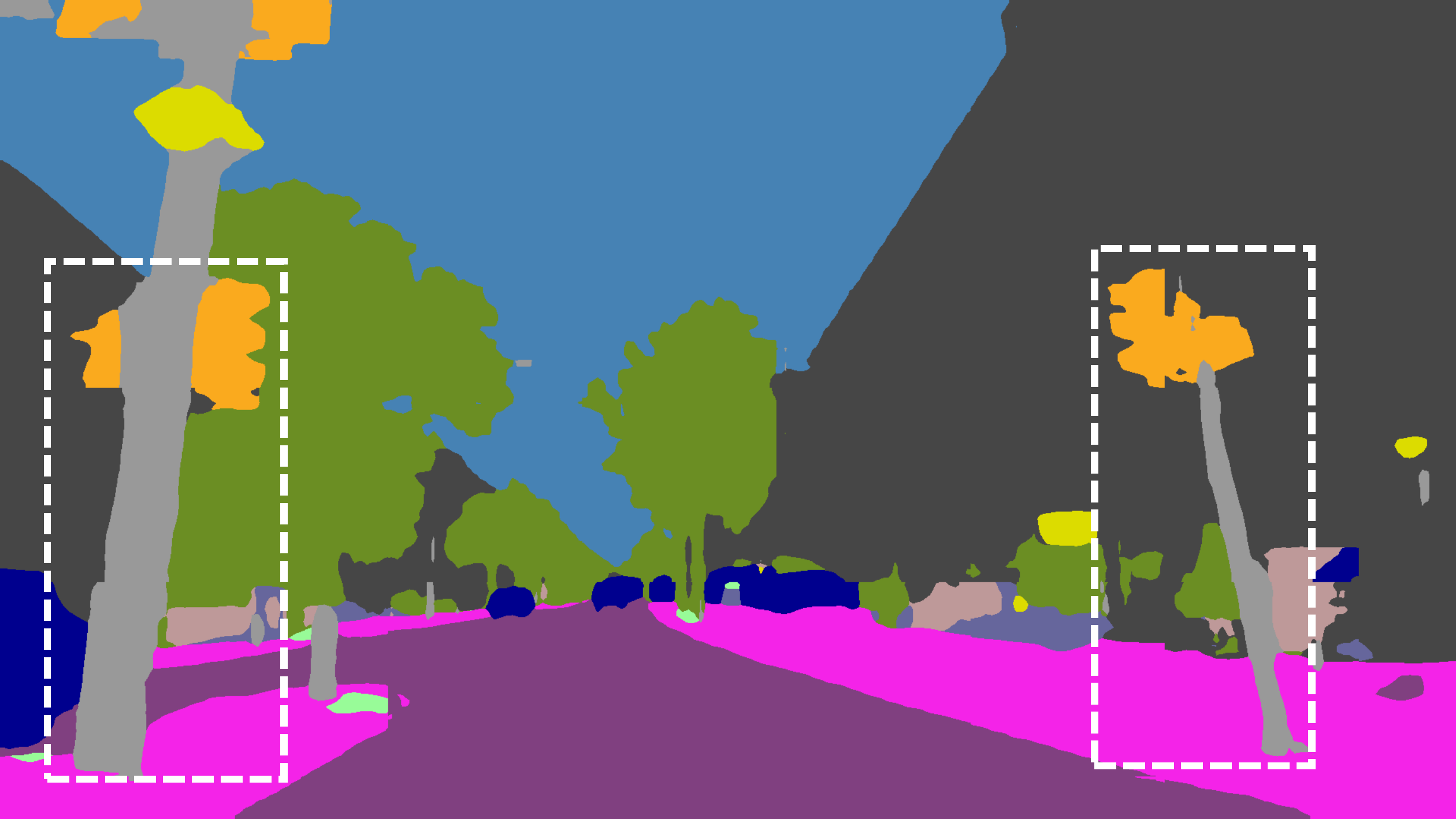}} \hfill
    \subfloat[]{\includegraphics[width=0.142\linewidth]{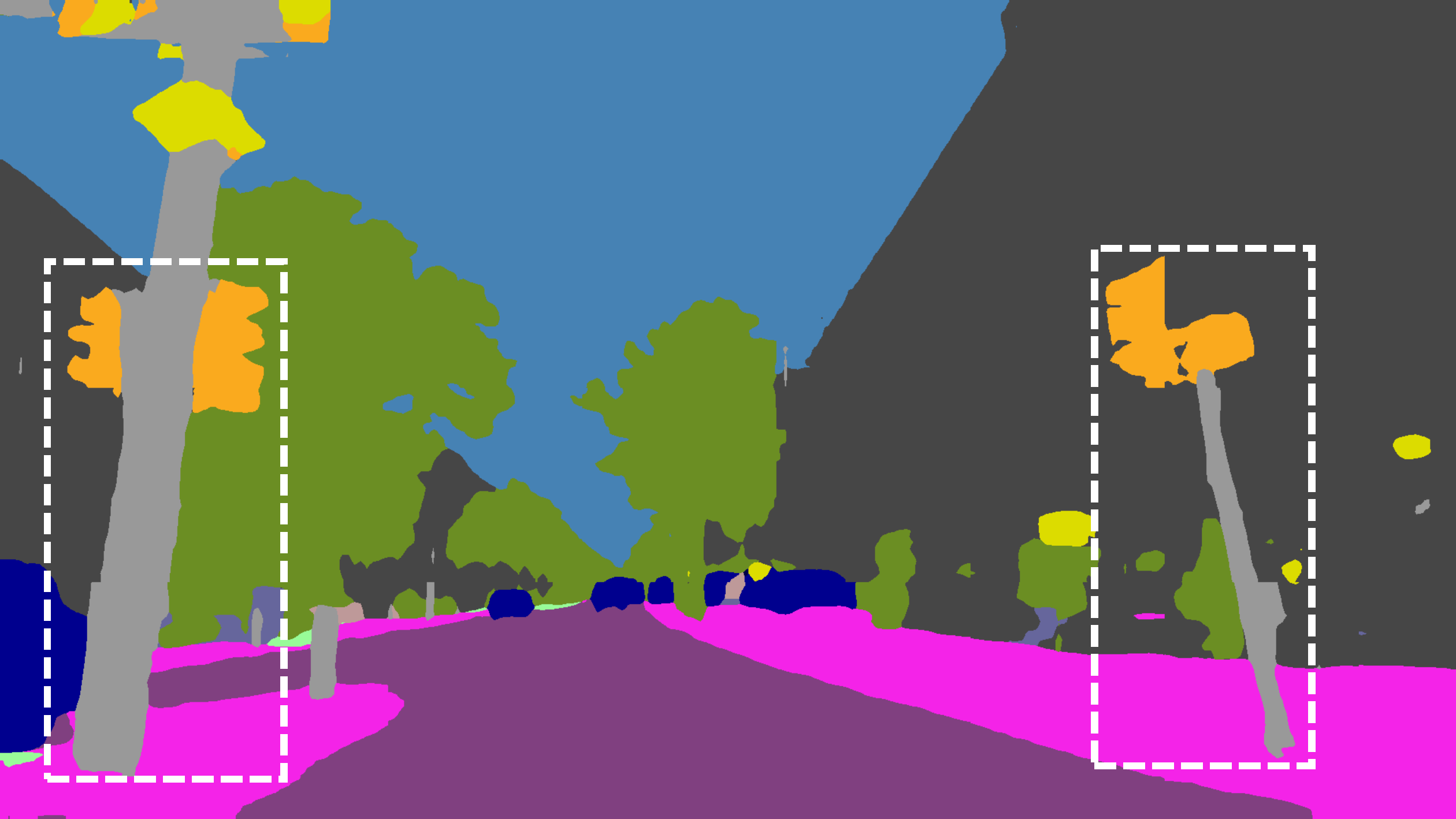}} \\ \vspace{-0.75cm}
    \subfloat[Image with GT]{\includegraphics[width=0.142\linewidth]{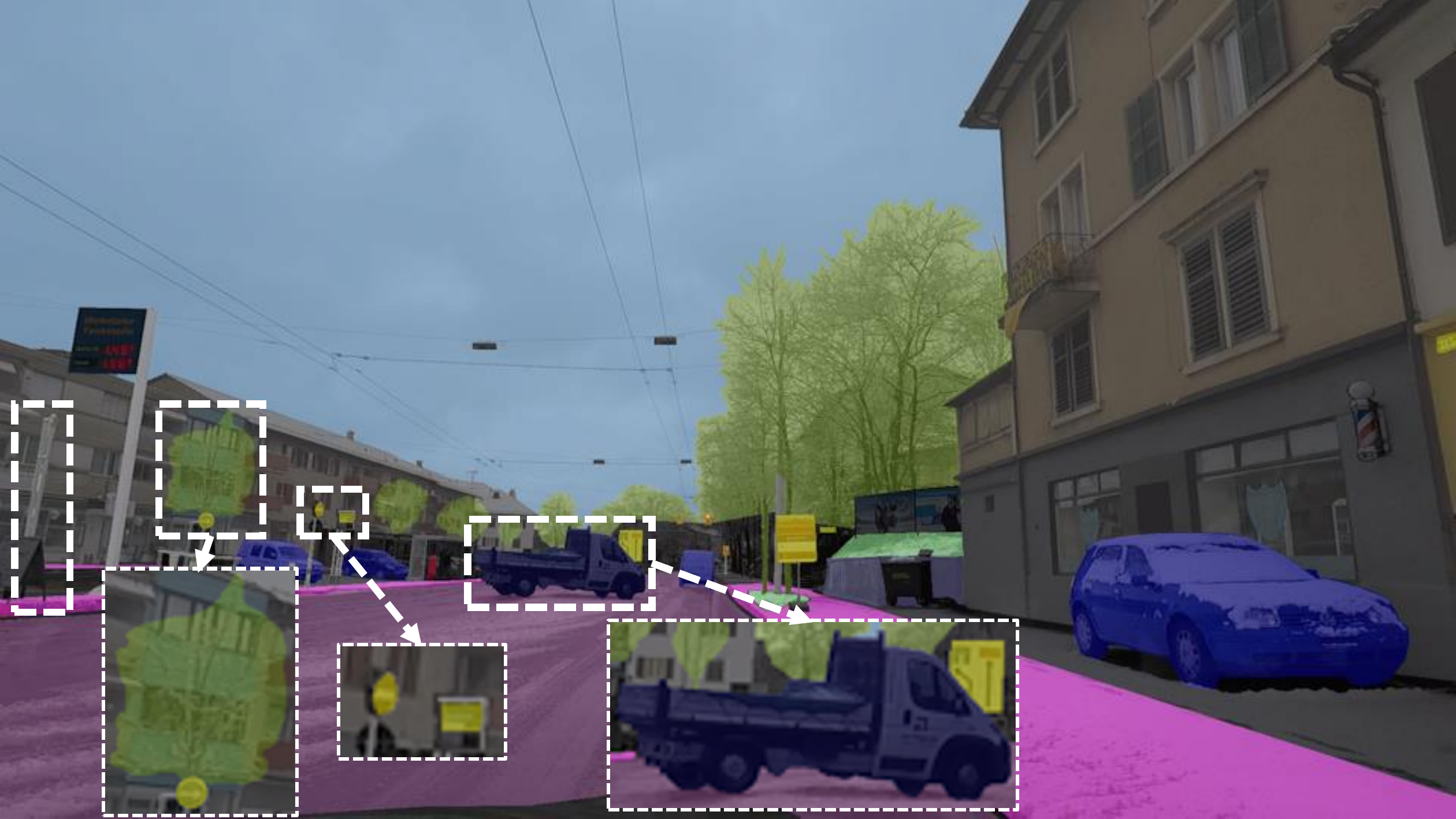}} \hfill
    \subfloat[Supervised (100\%)]{\includegraphics[width=0.142\linewidth]{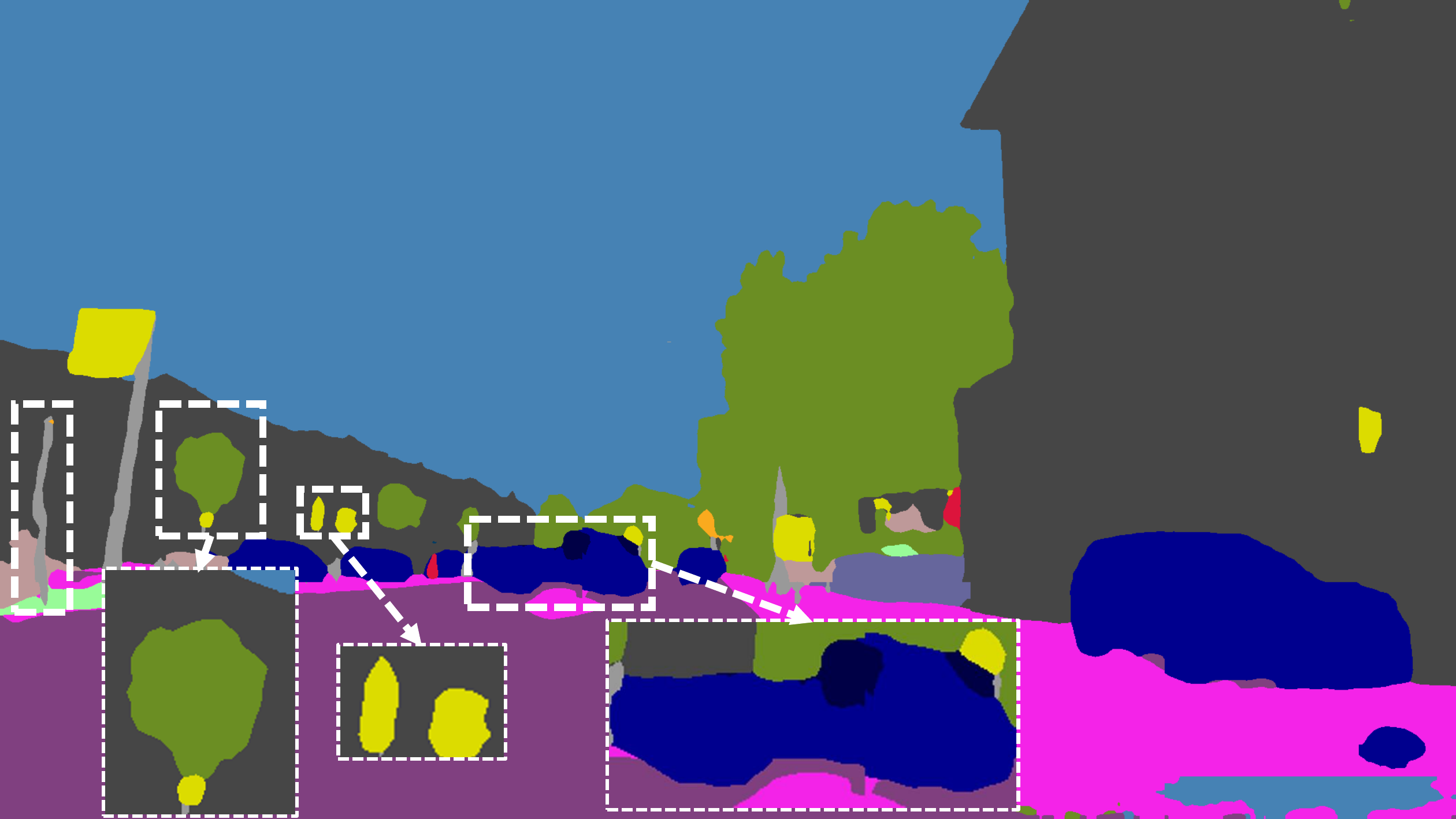}} \hfill
    \subfloat[RIPU (5\%)]{\includegraphics[width=0.142\linewidth]{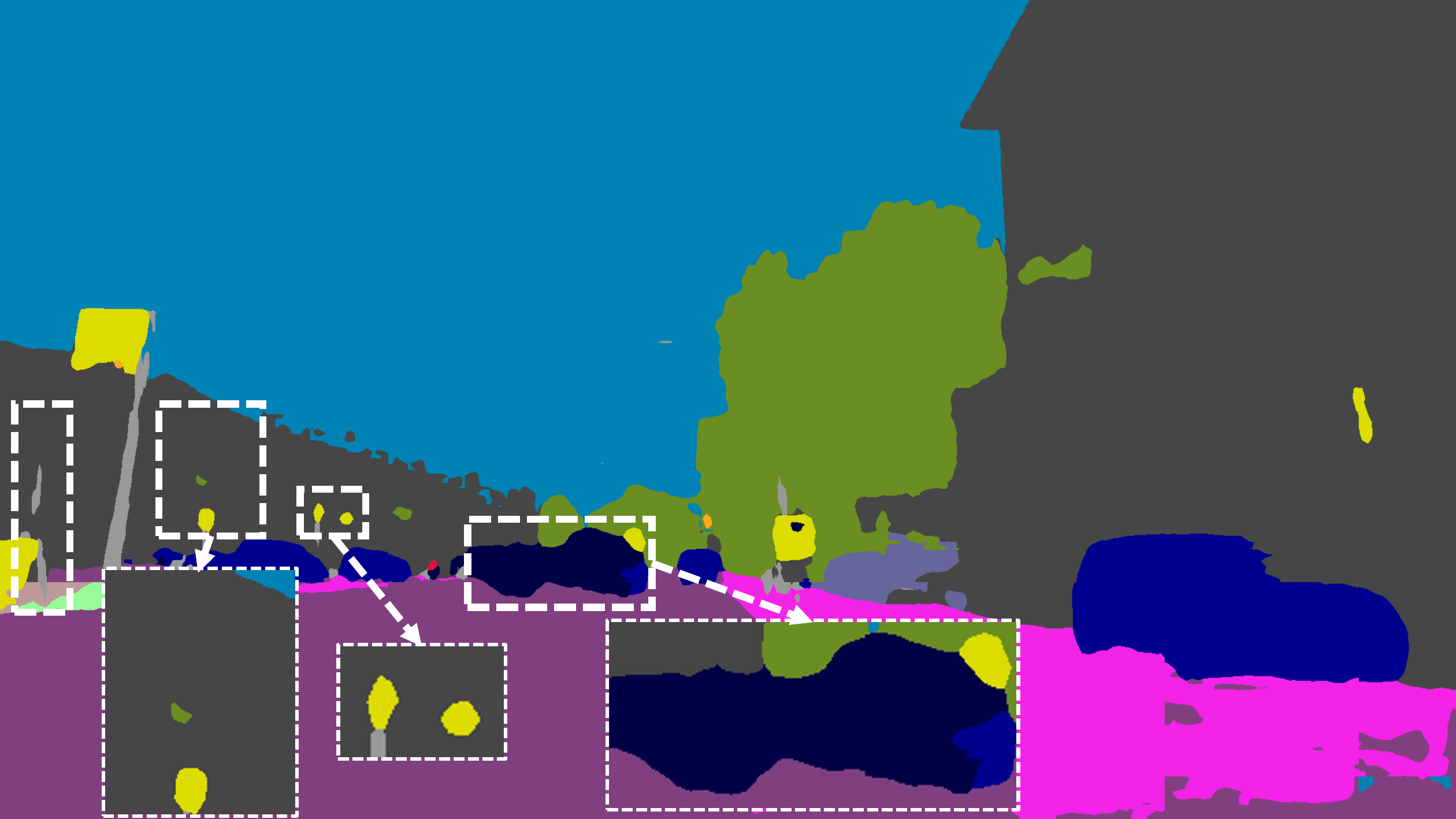}} \hfill
    \subfloat[D2ADA (5\%)]{\includegraphics[width=0.142\linewidth]{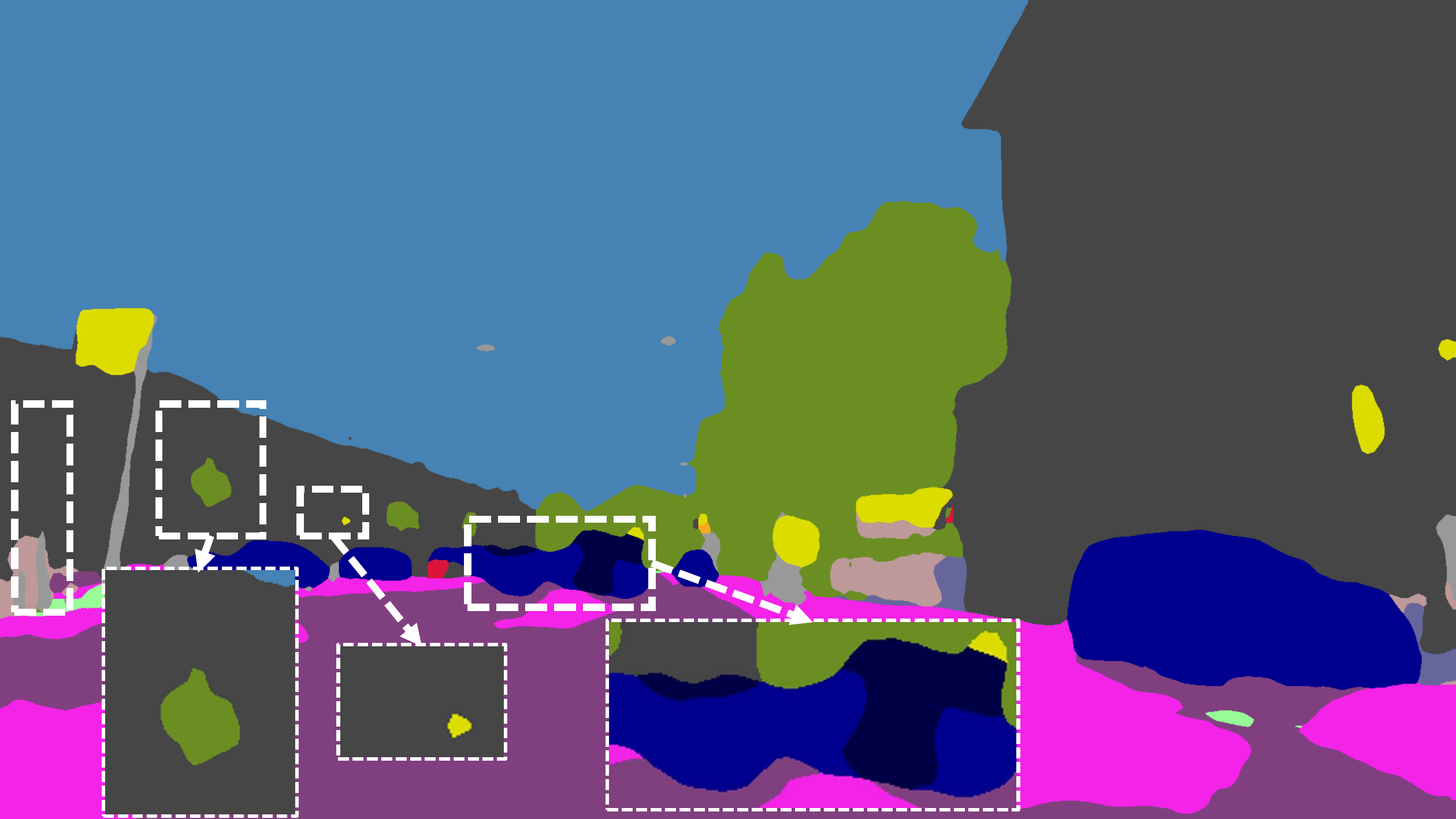}} \hfill
    \subfloat[Ours (5\%)]{\includegraphics[width=0.142\linewidth]{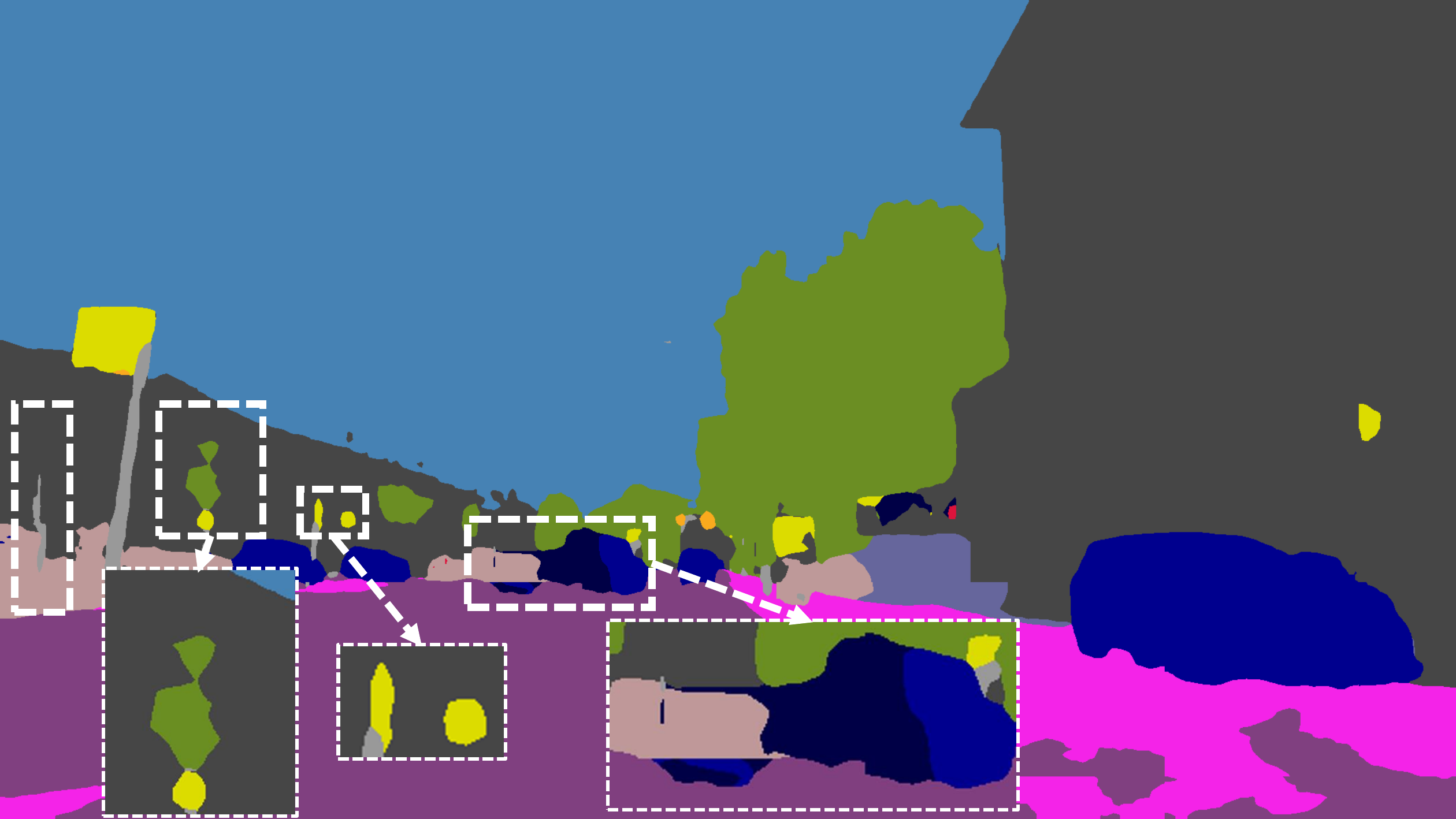}} \hfill
    \subfloat[Ours (25\%)]{\includegraphics[width=0.142\linewidth]{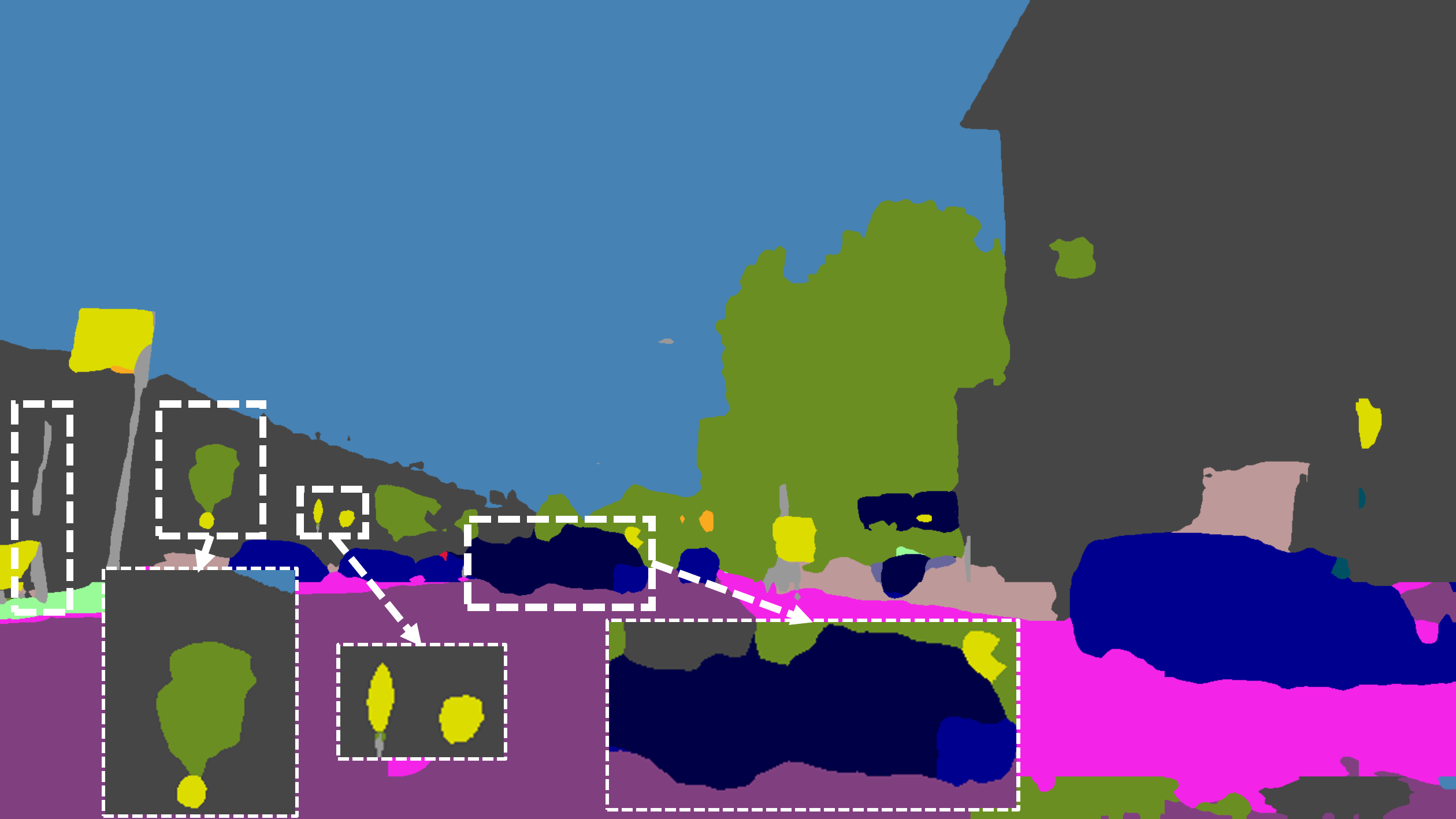}} \hfill
    \subfloat[Ours (50\%)]{\includegraphics[width=0.142\linewidth]{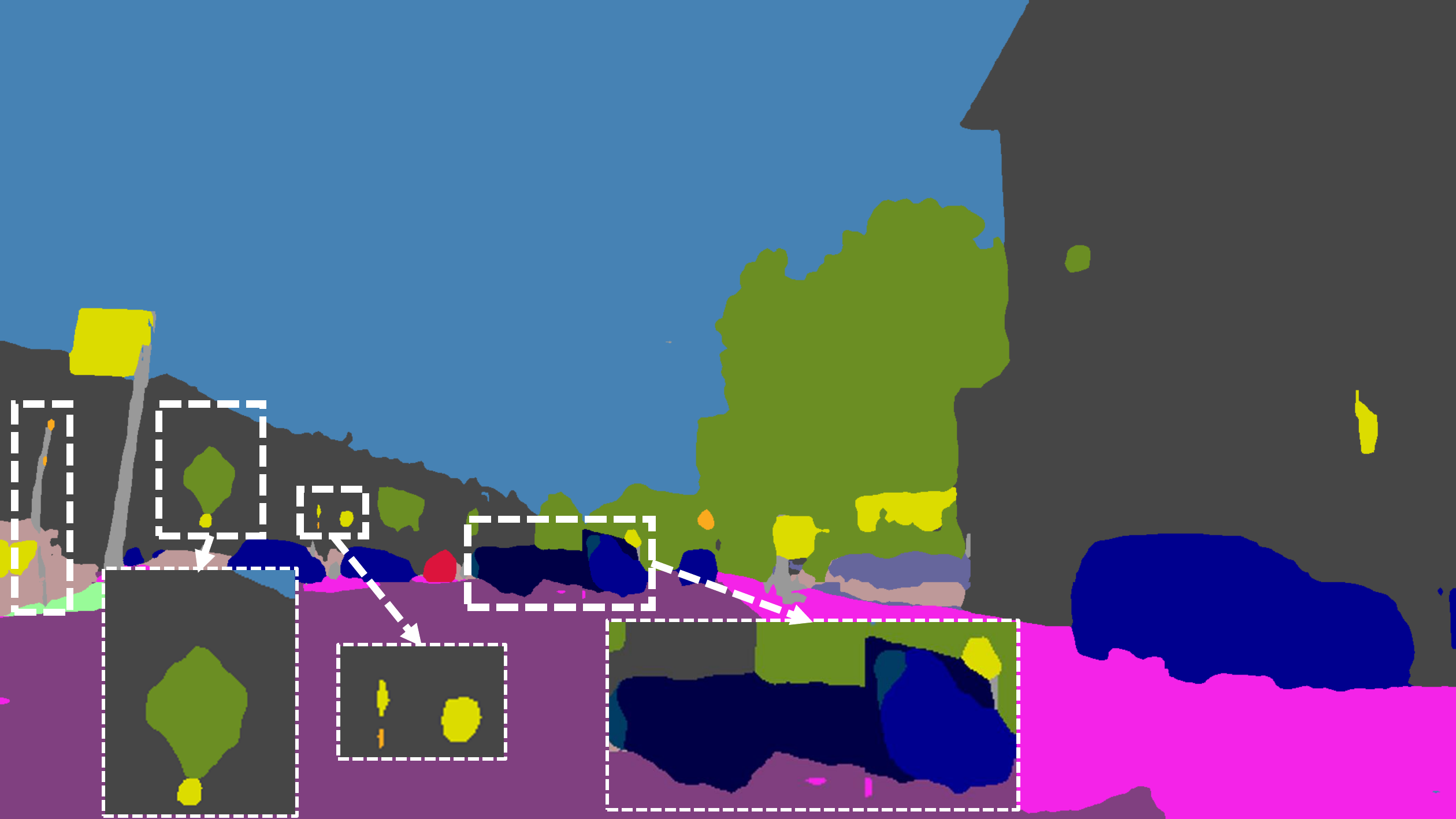}}
    \caption{Qualitative results of different methods on GTA5-to-Cityscapes (rows 1-2), SYNTHIA-to-Cityscapes (rows 3-4), and Cityscapes-to-ACDC (rows 5-6).
	From left to right: the image with ground truth, the prediction of supervised model using 100\% target labeled data, RIPU and D2ADA with 5\% target labeled data, Ours with 5\%, 25\% and 50\% target labeled data.}
    \label{fig:qualitative_comparison}
    \vspace{-0.35cm}
\end{figure*}

\begin{figure}
    \centering
    \vspace{-0.2cm}
    \captionsetup[subfloat]{font=scriptsize,labelfont=scriptsize,labelformat=empty}
    \subfloat[]{\includegraphics[width=0.245\linewidth]{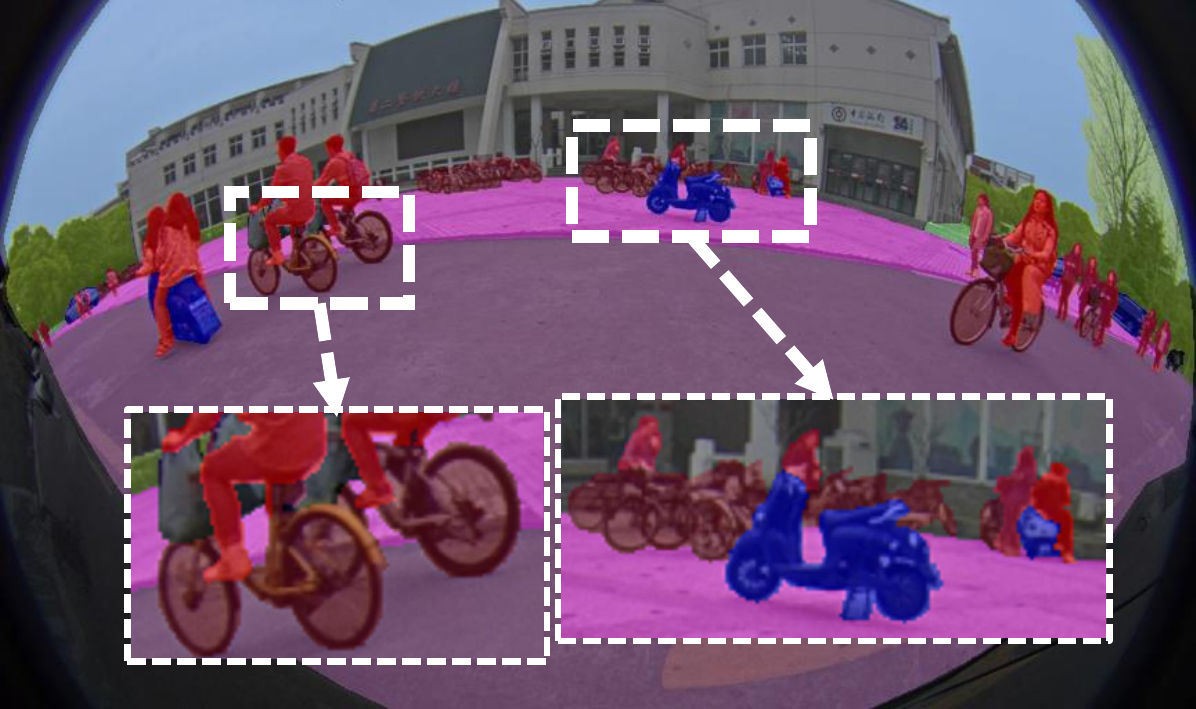}}\hfill
    \subfloat[]{\includegraphics[width=0.245\linewidth]{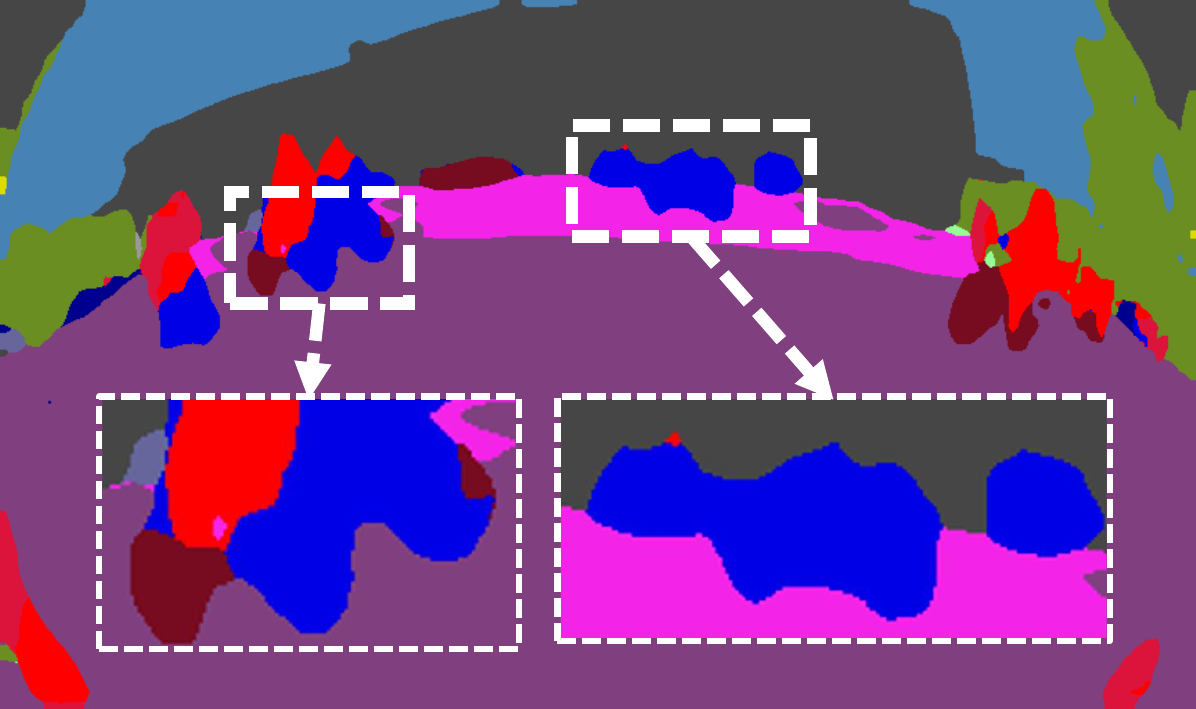}}\hfill
    \subfloat[]{\includegraphics[width=0.245\linewidth]{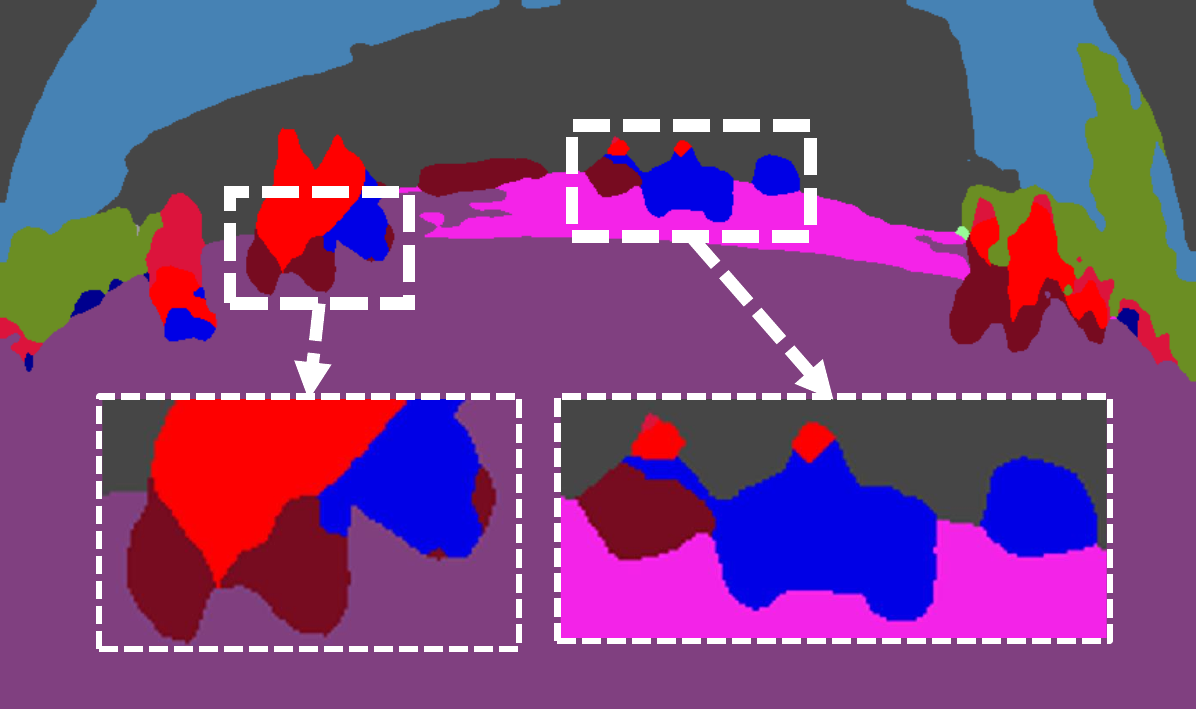}}\hfill
    \subfloat[]{\includegraphics[width=0.245\linewidth]{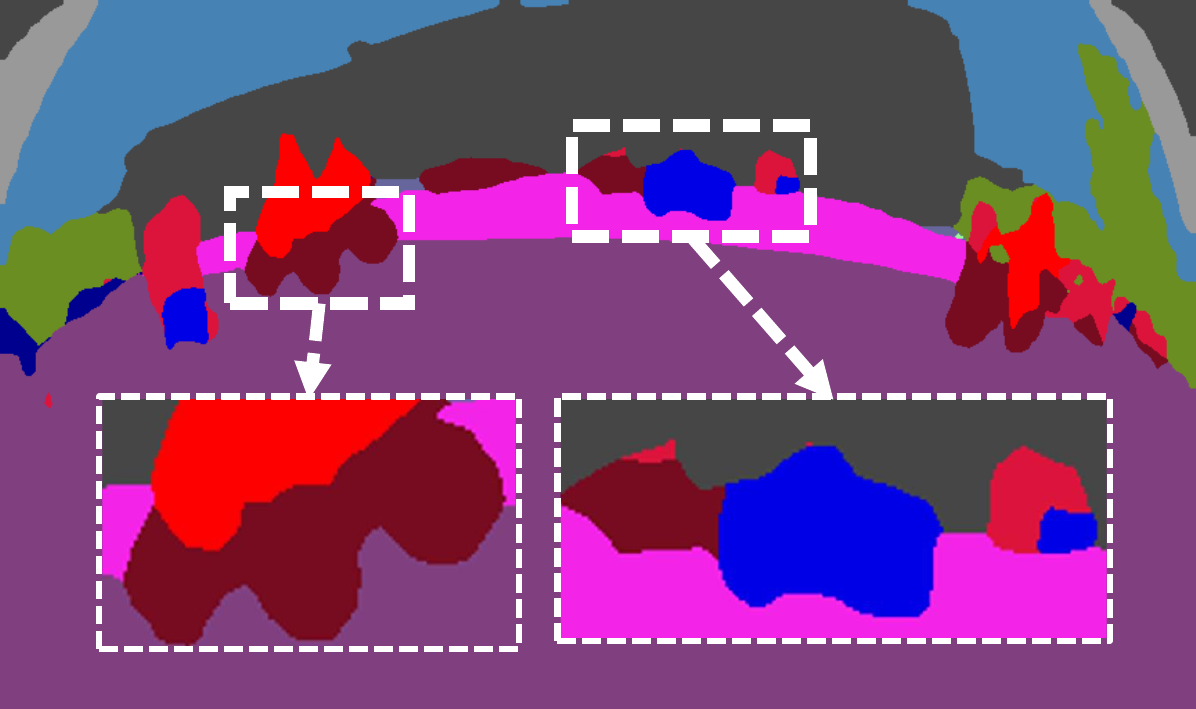}}\\ \vspace{-0.75cm}
    \subfloat[Image with GT]{\includegraphics[width=0.245\linewidth]{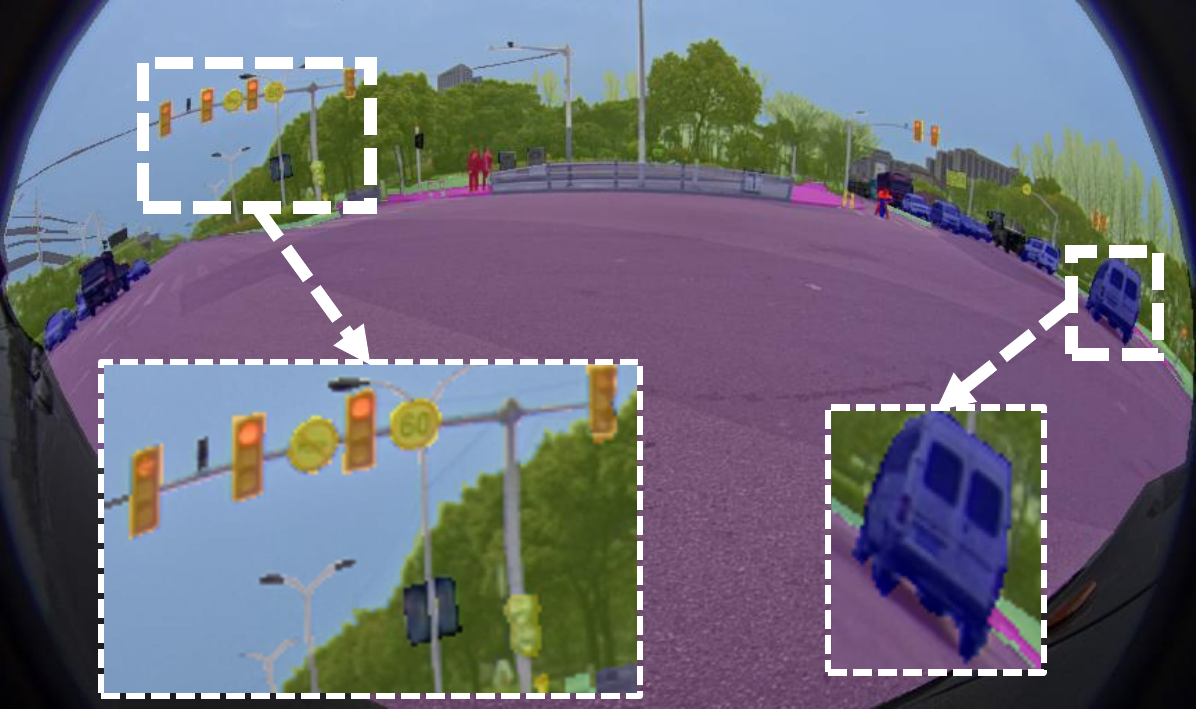}}\hfill
    \subfloat[Supervised (100\%)]{\includegraphics[width=0.245\linewidth]{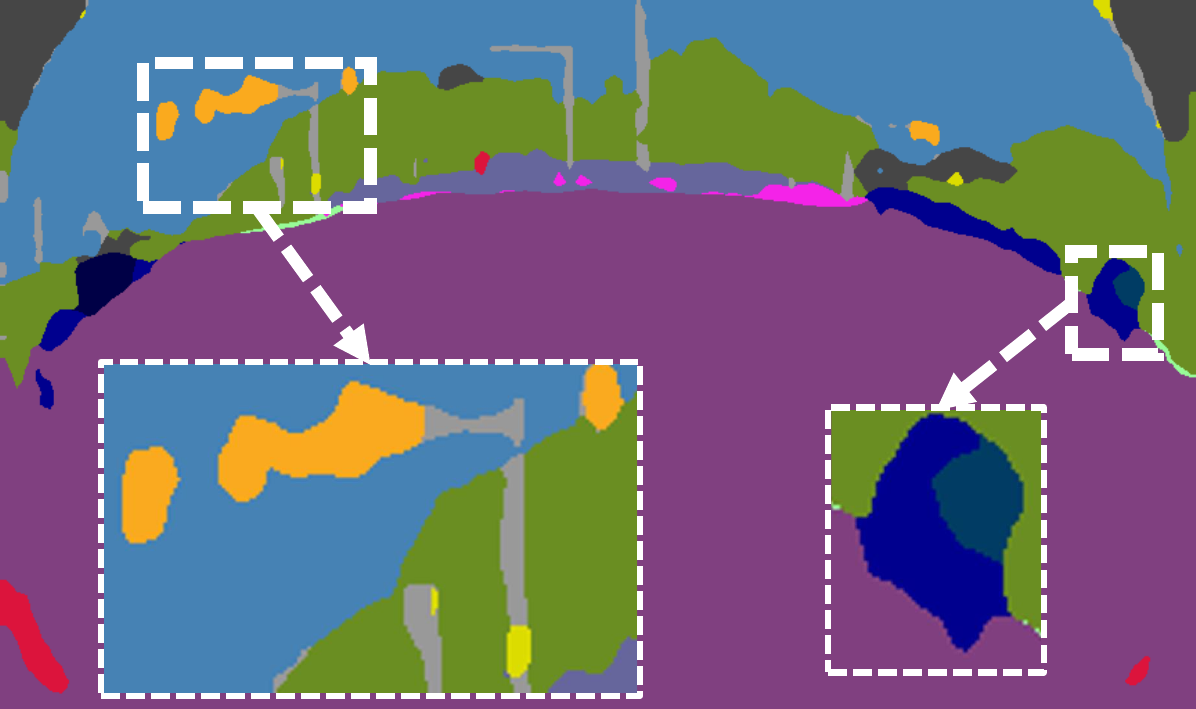}}\hfill
    \subfloat[Ours (25\%)]{\includegraphics[width=0.245\linewidth]{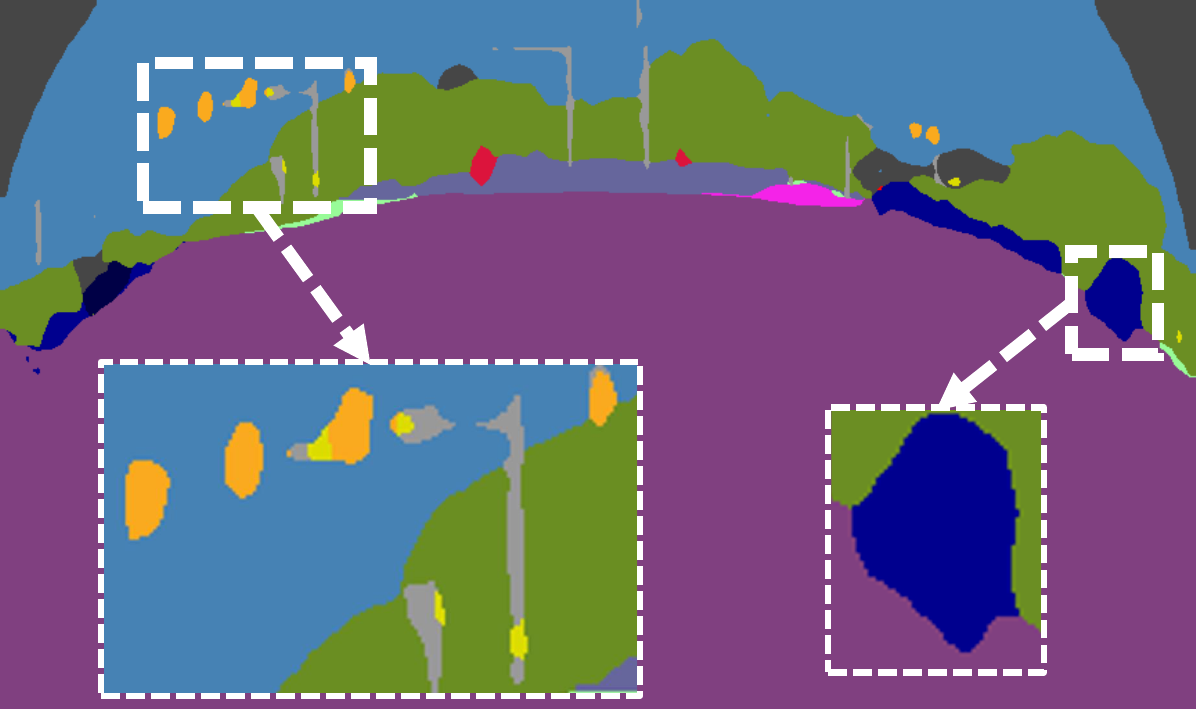}}\hfill
    \subfloat[Ours (50\%)]{\includegraphics[width=0.245\linewidth]{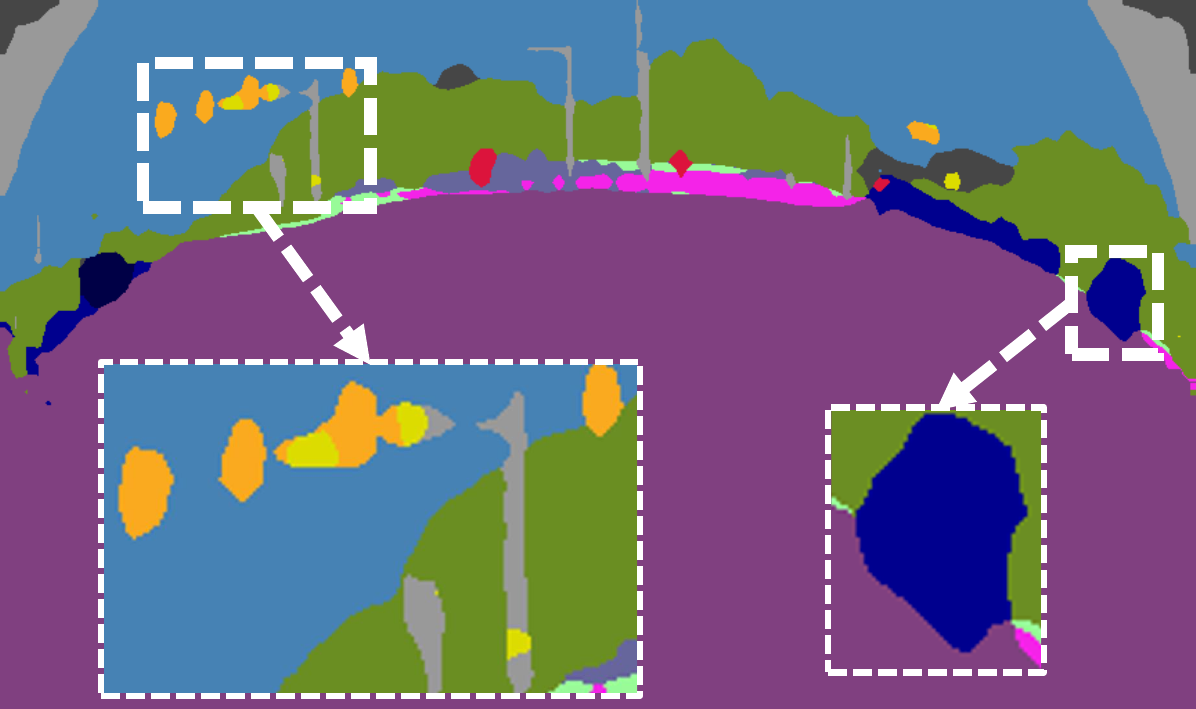}}\\
    \caption{The performance of SS-ADA on Cityscapes-to-FishEyeCampus. From left to right: image with ground truth, the prediction of supervised (100\%), ours (25\%), ours (50\%)}
    \label{fig:school_com}
    \vspace{-0.4cm}
\end{figure}

The first and second rows of Figure~\ref{fig:qualitative_comparison} display the qualitative segmentation results on GTA5-to-Cityscapes. SS-ADA excels in challenging classes such as bus, pole, rider, and bicycle, which are difficult in the Cityscapes dataset.
The third and fourth rows of Figure~\ref{fig:qualitative_comparison} depict the results on SYNTHIA-to-Cityscapes. SS-ADA segments the car, pole, traffic sign, and motorcycle better than RIPU and D2ADA, achieving results comparable to the supervised counterpart. 
The last two rows of Figure~\ref{fig:qualitative_comparison} show the results for Cityscapes-to-ACDC. In adverse weather conditions, SS-ADA segments the truck, pole, traffic light and sign better than RIPU and D2ADA, and is comparable to supervised counterpart.
Figure~\ref{fig:school_com} presents the segmentation performance of SS-ADA on Cityscapes-to-FishEyeCampus in a real-world scenario. SS-ADA exhibits comparable performance with the supervised model and sometimes classifies the bicycle, motorcycle, traffic light and sign better, which aligns with the quantitative results.

\subsection{Ablation Study}

\subsubsection{Effectiveness of each module}

We conducted an ablation study on GTA5-to-Cityscapes with 50\% of target labeled data, and Table~\ref{tab:ablation_study} presents the results. The source-only model has only 40.8\% mIoU on Cityscapes. When adding the semi-supervised learning module, the mIoU increases to 68.5\%. Incorporating only the active learning module results in an mIoU of 70.4\%. Combining both of them raises the mIoU to 71.4\%. Finally, integrating the IoU-based class weighting achieves the highest mIoU of 72.1\%. These results demonstrate the effectiveness of each module in SS-ADA, with each component contributing to the overall performance improvement. 


\begin{table}[ht]
    \centering
    \caption{The effect of each module in SS-ADA. Using 50\% of target labeled data in GTA5-to-Cityscapes.}
    \label{tab:ablation_study}
    \begin{tabular}{c|c|c|c}
        \hline
        Semi Supervised & Active Learning & Class Weight & mIoU \\ \hline
        & & & 40.8 \\ \hline 
        $\surd$ & & & 68.5 \\ \hline
        & $\surd$ & & 70.4 \\ \hline
        $\surd$ & $\surd$ & & 71.4 \\ \hline
        $\surd$ & $\surd$ & $\surd$ & 72.1 \\ \hline
    \end{tabular}
    \vspace{-0.2cm}
\end{table}

\subsubsection{The impact of acquisition strategy}

We conducted experiments on GTA5-to-Cityscapes using image-level entropy and confidence as acquisition strategies, and the results are shown in Table~\ref{tab:sample_selection}. Across four different annotation ratios, the differences in performance between using entropy and confidence were not significant. Since the overall performance of entropy is slightly better, we adopt it in our experiments.
\begin{table}[ht]
    \centering
    \renewcommand{\arraystretch}{1.05}
    \caption{The impact of sample acquisition strategy. Using 50\% of target labeled data in GTA5-to-Cityscapes.} 
    \label{tab:sample_selection}
    \begin{tabular}{c|cccc}
        \hline
        \multirow{2}{*}{Strategy} & \multicolumn{4}{c}{mIoU} \\ \cline{2-5}
         & 50\% & 25\% & 12.5\% & 6.25\% \\ \hline
        entropy & \textbf{72.1} & \textbf{71.3} & \textbf{70.8} & \textbf{67.7} \\ \hline
        confidence & 71.3 & 70.2 & 69.5 & 67.3 \\ \hline
    \end{tabular}
    \vspace{-0.2cm}
\end{table}

\subsubsection{The impact of active learning trigger epoches $T_{ac}$}
\begin{table}[!ht]
    \centering
    \caption{The impact of active learning trigger epoches.} 
    \label{tab:sampling_epoches}
    \begin{tabular}{c|c|c|c|c|c}
        \hline
        $T_{ac}$ & mIoU & $T_{ac}$ & mIoU & $T_{ac}$ & mIoU \\ \hline
        {10, 20} & 64.8 & {10, 20, 30} & 65.3 & {10, 20, 30, 40} & 65.5 \\
        {20, 40} & 64.0 & {20, 40, 60} & \textbf{66.3} & {20, 40, 60, 80} & \textbf{65.9} \\
        {30, 60} & \textbf{65.0} & {30, 60, 90} & 65.8 & {30, 60, 90, 120} & 64.1 \\
        \hline
        average & 64.6 & average & \textbf{65.8} & average & 65.2 \\
        \hline
    \end{tabular}
    \vspace{-0.15cm}
\end{table}

We conduct experiments on Cityscapes-to-FishEyeCampus with 12.5\% of target labeled data under different $T_{ac}$ and the results are shown in Table~\ref{tab:sampling_epoches}. Overall, the impact of different $T_{ac}$ on model accuracy is not particularly significant. Setting the trigger epoches within the first half of the total training epochs yields satisfactory results. As the number of trigger epoches increases from two to four, the average mIoU initially improves and then declines, with three trigger epoches achieving the best performance. In the three-sampling setup, the configuration of sampling at epochs 20, 40, and 60 shows the best results, and we use this setting in our experiments.

\subsubsection{The impact of the upper bound of class weight $u$}
\begin{figure}
    \vspace{-0.15cm}
    \centering
    \includegraphics[width=0.9\linewidth]{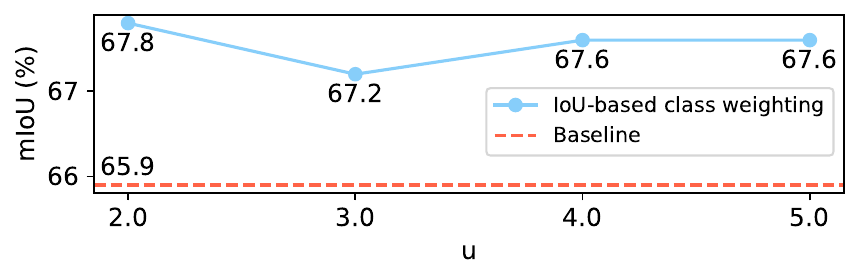}
    \vspace{-0.15cm}
    \caption{The impact of the upper bound $u$.}
    \label{fig:class_weight}
    \vspace{-0.6cm}
\end{figure}

We set $u$ to different values and the results on Cityscapes-to-FishEyeCampus with 25\% of target labeled data are shown in Figure~\ref{fig:class_weight}. SS-ADA consistently outperforms the baseline of 65.9\% mIoU across different $u$, with no significant performance differences observed. This indicates that SS-ADA is insensitive to variations in $u$, and we set it to 2.0. 
\vspace{-0.2cm}

%% file: tex/5_conclusion.tex
\section{Conclusion}
In this paper, we propose SS-ADA, a novel semi-supervised active domain adaptation framework for semantic segmentation in driving scenes. SS-ADA incorporates active learning into semi-supervised semantic segmentation to select the more informative unlabeled data for manual annotation and model training. This approach facilitates the application of semantic segmentation models to new driving scenarios and significantly reduces annotation costs. We find that image-level acquisition strategies are less sensitive to class imbalance issue and design an IoU-based class weighting strategy to alleviate this. Experimental results demonstrate the effectiveness of SS-ADA across various driving scenarios, including the synthetic-to-real, normal-to-adverse weather, and pinhole-to-fisheye camera adaptation datasets. SS-ADA achieves the accuracy of supervised learning using only 25\% of the labeled samples from the target domain. 

In future work, we will explore the design of more effective data acquisition strategies, the application of SS-ADA to other tasks and domains, and consider incorporating vision foundation models into the framework. 

%% file: acda.bbl
\begin{thebibliography}{10}
\providecommand{\url}[1]{#1}
\csname url@samestyle\endcsname
\providecommand{\newblock}{\relax}
\providecommand{\bibinfo}[2]{#2}
\providecommand{\BIBentrySTDinterwordspacing}{\spaceskip=0pt\relax}
\providecommand{\BIBentryALTinterwordstretchfactor}{4}
\providecommand{\BIBentryALTinterwordspacing}{\spaceskip=\fontdimen2\font plus
\BIBentryALTinterwordstretchfactor\fontdimen3\font minus
  \fontdimen4\font\relax}
\providecommand{\BIBforeignlanguage}[2]{{%
\expandafter\ifx\csname l@#1\endcsname\relax
\typeout{** WARNING: IEEEtran.bst: No hyphenation pattern has been}%
\typeout{** loaded for the language `#1'. Using the pattern for}%
\typeout{** the default language instead.}%
\else
\language=\csname l@#1\endcsname
\fi
#2}}
\providecommand{\BIBdecl}{\relax}
\BIBdecl

\bibitem{hyseg}
Y.~Qian, X.~Wang, Z.~Chen, C.~Wang, and M.~Yang, ``Hy-seg: A hybrid method for
  ground segmentation using point clouds,'' \emph{IEEE Transactions on
  Intelligent Vehicles}, vol.~8, no.~2, pp. 1597--1606, 2022.

\bibitem{restricted}
L.~Deng, M.~Yang, H.~Li, T.~Li, B.~Hu, and C.~Wang, ``Restricted deformable
  convolution-based road scene semantic segmentation using surround view
  cameras,'' \emph{IEEE Transactions on Intelligent Transportation Systems},
  vol.~21, no.~10, pp. 4350--4362, 2019.

\bibitem{gated}
Y.~Qian, L.~Deng, T.~Li, C.~Wang, and M.~Yang, ``Gated-residual block for
  semantic segmentation using rgb-d data,'' \emph{IEEE Transactions on
  Intelligent Transportation Systems}, vol.~23, no.~8, pp. 11\,836--11\,844,
  2021.

\bibitem{segtransconv}
J.~Fan, B.~Gao, Q.~Ge, Y.~Ran, J.~Zhang, and H.~Chu, ``Segtransconv:
  Transformer and cnn hybrid method for real-time semantic segmentation of
  autonomous vehicles,'' \emph{IEEE Transactions on Intelligent Transportation
  Systems}, vol.~25, no.~2, pp. 1586--1601, 2024.

\bibitem{transfer}
J.~Zhang, C.~Ma, K.~Yang, A.~Roitberg, K.~Peng, and R.~Stiefelhagen, ``Transfer
  beyond the field of view: Dense panoramic semantic segmentation via
  unsupervised domain adaptation,'' \emph{IEEE Transactions on Intelligent
  Transportation Systems}, vol.~23, no.~7, pp. 9478--9491, 2021.

\bibitem{cmx}
J.~Zhang, H.~Liu, K.~Yang, X.~Hu, R.~Liu, and R.~Stiefelhagen, ``Cmx:
  Cross-modal fusion for rgb-x semantic segmentation with transformers,''
  \emph{IEEE Transactions on Intelligent Transportation Systems}, vol.~24,
  no.~12, pp. 14\,679--14\,694, 2023.

\bibitem{cityscapes}
M.~Cordts, M.~Omran, S.~Ramos, T.~Rehfeld, M.~Enzweiler, R.~Benenson,
  U.~Franke, S.~Roth, and B.~Schiele, ``The cityscapes dataset for semantic
  urban scene understanding,'' in \emph{Proceedings of the IEEE/CVF Conference
  on Computer Vision and Pattern Recognition (CVPR)}, 2016, pp. 3213--3223.

\bibitem{acdc}
C.~Sakaridis, D.~Dai, and L.~Van~Gool, ``Acdc: The adverse conditions dataset
  with correspondences for semantic driving scene understanding,'' in
  \emph{Proceedings of the IEEE/CVF International Conference on Computer Vision
  (ICCV)}, 2021, pp. 10\,765--10\,775.

\bibitem{tits-semi}
J.~Mei, B.~Gao, D.~Xu, W.~Yao, X.~Zhao, and H.~Zhao, ``Semantic segmentation of
  3d lidar data in dynamic scene using semi-supervised learning,'' \emph{IEEE
  Transactions on Intelligent Transportation Systems}, vol.~21, no.~6, pp.
  2496--2509, 2020.

\bibitem{u2pl}
Y.~Wang, H.~Wang, Y.~Shen, J.~Fei, W.~Li, G.~Jin, L.~Wu, R.~Zhao, and X.~Le,
  ``Semi-supervised semantic segmentation using unreliable pseudo-labels,'' in
  \emph{Proceedings of the IEEE/CVF Conference on Computer Vision and Pattern
  Recognition (CVPR)}, 2022, pp. 4248--4257.

\bibitem{unimatch}
L.~Yang, L.~Qi, L.~Feng, W.~Zhang, and Y.~Shi, ``Revisiting weak-to-strong
  consistency in semi-supervised semantic segmentation,'' in \emph{Proceedings
  of the IEEE/CVF Conference on Computer Vision and Pattern Recognition
  (CVPR)}, 2023, pp. 7236--7246.

\bibitem{seg2depth}
H.~Ibrahem, A.~Salem, and H.-S. Kang, ``Seg2depth: Semi-supervised depth
  estimation for autonomous vehicles using semantic segmentation and single
  vanishing point fusion,'' \emph{IEEE Transactions on Intelligent Vehicles},
  2024.

\bibitem{deepbayesian}
Y.~Gal, R.~Islam, and Z.~Ghahramani, ``Deep bayesian active learning with image
  data,'' in \emph{International Conference on Machine Learning}.\hskip 1em
  plus 0.5em minus 0.4em\relax PMLR, 2017, pp. 1183--1192.

\bibitem{wu2021redal}
T.-H. Wu, Y.-C. Liu, Y.-K. Huang, H.-Y. Lee, H.-T. Su, P.-C. Huang, and W.~H.
  Hsu, ``Redal: Region-based and diversity-aware active learning for point
  cloud semantic segmentation,'' in \emph{Proceedings of the IEEE/CVF
  International Conference on Computer Vision (ICCV)}, 2021, pp.
  15\,510--15\,519.

\bibitem{ripu}
B.~Xie, L.~Yuan, S.~Li, C.~H. Liu, and X.~Cheng, ``Towards fewer annotations:
  Active learning via region impurity and prediction uncertainty for domain
  adaptive semantic segmentation,'' in \emph{Proceedings of the IEEE/CVF
  Conference on Computer Vision and Pattern Recognition (CVPR)}, 2022, pp.
  8068--8078.

\bibitem{activecontrastive}
B.~Gao, X.~Zhao, and H.~Zhao, ``An active and contrastive learning framework
  for fine-grained off-road semantic segmentation,'' \emph{IEEE Transactions on
  Intelligent Transportation Systems}, vol.~24, no.~1, pp. 564--579, 2023.

\bibitem{active_tits}
K.~Chitta, J.~M. Álvarez, E.~Haussmann, and C.~Farabet, ``Training data subset
  search with ensemble active learning,'' \emph{IEEE Transactions on
  Intelligent Transportation Systems}, vol.~23, no.~9, pp. 14\,741--14\,752,
  2022.

\bibitem{activesurvey}
X.~Li, Z.~Wang, Y.~Huang, and H.~Chen, ``A survey on self-evolving autonomous
  driving: a perspective on data closed-loop technology,'' \emph{IEEE
  Transactions on Intelligent Vehicles}, vol.~8, no.~11, pp. 4613--4631, 2023.

\bibitem{d2ada}
T.-H. Wu, Y.-S. Liou, S.-J. Yuan, H.-Y. Lee, T.-I. Chen, K.-C. Huang, and W.~H.
  Hsu, ``D 2 ada: Dynamic density-aware active domain adaptation for semantic
  segmentation,'' in \emph{European Conference on Computer Vision
  (ECCV)}.\hskip 1em plus 0.5em minus 0.4em\relax Springer, 2022, pp. 449--467.

\bibitem{playingfordata}
S.~R. Richter, V.~Vineet, S.~Roth, and V.~Koltun, ``Playing for data: Ground
  truth from computer games,'' in \emph{European Conference on Computer Vision
  (ECCV)}.\hskip 1em plus 0.5em minus 0.4em\relax Springer, 2016, pp. 102--118.

\bibitem{synthia}
G.~Ros, L.~Sellart, J.~Materzynska, D.~Vazquez, and A.~M. Lopez, ``The synthia
  dataset: A large collection of synthetic images for semantic segmentation of
  urban scenes,'' in \emph{Proceedings of the IEEE/CVF Conference on Computer
  Vision and Pattern Recognition (CVPR)}, 2016, pp. 3234--3243.

\bibitem{simulator}
Y.~Li, W.~Yuan, S.~Zhang, W.~Yan, Q.~Shen, C.~Wang, and M.~Yang, ``Choose your
  simulator wisely: A review on open-source simulators for autonomous
  driving,'' \emph{IEEE Transactions on Intelligent Vehicles}, 2024.

\bibitem{halo}
L.~Franco, P.~Mandica, K.~Kallidromitis, D.~Guillory, Y.-T. Li, T.~Darrell, and
  F.~Galasso, ``Hyperbolic active learning for semantic segmentation under
  domain shift,'' in \emph{International Conference on Machine Learning}.\hskip
  1em plus 0.5em minus 0.4em\relax PMLR, 2024.

\bibitem{Augmentation}
Z.~Zhao, L.~Yang, S.~Long, J.~Pi, L.~Zhou, and J.~Wang, ``Augmentation matters:
  A simple-yet-effective approach to semi-supervised semantic segmentation,''
  in \emph{Proceedings of the IEEE/CVF Conference on Computer Vision and
  Pattern Recognition (CVPR)}, 2023, pp. 11\,350--11\,359.

\bibitem{classmix}
V.~Olsson, W.~Tranheden, J.~Pinto, and L.~Svensson, ``Classmix:
  Segmentation-based data augmentation for semi-supervised learning,'' in
  \emph{Proceedings of the IEEE/CVF Winter Conference on Applications of
  Computer Vision (WACV)}, 2021, pp. 1369--1378.

\bibitem{st++}
L.~Yang, W.~Zhuo, L.~Qi, Y.~Shi, and Y.~Gao, ``St++: Make self-training work
  better for semi-supervised semantic segmentation,'' in \emph{Proceedings of
  the IEEE/CVF Conference on Computer Vision and Pattern Recognition (CVPR)},
  2022, pp. 4268--4277.

\bibitem{usrn}
D.~Guan, J.~Huang, A.~Xiao, and S.~Lu, ``Unbiased subclass regularization for
  semi-supervised semantic segmentation,'' in \emph{Proceedings of the IEEE/CVF
  Conference on Computer Vision and Pattern Recognition (CVPR)}, 2022, pp.
  9968--9978.

\bibitem{CorrMatch}
B.~Sun, Y.~Yang, L.~Zhang, M.-M. Cheng, and Q.~Hou, ``Corrmatch: Label
  propagation via correlation matching for semi-supervised semantic
  segmentation,'' \emph{IEEE Computer Vision and Pattern Recognition (CVPR)},
  2024.

\bibitem{cat_adversal}
Z.~Yuan, C.~Wen, M.~Cheng, Y.~Su, W.~Liu, S.~Yu, and C.~Wang, ``Category-level
  adversaries for outdoor lidar point clouds cross-domain semantic
  segmentation,'' \emph{IEEE Transactions on Intelligent Transportation
  Systems}, vol.~24, no.~2, pp. 1982--1993, 2023.

\bibitem{daformer}
L.~Hoyer, D.~Dai, and L.~Van~Gool, ``Daformer: Improving network architectures
  and training strategies for domain-adaptive semantic segmentation,'' in
  \emph{Proceedings of the IEEE/CVF Conference on Computer Vision and Pattern
  Recognition (CVPR)}, 2022, pp. 9924--9935.

\bibitem{mic}
L.~Hoyer, D.~Dai, H.~Wang, and L.~Van~Gool, ``Mic: Masked image consistency for
  context-enhanced domain adaptation,'' in \emph{Proceedings of the IEEE/CVF
  Conference on Computer Vision and Pattern Recognition (CVPR)}, 2023, pp.
  11\,721--11\,732.

\bibitem{sam4udass}
W.~Yan, Y.~Qian, H.~Zhuang, C.~Wang, and M.~Yang, ``Sam4udass: When sam meets
  unsupervised domain adaptive semantic segmentation in intelligent vehicles,''
  \emph{IEEE Transactions on Intelligent Vehicles}, vol.~9, no.~2, pp.
  3396--3408, 2024.

\bibitem{tits_active}
K.~Chitta, J.~M. Álvarez, E.~Haussmann, and C.~Farabet, ``Training data subset
  search with ensemble active learning,'' \emph{IEEE Transactions on
  Intelligent Transportation Systems}, vol.~23, no.~9, pp. 14\,741--14\,752,
  2022.

\bibitem{labor}
I.~Shin, D.-J. Kim, J.~W. Cho, S.~Woo, K.~Park, and I.~S. Kweon, ``Labor:
  Labeling only if required for domain adaptive semantic segmentation,'' in
  \emph{Proceedings of the IEEE/CVF International Conference on Computer Vision
  (ICCV)}, 2021, pp. 8588--8598.

\bibitem{iterative}
L.~Guan and X.~Yuan, ``Iterative loop method combining active and
  semi-supervised learning for domain adaptive semantic segmentation,''
  \emph{arXiv preprint arXiv:2301.13361}, 2023.

\bibitem{TUFL}
W.~Yan, Y.~Qian, C.~Wang, and M.~Yang, ``Threshold-adaptive unsupervised focal
  loss for domain adaptation of semantic segmentation,'' \emph{IEEE
  Transactions on Intelligent Transportation Systems}, vol.~24, no.~1, pp.
  752--763, 2023.

\bibitem{dropout}
N.~Srivastava, G.~Hinton, A.~Krizhevsky, I.~Sutskever, and R.~Salakhutdinov,
  ``Dropout: a simple way to prevent neural networks from overfitting,''
  \emph{The Journal of Machine Learning Research}, vol.~15, no.~1, pp.
  1929--1958, 2014.

\bibitem{bisenet}
C.~Yu, J.~Wang, C.~Peng, C.~Gao, G.~Yu, and N.~Sang, ``Bisenet: Bilateral
  segmentation network for real-time semantic segmentation,'' in \emph{European
  Conference on Computer Vision (ECCV)}.\hskip 1em plus 0.5em minus 0.4em\relax
  Springer, 2018, pp. 325--341.

\bibitem{deeplabv2}
L.-C. Chen, G.~Papandreou, I.~Kokkinos, K.~Murphy, and A.~L. Yuille, ``Deeplab:
  Semantic image segmentation with deep convolutional nets, atrous convolution,
  and fully connected crfs,'' \emph{IEEE Transactions on Pattern Analysis and
  Machine Intelligence}, vol.~40, no.~4, pp. 834--848, 2017.

\bibitem{neutral}
H.~Xu, M.~Yang, L.~Deng, Y.~Qian, and C.~Wang, ``Neutral cross-entropy loss
  based unsupervised domain adaptation for semantic segmentation,'' \emph{IEEE
  Transactions on Image Processing}, vol.~30, pp. 4516--4525, 2021.

\bibitem{class_weight}
Y.~Cui, M.~Jia, T.-Y. Lin, Y.~Song, and S.~Belongie, ``Class-balanced loss
  based on effective number of samples,'' in \emph{Proceedings of the IEEE/CVF
  Conference on Computer Vision and Pattern Recognition (CVPR)}, 2019, pp.
  9268--9277.

\end{thebibliography}
